\documentclass{article} %
\usepackage[accepted]{icml2024}
\usepackage[toc,page,header]{appendix}
\usepackage{minitoc}

\usepackage{natbib}
\usepackage{doi}
\usepackage{graphicx}
\usepackage{float}
\usepackage{amsmath,amsfonts,bm}

\def\eqref#1{equation~\ref{#1}}
\def\1{\bm{1}}

\DeclareMathAlphabet{\mathsfit}{\encodingdefault}{\sfdefault}{m}{sl}
\SetMathAlphabet{\mathsfit}{bold}{\encodingdefault}{\sfdefault}{bx}{n}

\newcommand{\Dcur}{D_{\mathrm{curated}}}

\newcommand{\Yxy}{\hat{Y}_{\mathcal{F}}(x,y)}

\newcommand{\Dcurated}{D_{\mathrm{curated}}}
\newcommand{\Doracle}{D_{\mathrm{oracle}}}
\newcommand{\Dtrain}{D_{\mathrm{train}}}
\newcommand{\Drest}{D_{\mathrm{discarded}}}
\newcommand{\Dsyn}{D_{\mathrm{syn}}}
\newcommand{\Dtest}{D_{\mathrm{test}}}
\newcommand{\Ddiscarded}{D_{\mathrm{discarded}}}
\newcommand{\name}{\texttt{CLLM}}

\usepackage{makecell}
\usepackage{hyperref}
\usepackage{float}

\hypersetup{
    colorlinks = true,
    linkcolor = blue,
    anchorcolor = blue,
    citecolor = blue,
    filecolor = blue,
    urlcolor = blue
    }
\usepackage{url}
\usepackage{makecell}
\usepackage{amsthm}
\usepackage{caption} %
\usepackage{subcaption}
\usepackage{booktabs}
\usepackage{wrapfig}
\usepackage{colortbl}
\usepackage{enumitem}
\usepackage{amsmath,amssymb}

\usepackage{tabularx}
\usepackage{listings}

\definecolor{codegreen}{rgb}{0,0.6,0}
\definecolor{codegray}{rgb}{0.5,0.5,0.5}
\definecolor{codepurple}{rgb}{0.58,0,0.82}
\definecolor{backcolour}{rgb}{0.95,0.95,0.92}
\definecolor{TealBlue}{rgb}{0., 0.0, 0}

\definecolor{originalcolor}{RGB}{77, 77, 77}
\definecolor{gpt4color}{RGB}{241, 106, 112}
\definecolor{tvaecolor}{RGB}{177, 216, 119}
\definecolor{oraclecolor}{RGB}{140, 220, 218}
\definecolor{gpt4nocontextcolor}{RGB}{0, 109, 119}

\lstdefinestyle{mystyle}{
    backgroundcolor=\color{backcolour},   
    commentstyle=\color{codegreen},
    keywordstyle=\color{magenta},
    numberstyle=\tiny\color{codegray},
    stringstyle=\color{codepurple},
    basicstyle=\ttfamily\footnotesize,
    breakatwhitespace=false,         
    breaklines=true,                 
    captionpos=b,                    
    keepspaces=true,                 
    numbers=left,                    
    numbersep=5pt,                  
    showspaces=false,                
    showstringspaces=false,
    showtabs=false,                  
    tabsize=2
}

\definecolor{ForestGreen}{RGB}{34, 139, 34}
\definecolor{Dandelion}{RGB}{253, 219, 109}
\definecolor{BrickRed}{RGB}{203, 65, 84}

\usepackage{tcolorbox}
\definecolor{lightblue}{HTML}{84C7F9}
\definecolor{lighterblue}{HTML}{D4ECFF}
\definecolor{coolblue}{HTML}{3967BD}

\newtcolorbox{mybox}{colback=lighterblue,colframe=lightblue}

\usepackage{pgfplots}
\usepackage[framemethod=TikZ]{mdframed}
\usepackage{comment}
\usepackage{fontawesome}
\usepackage{pgfplotstable}
\usepackage{colortbl}

\mdfsetup{
    middlelinecolor =   lightblue,
    middlelinewidth =   1.5pt,
    backgroundcolor =   lighterblue,
    roundcorner     =   5pt,
}

\usepackage{microtype}
\usepackage{graphicx}
\usepackage{arydshln}
\usepackage{booktabs} 
\usepackage{tabularx}
\usepackage{ltablex}
\usepackage{makecell}

\usepackage{hyperref}
\usepackage{makecell}
\usepackage{multirow}

\definecolor{headercolor}{rgb}{0.42, 0.56, 0.14} 
\definecolor{rowcolor}{rgb}{0.9, 0.9, 0.9} % A light gray color

\usepackage{colortbl}
\definecolor{Gray}{gray}{0.9}
\newcolumntype{g}{>{\columncolor{Gray}}c}

\usepackage{amsmath}
\usepackage{amssymb}
\usepackage{mathtools}
\usepackage{amsthm}

\usepackage{hyperref}
\hypersetup{
    colorlinks = true,
    linkcolor = blue,
    anchorcolor = blue,
    citecolor = black,
    filecolor = blue,
    urlcolor = blue
    }

\usepackage[capitalize,noabbrev]{cleveref}

\theoremstyle{plain}
\newtheorem{theorem}{Theorem}[section]

\theoremstyle{definition}
\newtheorem{definition}[theorem]{Definition}

\theoremstyle{remark}

\usepackage{booktabs}
\usepackage{pifont}
\usepackage{siunitx}

\usepackage[textsize=tiny]{todonotes}
\definecolor{ForestGreen}{RGB}{34, 139, 34}

\usepackage{pgfplots}
\usepackage[framemethod=TikZ]{mdframed}
\usepackage{comment}
\usepackage{fontawesome}
\usepackage{pgfplotstable}
\usepackage{colortbl}

\mdfsetup{% 
    middlelinecolor =   none,
    middlelinewidth =   1pt,
    backgroundcolor =   blue!5,
    roundcorner     =   5pt,
}

\icmltitlerunning{Curated LLM: Synergy of LLMs and Data Curation for tabular augmentation in low-data regimes}

\begin{document}

\twocolumn[
\icmltitle{Curated LLM: Synergy of LLMs and Data Curation \\ for tabular augmentation in low-data regimes}

\icmlsetsymbol{equal}{*}

\begin{icmlauthorlist}
\icmlauthor{Nabeel Seedat}{equal,cam}
\icmlauthor{Nicolas Huynh}{equal,cam}
\icmlauthor{Boris van Breugel}{cam}
\icmlauthor{Mihaela van der Schaar}{cam}
\end{icmlauthorlist}

\icmlaffiliation{cam}{University of Cambridge}

\icmlcorrespondingauthor{Nabeel Seedat}{ns741@cam.ac.uk}
\icmlcorrespondingauthor{Nicolas Huynh}{nvth2@cam.ac.uk}

\icmlkeywords{LLM, Synthetic data, data-centric AI}

\vskip 0.3in
]

\printAffiliationsAndNotice{\icmlEqualContribution}

\doparttoc
\faketableofcontents

\begin{abstract}
 Machine Learning (ML) in low-data settings remains an underappreciated yet crucial problem. Hence, data augmentation methods to increase the sample size of datasets needed for ML are key to unlocking the transformative potential of ML in data-deprived regions and domains. 
Unfortunately, the limited training set constrains traditional tabular synthetic data generators in their ability to generate a large and diverse augmented dataset needed for ML tasks. 
To address this challenge, we introduce \name, which leverages the prior knowledge of Large Language Models (LLMs) for data augmentation in the low-data regime.  However, not all the data generated by LLMs will improve downstream utility, as for any generative model. Consequently, we introduce a principled curation mechanism, leveraging learning dynamics, coupled with confidence and uncertainty metrics, to obtain a high-quality dataset. Empirically, on multiple real-world datasets, we demonstrate the superior performance of \name~ in the low-data regime compared to conventional generators. Additionally, we provide insights into the LLM generation and curation mechanism, shedding light on the features that enable them to output high-quality augmented datasets. 
\end{abstract}

\section{Introduction} \label{introduction}

 \textbf{No data, No Machine Learning.} 
 Machine learning (ML) has transformed numerous industries, but its wider adoption is hindered by a pervasive roadblock: insufficient data.
 Specifically, the use of ML algorithms presumes the availability and access to large datasets for training, be it labeled or unlabeled. Unfortunately, real-world domains are often data scarce: (i) in healthcare and finance, collecting annotations can be expensive or practically impossible; (ii) in developing and low-to-middle income countries (LMICs), digital infrastructure (such as electronic healthcare records (EHRs)) can be limited or nonexistent \citep{ade2023artificial,asiedu2023globalizing,owoyemi2020artificial,mollura2020artificial,alami2020artificial,ciecierski2022artificial} and (iii) within large datasets, there can be (ethnic) minorities that are underrepresented. This lack of data has serious consequences: to sideline these settings to the peripheries of ML advancements and prevent the development of accurate models. 
 How can we build a reliable ML model in this \emph{low-data regime}, with so few samples?
 Solving this problem is a major opportunity that would unlock the potential of ML across society, domains, and regions. 

\textbf{Aim.} To address this important yet undervalued low-data problem, we aim to augment the \emph{small labeled dataset} ($n<100$) with synthetic samples. We focus on tabular data, as defining augmentations is non-trivial and can easily result in nonsensical or invalid samples. Moreover, tabular domains like healthcare are often where data scarcity is acute.

\begin{figure*}[!t]
    \centering
\includegraphics[width=0.92\linewidth]{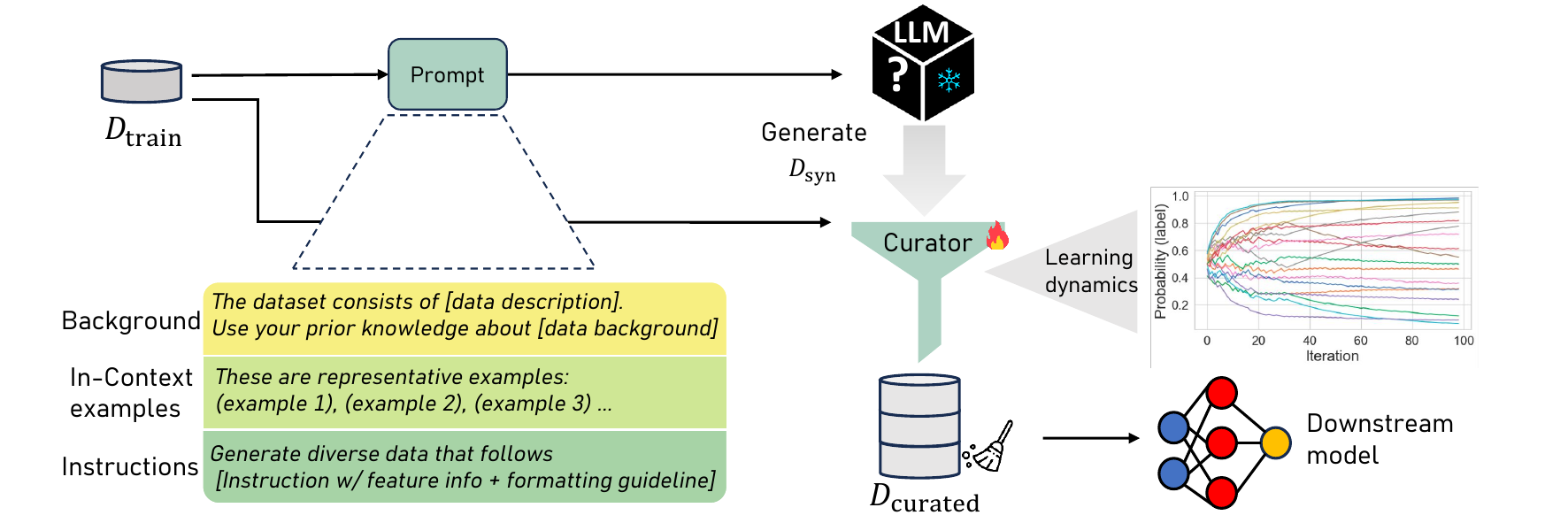}
    \caption{ \name~ uses a small dataset $\Dtrain$ and a frozen black-box LLM to generate a larger synthetic set $\Dsyn$. The curator computes the learning dynamics of samples in $\Dsyn$, assessing samples based on their aleatoric uncertainty and predictive confidence, then curates $\Dsyn$ with the goal that a downstream model trained on the curated $\Dcur$ will have improved performance.}
    \vspace{-3mm}
     \rule{\linewidth}{.5pt}
     \vspace{-8mm}
    \label{fig:overview}
\end{figure*}

\textbf{Related work.} Data augmentation is a widely used and different approach to address data scarcity in tabular data contexts. Methods are either based on generative models \citep{ghosheh2023synthesizing, biswas2023generative, wang2023enhancing,machado2022benchmarking,tanaka2019data} such as GANs \citep{xu2019modeling}, VAEs \citep{xu2019modeling}, Normalizing Flows \citep{papamakarios2021normalizing}, \textcolor{black}{Score-based models \citep{kotelnikov2022tabddpm, kim2022stasy}}, or alternatively traditional methods such as SMOTE \citep{chawla2002smote, wang2023enhancing,machado2022benchmarking}.  
However, in low-data regimes ($n<100$), the training data may not describe the full data distribution well, despite it being i.i.d. draws. Consequently, this harms conventional methods since the augmented data may not be sufficiently diverse and accurate, restricting the generalizability of predictive models trained on such data. 

Recent work has shown the potential of fine-tuning Large Language Models (LLMs) for tabular data generation \citep{borisov2023language}. While LLMs offer some degree of prior knowledge, there are two challenges in our setting. First, it is computationally expensive to fine-tune LLMs, while needing specialized hardware ---luxuries often not available in LMICs, thereby limiting applicability in such settings. Second, fine-tuning 
often assumes a large number of samples. In our low-data setting, it could lead to overfitting and low-quality generated samples, and hence poor downstream models---as we show for \cite{borisov2023language} in Sec. \ref{sec:exps}. 
Tangential to data augmentation, prior work has tackled data scarcity in the tabular setting via the lens of transfer learning or few-short learning, by using a knowledge graph \citep{margeloiu2022graph,ruiz2023enabling} (which might not be available) or a pretrained model \citep{levin2022transfer,jin2023benchmarking, tabllm}.
However, unlike data augmentation, these approaches are not flexible, as they tie the data customer to use a certain downstream predictor. We provide an extended discussion on this point in \cref{appendix:extended}.

\textbf{Curated LLMs.} To address the shortcomings of the aforementioned augmentation approaches, we propose \texttt{Curated LLM} (\name). First,  \name~ leverages the in-context capabilities of LLMs for generation, thereby reducing the computational burden compared to fine-tuning.  We also posit for the low-data regime; the diverse pretraining corpus of LLMs carries valuable prior knowledge, which may offer more diversity in their generation compared to other conventional tabular generators. \emph{Of course, LLMs are not perfect}. Consequently, balancing the utility of LLMs against the risk of noisy, irrelevant data is important to ensure reliable downstream performance. Hence, this necessitates systematic assessment of the generated data. In fact, this issue is vital for \emph{any} generative model.

 This motivates the second key aspect of \name, i.e. a post-generation data curation mechanism. This addresses the \emph{overlooked} aspect that not all of the synthetic samples are useful to downstream model performance, with some samples even harmful. 
 We anchor our approach with ideas from learning theory that show the behavior of individual data samples during training, called learning dynamics, provides a salient signal about the value of samples to a learner \citep{arpit2017closer,arora2019fine,li2020learning}. To provide intuition, samples with variable predictions might be considered ambiguous or other samples might never be learned correctly and could harm a model.  In \name, we study the learning dynamics of the synthetic data samples, with respect to a model trained on the small real dataset.
We then analyze these dynamics by computing two key metrics: confidence and aleatoric (data) uncertainty. These metrics form the basis for curating the synthetic samples. We then aim to enable a highly performant downstream model when trained on the curated dataset.

\begin{mdframed}[leftmargin=0pt, rightmargin=0pt, innerleftmargin=10pt, innerrightmargin=10pt, skipbelow=0pt]
\textbf{Contributions:} \name~ is a novel data augmentation approach allying the strengths of LLMs with a robust data curation mechanism to improve data augmentation in the 
\emph{low-data regime} $(n < 100)$, 
bringing several contributions: 
 \textbf{\textcolor{ForestGreen}{\textcircled{1}} Improved performance:} we empirically demonstrate on 7 real-world datasets that \name~ enables superior downstream performance compared to 6 widely used tabular data generative models and data augmentation techniques.
\textbf{\textcolor{ForestGreen}{\textcircled{2}} Value of curation:} we show the \emph{overlooked} aspect of synthetic data curation improves downstream performance across the generative models. This highlights the flexibility and broad utility of our curation mechanism for data augmentation. \textbf{\textcolor{ForestGreen}{\textcircled{3}} Insights:}  we dissect the two aspects of \name~(LLM and data curation) along a variety of dimensions, providing insights and understanding into why the approach is beneficial. We show the largest gains are for underrepresented subgroups and in low-data settings. These contributions, which address the data logjam, pave the way towards wider usage of ML across society, domains 
and regions.
\end{mdframed}

\vspace{-0mm}
\section{\name: Synergy of LLM Generation and Data Curation}
\vspace{-0mm}
\textbf{Set-up.} Given feature space $\mathcal{X}$, and label space $\mathcal{Y} = \{1, ..., k\}$, we assume that we only have a small labeled dataset $\Dtrain = \{(x_i, y_i)\}_{i=1}^{n}$, with $x_i \in \mathcal{X}$, $y_i \in \mathcal{Y}$ and $n<100$ (\textcolor{TealBlue}{low data setting}). Assume $\Dtrain$ is drawn i.i.d. from the real distribution $p_R(X,Y)$. We also assume access to a pretrained LLM to generate samples. 
We denote the output distribution of the LLM as $p_\Phi(X,Y)$, with $\Phi$ containing parameters that we control (e.g., input prompts). Our goal is to generate a dataset to augment the small $\Dtrain$, and subsequently use it to train a classifier $f: \mathcal{X} \rightarrow \mathcal{Y}$. 
Successful augmentation will provide a better classifier $f$, than if we had trained $f$ on the small $\Dtrain$ itself. We measure downstream performance on a separate held-out dataset of real data, $\Dtest$.

\textbf{Our Approach.} To address this challenge, we introduce \name, an approach for data augmentation in low-data regimes. 
As shown in Figure \ref{fig:overview}, \name~ leverages LLMs to \textbf{generate} a synthetic dataset $\Dsyn$ using a small dataset $\Dtrain$ (Sec.~\ref{sec:generate}). It exploits the LLMs' prior knowledge via \textcolor{TealBlue}{in-context learning (ICL)} and contextual information. \name~ then \textbf{curates} $\Dsyn$ by analyzing the learning dynamics of samples in $\Dsyn$ based on predictive confidence and aleatoric (data) uncertainty. These metrics are obtained by training a supervised model on $\Dtrain$. We leverage them to define a curated dataset $\Dcur$, which is used to train a downstream classifier (Sec.~\ref{sec:curation}). 

In each sub-section we describe and motivate the design of the different aspects of \name~ (LLM and curation mechanism). Furthermore, we provide insights and understanding into their role in improving data utility, which
we later quantify on multiple real-world datasets in Sec.~\ref{sec:exps}.

  \vspace{-0mm}
\subsection{Data generation with LLMs based on a small $\Dtrain$}\label{sec:generate}
  \vspace{-0mm}

As outlined in Sec.~\ref{introduction}, in the low-data regime, conventional tabular generative models (e.g. CTGAN, TVAE) are constrained by the limited $\Dtrain$ and may not generate sufficiently diverse and/or accurate synthetic data. To address this, we propose to leverage LLMs, building on their large-scale pretraining. We first outline the appealing properties of LLMs for tabular data generation when we have very few samples, then describe design choices to exploit these. 
\begin{itemize}[leftmargin=*]
    \itemsep0em 
    \vspace{-0mm}
    \item \textbf{Prior knowledge.} LLMs are pretrained with a vast corpus of information \citep{chowdhery2022palm, singhal2023large}. When prompted to generate samples with limited real data, LLMs can leverage this encoded prior information about similar problems and feature-label relationships to enhance both accuracy and diversity of generation. 
    \item \textbf{Contextual understanding.} LLMs can process background and contextual information about the problem via natural language \citep{yang2023large}. For example, a high-level description of the task, features and their meanings can be conveniently described through natural language. Such information is unavailable to conventional generators that only utilize numerical examples. 
    \item \textbf{Few-shot capabilities.}  LLMs have demonstrated proficiency in generalizing to tasks with just a few examples \citep{brown2020language,wei2023larger,mirchandani2023large}. In the context of generation, we envision the idea of in-context generation using limited real examples. 
\end{itemize}

To benefit from these capabilities, we craft the LLM prompt with three different parts (see Fig. \ref{fig:overview}): (1) \emph{Background}: text description of the dataset and task (e.g. predict Covid mortality). Additionally, we include a description of what each feature means, explicitly prompting the LLM to use prior knowledge about these features. (2) \emph{Examples}: we serialize the samples in $\Dtrain$ as example demonstrations and provide both the features and the label in text format.  
(3) \emph{Instructions}: To generate a synthetic dataset $\Dsyn$, we instruct the LLM to leverage the contextual information and provided examples as an i.i.d. draw from the distribution. We instruct the LLM to identify structural and feature-label relationships in the data and generate diverse data following the structure and format of the provided examples.
We provide more details on the prompts in Appendix \ref{appx:B}.

\textbf{Motivation for a frozen LLM.} Using a frozen black-box LLM (e.g. GPT-4 or GPT-3.5) is computationally cheaper and requires less specialized hardware (i.e. GPUs) compared to fine-tuning. This relates to settings described in Sec. \ref{introduction}, such as LMICs, where we may not have the computational resources to fine-tune an LLM. Even in settings where fine-tuning is possible, we show empirically in Sec. \ref{sec:exps} that LLM fine-tuning (e.g. GReaT baseline) is suboptimal in low-data settings ($n<100$) compared to providing in-context examples coupled with curation.
\begin{mybox}
    \textbf{Dissecting the LLM's generative features. } 
    We now investigate various dimensions to understand and illustrate empirically the appealing features of LLMs as data generators in the low-data regime, and how our design choices unlock them. We use the Brazil \emph{Covid-19} dataset \citep{covid} as a running example and focus on GPT-4 as the LLM. 
\end{mybox}

\begin{figure}[!t]
    \centering
    \includegraphics[scale=0.2]{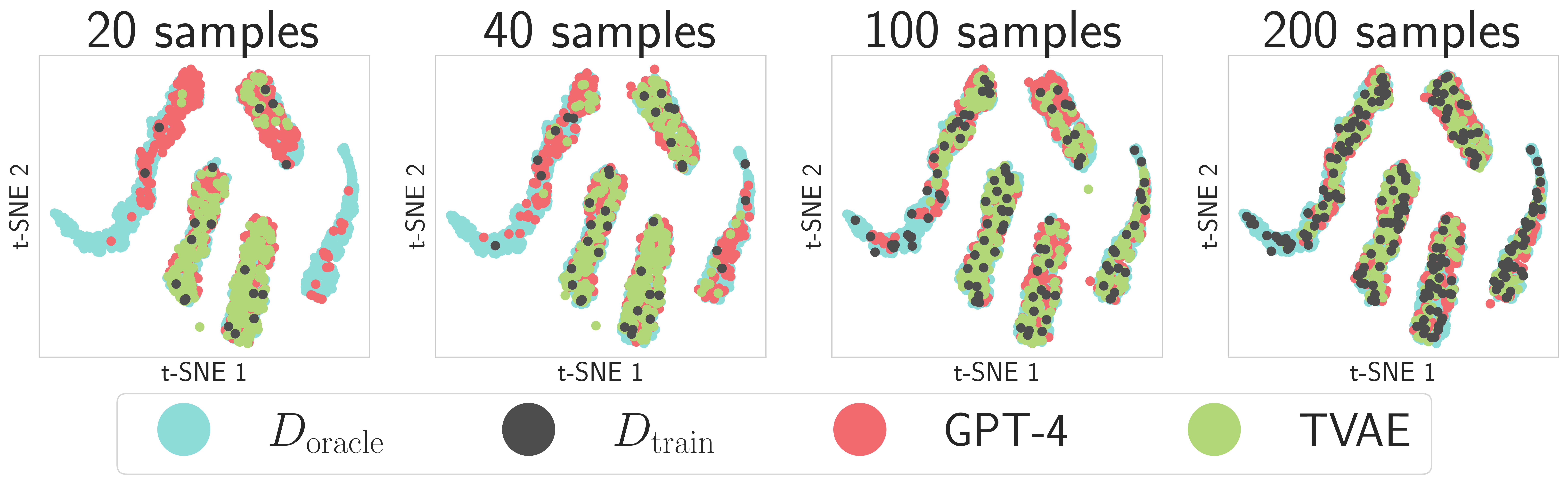}
    \caption{\textcolor{gpt4color}{GPT-4} is able to extrapolate to regions of the \textcolor{oraclecolor}{oracle} (true manifold) even where there is no training data covering them, \textcolor{TealBlue}{as can be seen by the overlap with the turquoise dots}, with the effect more pronounced when $\Dtrain$ is small}
    \label{tsne-understand}
     \rule{\linewidth}{.5pt}
\end{figure}

$\blacktriangleright$ \textcolor{black}{\textbf{Extrapolation to unseen regions of the manifold.}} We compare the samples generated by \textcolor{gpt4color}{GPT-4} to \textcolor{tvaecolor}{TVAE}, a widely used tabular data generator. We consider $\Doracle$, a held-out dataset from the same distribution as $\Dtrain$, such that $\vert\Doracle\vert \gg \vert\Dtrain\vert$, thereby providing an approximation for the true manifold.  The t-SNE plots in Fig.~\ref{tsne-understand}
shows, when \textcolor{originalcolor}{$\Dtrain$} is very small ($n=20$ samples), that its samples do not cover all regions of \textcolor{oraclecolor}{$\Doracle$}. For example, \textcolor{originalcolor}{$\Dtrain$} does not contain samples from specific demographic subgroups (e.g. people with age 40 or below).
As expected, \textcolor{tvaecolor}{TVAE} only generates samples constrained by the limited \textcolor{originalcolor}{$\Dtrain$}. In contrast, \textcolor{gpt4color}{GPT-4} is able to extrapolate and generate samples even in unseen regions of \textcolor{originalcolor}{$\Dtrain$}, thereby better covering \textcolor{oraclecolor}{$\Doracle$}. This stems from its \emph{contextual understanding} of the features, unlocking the use of its \emph{prior knowledge}. It leads to better coverage in the low-data regime, consequently aiding in superior downstream performance, as shown in Table \ref{table:results_comparison}. As $n$ increases ($\geq 100$), $\Dtrain$ provides better coverage, which naturally benefits both \textcolor{gpt4color}{GPT-4} and \textcolor{tvaecolor}{TVAE}. Overall, this result shows how prior knowledge encoded in LLMs addresses shortcomings of conventional generative approaches (e.g. \textcolor{tvaecolor}{TVAE}) in the low-data regime.

$\blacktriangleright$
 \textbf{GPT-4 benefits underrepresented groups the most.} Having illustrated the extrapolation capabilities of GPT-4, we now ask:~\emph{where does augmentation benefit downstream performance the most?} We evaluate performance gains for different demographic subgroups, such as age groups and ethnic groups (Amarela, Prada). 
 \begin{wrapfigure}{r}{0.25\textwidth}
        \centering
        \includegraphics[width = 1.05\linewidth]{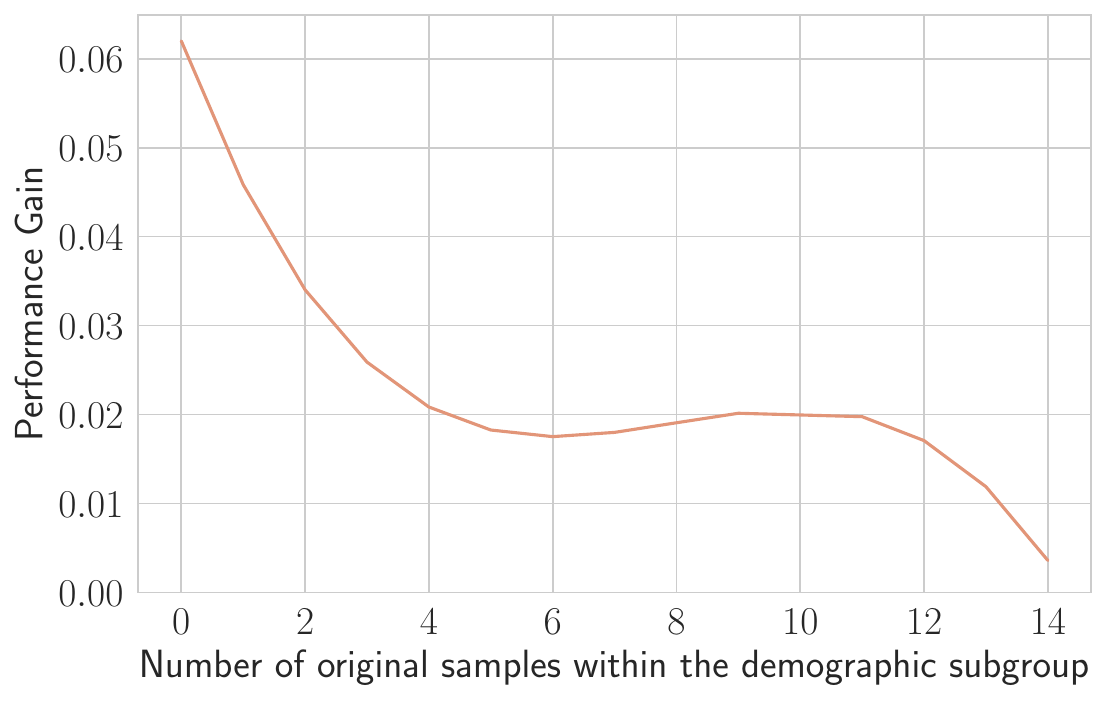}
        \captionof{figure}{Subgroups with fewest samples in $\Dtrain$ benefit the most from data augmentation, on average.}
        \label{fig:nsweep}
     \rule{\linewidth}{.5pt}
     \vspace{-5mm}
 \end{wrapfigure}
 Fig.~\ref{fig:nsweep} shows the performance gain obtained by training a classifier on data generated by GPT-4 compared to training on the small $\Dtrain$. The greatest gains, on average, are for subgroups for which we have \emph{no data} in $\Dtrain$, yet GPT-4 can extrapolate and generate samples for these subgroups. This further validates the rationale of extrapolation via prior knowledge as a key source of gain for GPT-4.

 Table \ref{tab:subgroup} shows fine-grained results (across 10 different seeds) for the 5 subgroups that benefit the most from data augmentation, which are small-sized demographic subgroups. This finding has real-world implications for equity, showing we can improve performance for underrepresented subgroups even when we lack data or collecting data is difficult/costly.

    \begin{table}[!h]
        \vspace{-0mm}
        \centering
        \caption{Deep dive into the top 5 demographic subgroups \textcolor{TealBlue}{in the Covid dataset} with the largest gains, across 10 seeds, for $\vert \Dtrain \vert=20$. GPT-4 improves performance on the smallest groups.}
        \vspace{-1mm}
        \scalebox{0.85}{
            \begin{tabular}{c|c|cc}
                \hline
                Subgroup & $n_{\mathrm{samples}}$ in $\Dtrain$& \multicolumn{2}{c}{ Avg. Acc. Gain v. $\Dtrain$} \\
                & (min - max)  & GPT-4 & TVAE  \\
                \hline
                Age\_40 & 0-6  & \textbf{6.38} +- \footnotesize{2.09} & -3.37 +- \footnotesize{2.86} \\
                Liver & 0-1  & \textbf{3.85} +- \footnotesize{3.37} & -13.1 +- \footnotesize{3.38} \\
                Renal & 0-3  & \textbf{4.52} +- \footnotesize{2.01} & -18.0 +- \footnotesize{3.22} \\
                Amarela & 0-1 & \textbf{8.71} +- \footnotesize{1.40} & -2.03 +- \footnotesize{2.88} \\
                Parda & 3-11   & \textbf{5.07} +- \footnotesize{1.50}  & -6.57 +- \footnotesize{1.61}\\
                \hline
            \end{tabular}
        \label{tab:subgroup}
        }
    \end{table}

$\blacktriangleright$ 
\textbf{Importance of contextual information in the prompt. } A natural question is:~\emph{how important is the prompt to elicit the prior knowledge of the LLM?} We explore two variants:
(1) \textcolor{gpt4color}{\emph{Prompt w/ context}}:  provides contextual information including background about the dataset, feature names and descriptions (our approach) 
and (2) \textcolor{gpt4nocontextcolor}{\emph{Prompt w/ no context}}: only provides the numerical in-context examples (ablation).  
\begin{wrapfigure}{r}{0.275\textwidth}
    \centering
    \includegraphics[width=1\linewidth]{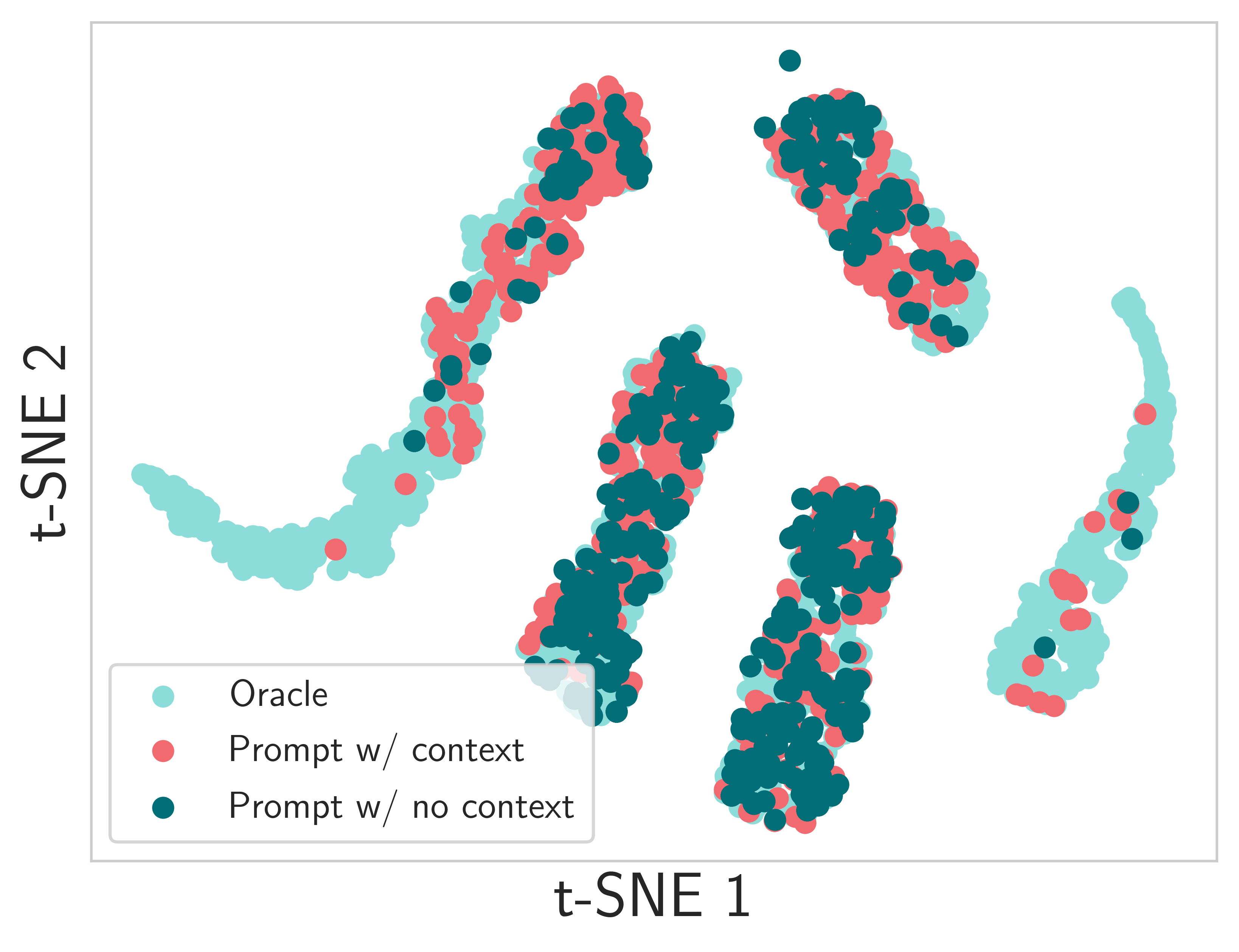}
    \captionof{figure}{Contextual information in the prompt is important for extrapolation.}
    \label{fig:prompt}
     \rule{\linewidth}{.5pt}
     \vspace{-5mm}
\end{wrapfigure}
Fig.~\ref{fig:prompt} qualitatively shows that not including contextual knowledge in the prompt gives lower coverage of \textcolor{oraclecolor}{$\Doracle$} with less extrapolation beyond $\Dtrain$. We quantify this in Table \ref{tab:metrics_covid} using Precision (Quality) and Recall (Diversity) metrics \citep{sajjadi2018assessing}, as well as Utility (Downstream performance). \textcolor{gpt4color}{GPT-4} \emph{with contextual information} has superior precision and recall in the low data setting. Furthermore, we show that \emph{ the lack} of contextual information in the prompt significantly harms the precision (quality) of the data even compared to TVAE. This highlights that LLMs need guidance, as we are only able to get the extrapolation and performance benefits by including contextual information, further motivating our design choices in the prompt. We conduct the same experiment with the Compas dataset in Appendix \ref{app:compas_context}.

\begingroup
\setlength{\tabcolsep}{1pt} %
\renewcommand{\arraystretch}{1}
\begin{table}[!h]
    \centering
    \caption{Including contextual information in the prompt improves  precision (P), recall (R), and utility (U) in low-sample settings (\textcolor{TealBlue}{results shown for the Covid  dataset}).}
    \scalebox{0.55}{

\begin{tabular}{c|ccccccccc}
\hline
\makecell{$n_{\mathrm{samples}}$ \\ in \\ $\Dtrain$} & \multicolumn3{c}{\makecell{GPT-4 \\  w/ context}} & \multicolumn3{c}{\makecell{GPT-4 \\  no context}} & \multicolumn3{c}{TVAE} \\
\cmidrule(lr){2-4}\cmidrule(lr){5-7}\cmidrule(lr){8-10}
 & P & R & U & P & R & U & P & R & U  \\
\hline
20 & $\bf 0.41_{\scriptsize(0.04)}$ & $\bf 0.87_{\scriptsize(0.03)}$ & $\bf 0.74_{\scriptsize(0.01)}$ & $0.13_{\scriptsize(0.0)}$ & $0.82_{\scriptsize(0.01)}$ & $0.66_{\scriptsize(0.01)}$ & $0.33_{\scriptsize(0.07)}$ & $0.50_{\scriptsize(0.03)}$ & $0.59_{\scriptsize(0.02)}$\\
40 & $\bf 0.40_{\scriptsize(0.01)}$ & $\bf 0.91_{\scriptsize(0.01)}$ & $\bf 0.76_{\scriptsize(0.0)}$ & $0.11_{\scriptsize(0.0)}$ & $0.89_{\scriptsize(0.0)}$ & $0.69_{\scriptsize(0.0)}$ & $0.27_{\scriptsize(0.01)}$ & $0.68_{\scriptsize(0.01)}$ & $0.62_{\scriptsize(0.03)}$\\
100 & $\bf 0.42_{\scriptsize(0.01)}$ & $0.86_{\scriptsize(0.02)}$ & $\bf 0.75_{\scriptsize(0.01)}$ & $0.11_{\scriptsize(0.01)}$ & $\bf 0.90_{\scriptsize(0.01)}$ & $0.74_{\scriptsize(0.01)}$ & $0.39_{\scriptsize(0.02)}$ & $0.67_{\scriptsize(0.03)}$ & $0.64_{\scriptsize(0.06)}$\\
200 & $0.44_{\scriptsize(0.02)}$ & $0.85_{\scriptsize(0.02)}$ & $\bf 0.75_{\scriptsize(0.0)}$ & $0.08_{\scriptsize(0.01)}$ & $\bf 0.90_{\scriptsize(0.0)}$ & $0.60_{\scriptsize(0.01)}$ & $\bf 0.47_{\scriptsize(0.0)}$ & $0.73_{\scriptsize(0.01)}$ & $0.65_{\scriptsize(0.02)}$\\
   \hline
\end{tabular}}
        \label{tab:metrics_covid}
\end{table}

\subsection{Data curation with learning dynamics}\label{sec:curation}
When prompted with $\Phi$ (which contains the in-context samples of $\Dtrain$), the LLM  generates samples from a distribution $p_\Phi(X,Y)$ that approximates $p_R(X,Y)$, 
implicitly exploiting its large-scale pretraining and few-shot capabilities. LLMs are of course not perfect and could generate noisy samples, hence this distribution may be inaccurate %
\footnote{We could finetune the model on the scarce $\Dtrain$ we have, but is likely to still lead to overfitting due to the extreme data scarcity and LLM parameter size.}. To make this distribution more relevant to the downstream task, we include a data curation mechanism. Specifically, we focus on the noisy feature-label relationship $p_{\Phi}(Y \vert X)$, for which we expect $p_{\Phi}(Y \vert X) \neq p_{R}(Y \vert X)$ given the small size of $\Dtrain$. This motivates us to curate $\Dsyn$ and discard likely mislabeled samples.

We anchor our approach with ideas from learning theory that show that the behavior of individual samples during model training (called \emph{learning dynamics}) contains signal about the nature of the samples themselves \citep{arpit2017closer,arora2019fine,li2020learning}.
Some samples are easily and confidently predicted over different model checkpoints, whereas other samples might be challenging (e.g. due to mislabeling) and hence might be incorrectly predicted for the given label. Consequently, we operationalize \emph{learning dynamics} as the basis of our proposed curation mechanism. Specifically, we analyze samples in $\Dsyn$ by studying their learning dynamics computed with a classifier trained on $\Dtrain$. We then categorize and filter samples in $\Dsyn$, and produce a curated dataset $\Dcur \subset \Dsyn$. 

\textbf{Learning dynamics.} We now formalize how we compute learning dynamics for individual samples. Assume that a classifier $f$ is trained in an iterative scheme (e.g. neural networks or XGBoost trained over iterations) on $\Dtrain$, which makes it possible to analyze the learning dynamics of samples in $\Dsyn$ over these iterations. \textcolor{TealBlue}{The classifier $f$ should be at least as flexible as the model that the practitioner intends to use for the downstream task.} $f$ is trained from scratch on $\Dtrain$ and goes through $e \in [E]$ different checkpoints leading to the set $ \mathcal{F} = \{f_{1} , f_{2}, \dots, f_{E} \}$, such that $f_{e}$ is the classifier at the $e$-th checkpoint. Let $[f_{e}(x)]_{y}$ denote the predicted probability for class $y$ and sample $x$. Our goal is to assess the learning dynamics of samples in $\Dsyn$ over these $E$ training checkpoints, while we train $f$ on $\Dtrain$. For this, we
define $H$, a random variable following a uniform distribution $\mathcal{U}_{\mathcal{F}}$ over the set of checkpoints $\mathcal{F}$.
Specifically, given $H = h$ and a sample $(x,y)$, we define the correctness in the prediction of $H$ as a binary random variable $\Yxy$ with the following conditional:  $P(\Yxy = 1 \vert H=h) = [h(x)]_{y}$ and $P(\Yxy = 0 \vert H = h) = 1 - P(\Yxy = 1 \vert H=h)$.

\textbf{Curation metrics.} Equipped with a probabilistic interpretation of the predictions of a model, we now define two characterization metrics that we use for curation: (i) average confidence and (ii) aleatoric (data) uncertainty, inspired by \citep{kwon2018uncertainty, seedat2022data}.

\begin{definition}[Average confidence]
For any set of checkpoints $\mathcal{F}= \{f_1, ..., f_E \}$, the average confidence for a sample $(x,y)$ is defined as the following marginal:
\begin{align*} \bar{P}_{\mathcal{F}}(x,y) &:= P(\Yxy = 1) \\
&=\mathbb{E}_{H \sim \mathcal{U}_{\mathcal{F}}}[P(\Yxy = 1 \vert H)] \\ &=\frac{1}{E}\sum_{e = 1}^{E} [f_{e}(x)]_{y}
\end{align*}

\end{definition}
\begin{definition}[Aleatoric uncertainty]
For any set of checkpoints $\mathcal{F}= \{f_1, ..., f_E \}$, the aleatoric uncertainty for a sample $(x,y)$ is defined as:
\begin{align*}
    v_{al,\mathcal{F}}(x,y) &:= \mathbb{E}_{H \sim \mathcal{U}_{\mathcal{F}}}[Var(\Yxy \vert H)] \\ 
    &=\frac{1}{E}\sum_{e = 1}^{E} [f_{e}(x)]_{y}(1-[f_{e}(x)]_{y}) 
\end{align*}

\end{definition}

Intuitively, for binary classification ($k=2$), the aleatoric uncertainty for a sample $x$ is maximized when  $[f_{e}(x)]_{y} = \frac{1}{2}$ for all checkpoints $f_e$, akin to random guessing. Recall aleatoric uncertainty captures the inherent data uncertainty, hence is a principled way to capture issues such as mislabeling. This contrasts epistemic uncertainty, which is model-dependent and can be reduced simply by increasing model parameterization \citep{hullermeier2021uncerainty}.

Having defined sample-wise confidence and aleatoric uncertainty, we categorize samples in $\Dsyn$ as \textit{Selected} or \textit{Discarded}: for a sample $(x,y)$, a set of training checkpoints $\mathcal{F}$, and two thresholds $\tau_{\mathrm{conf}}$ and $\tau_{\mathrm{al}}$, we define the category $c(x,y, \mathcal{F})$ as
\textit{Discarded}  if $\Bar{P}_{\mathcal{F}}(x,y) < \tau_{\mathrm{conf}}$ and $v_{al, \mathcal{F}}(x,y) < \tau_{\mathrm{al}}$,  and
\textit{Selected}  otherwise.

Hence, a \textit{Discarded} sample is one for which we have a very low confidence in predicting its associated label whereas we also have low inherent data uncertainty. Finally, given a function $f$ associated with the set of checkpoints $\mathcal{F}$, we define the curated set $\Dcur = \{(x,y) \vert (x,y) \in \Dsyn, c(x,y,\mathcal{F}) = \textit{Selected}\}$. We also define  $\Ddiscarded = \Dsyn \setminus \Dcurated$.

To summarize, the objective of the curation step is that training on the curated synthetic data leads to a better classifier $f_{\Dcur}$ for the downstream task, compared to training on the uncurated synthetic data, i.e.   $M(f_{\Dcur}) > M(f_{\mathcal{\Dsyn}})$, where $M$ is a performance measure (for example accuracy).
In Sec.~\ref{sec:exps}, we empirically show how performance on this curated dataset is superior both for LLM generated data, as well as other classes of generative models.

\begin{mybox}
\textbf{Dissecting the role of curation.} 
We now empirically demonstrate the role of curation in correcting the noisy feature-label relationship present in $\Dsyn$, highlighting two insights: \\ (i) curation discards samples which are atypical in their label with respect to their neighbors in $\Dsyn$ (ii) discarded samples can be considered ``mislabeled", and we quantify their atypicality using a large held-out dataset $\Doracle$. 
\end{mybox}

$\blacktriangleright$  \textbf{Discarded samples conflict on the label with their neighbors in $\Dsyn$.}
We audit every synthetic sample $(x,y)$ generated by GPT-4 (across 7 datasets) and compute the proportion of its $k$ nearest neighbors in $\Dsyn$ which share the same label $y$. The agreement with the neighbors assesses the typicality of a sample's $y$ given $x$, where naturally lower agreement is linked to mislabeling, which we aim to detect via curation. Taking $k$ = 10, we obtain an average agreement of  $a_{\mathrm{curated}} = \textbf{0.74}$ for $\Dcur$, compared to $a_{\mathrm{discarded}} = \textbf{0.58}$ for $\Ddiscarded$. This shows that the samples removed by our curation mechanism are those which, despite having similar features $x$, do not agree with the labels of their surrounding neighbors. This corroborates ideas in \citep{ashmore2021assuring} of how proximity violations are useful to guide remedial action to improve models. Not removing these mislabeled samples injects noise into the downstream classifier, thus reducing performance.

$\blacktriangleright$ \textbf{Assessing discarded samples with  $\Doracle$.}
Ideally, the samples we select should better align with the true feature-label distribution. Since we don't have access to this distribution explicitly, we compute a proxy for $\eta(x) = \arg\max_{y} p(Y =y \vert X = x)$, which we call $\hat{\eta}$. It is obtained by training a classifier on a held-out dataset $\Doracle$---the same size as $\Dtest$ and an order of magnitude larger than $\Dtrain$. For each synthetic method, we then report the accuracy of $\hat{\eta}$ on both the curated  $\Dcurated$ and discarded $\Drest$ datasets ---see Fig. \ref{fig:bayes_classifier}.

\begin{figure}[h]
    \centering
    \includegraphics[width = 0.475\textwidth]{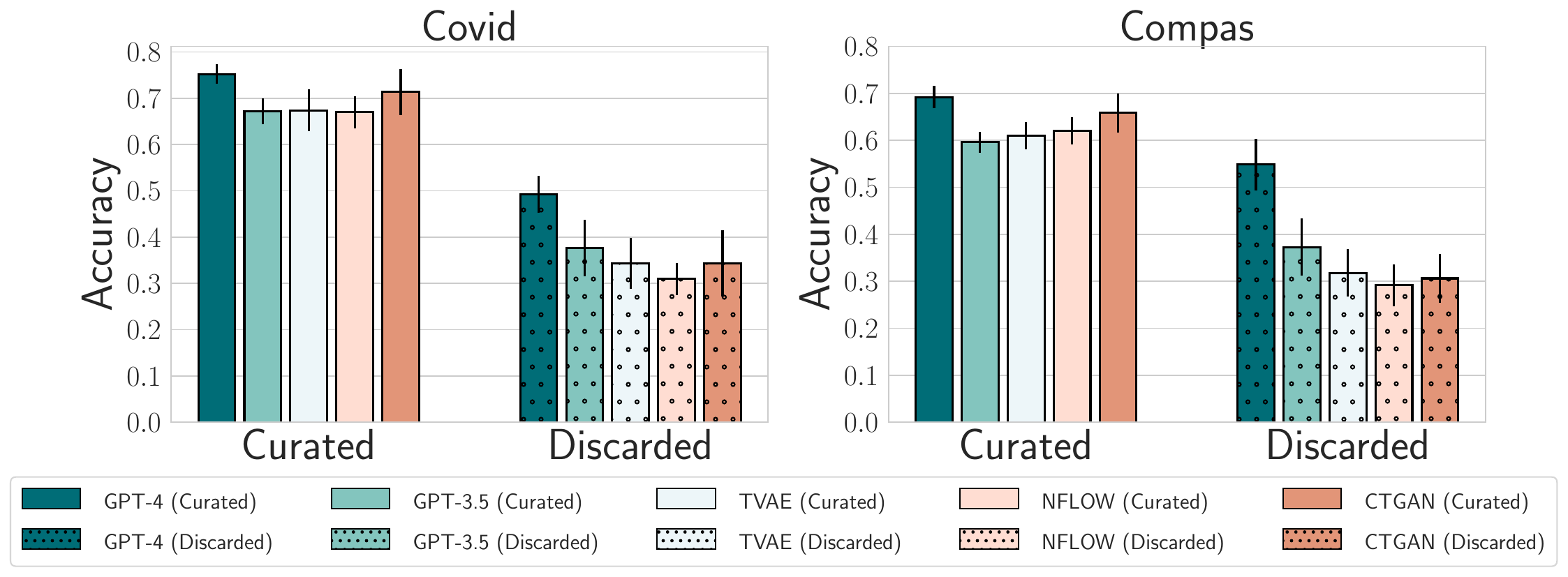}
    \caption{$\hat{\eta}$ aligns more with $\Dcur$ than $\Ddiscarded$ for each generative model: the curation step keeps high quality samples tailored to the downstream task.}
    \label{fig:bayes_classifier}
     \rule{\linewidth}{.5pt}
\end{figure}

We highlight two key observations. First, the curated datasets, for all the generative models, exhibit a higher agreement with the proxy $\hat{\eta}$ than the discarded datasets. This aligns with the desideratum of only keeping samples that exhibit the correct feature-label relationships. This provides a rationale for why curation helps improve discriminative performance, as samples in $\Dcurated$ are much more likely to have the correct feature-label relationship. 

Second, GPT-4 has a higher agreement with $\hat{\eta}$ on $\Drest$, compared to other generators. This illustrates that GPT-4's prior knowledge enables it to better capture the distribution $p(Y|X=x)$. Note that generative baselines (e.g. TVAE) model the joint $p(X,Y)$, \emph{without any context} of which is the set of features and which is the label. In contrast, we can define in the LLM prompt which column is the target $Y$, allowing the LLM to better capture the feature-label relationships. 
This complements the findings from Fig.~\ref{tsne-understand}, which showed that GPT-4 extrapolates to unseen regions of the feature manifold, captured by the support of $p(X)$.

\section{Curated LLMs for Better Data Augmentation}\label{sec:exps}
We now perform an end-to-end quantitative evaluation of \name~ \footnote{Code: https://github.com/seedatnabeel/CLLM or https://github.com/vanderschaarlab/CLLM}  across multiple real-world datasets, for \textbf{downstream utility}, demonstrating the value of allying the generative capabilities of LLMs with our curation mechanism. 

Sec.~\ref{performance} compares the downstream performance of models when trained on uncurated vs curated data for a variety of augmentation approaches.
Having evaluated \name~ on a range of datasets, we also demonstrate how we can leverage information extracted during curation to characterize datasets via a \textbf{hardness proxy}. Sec.~\ref{reliability} illustrates how our characterization of samples during the curation step can help to flag synthesized datasets (e.g via the LLM) which, if used for training, will result in poor downstream performance.

\begingroup
\setlength{\tabcolsep}{1.5pt} %
\renewcommand{\arraystretch}{1} %

\begin{table*}[!t]
\caption{ AUC averaged over 4 downstream models on $\Dtest$. Curation improves performance for all methods across all sample sizes $n$, as indicated by \textcolor{ForestGreen}{$\uparrow$}. \name~ w/ GPT-4 (Cur.) provides the strongest performance for both \colorbox{orange!20} {private/proprietary datasets} and \colorbox{green!20} {public datasets}}
\centering
\scalebox{0.75}{
\begin{tabular}{l|cc|cc|cc|cc|cc|cc|cc|cc|cc}
\toprule
& \multicolumn{2}{c|}{Real data} & \multicolumn{4}{c|}{\name ~(OURS)} &  \multicolumn{12}{c}{Baselines} \\
\cmidrule(lr){2-3}\cmidrule(lr){4-7}\cmidrule(lr){8-19}
& \multicolumn{2}{c|}{} & \multicolumn{2}{c}{GPT-4} & \multicolumn{2}{c|}{GPT-3.5} & \multicolumn{2}{c}{CTGAN} & \multicolumn{2}{c}{TabDDPM} & \multicolumn{2}{c}{GReaT}   & \multicolumn{2}{c}{NFLOW} & \multicolumn{2}{c}{SMOTE} & \multicolumn{2}{c}{TVAE} \\
\cmidrule(lr){2-3}\cmidrule(lr){4-5}\cmidrule(lr){6-7}\cmidrule(lr){8-9}\cmidrule(lr){10-11}\cmidrule(lr){12-13}\cmidrule(lr){14-15}\cmidrule(lr){16-17}\cmidrule(lr){18-19}
Dataset & $\Doracle$ & $\Dtrain$ & Uncur. & Cur. & Uncur. & Cur. & Uncur. & Cur. & Uncur. & Cur. & Uncur. & Cur. & Uncur. & Cur. & Uncur. & Cur. & Uncur. & Cur. \\
\hline\hline
\colorbox{orange!20} {covid (n=20)} & 74.41 & 68.50 & 73.78 & \textbf{73.87 \textcolor{ForestGreen}{$\uparrow$}} & 69.85 & 71.41 \textcolor{ForestGreen}{$\uparrow$} & 59.00 & 63.67 \textcolor{ForestGreen}{$\uparrow$} & 66.84 & 66.85 \textcolor{ForestGreen}{$\uparrow$} & 57.38 & 66.46 \textcolor{ForestGreen}{$\uparrow$} & 62.87 & 68.56 \textcolor{ForestGreen}{$\uparrow$} & 66.95 & 66.82 & 61.69 & 66.11 \textcolor{ForestGreen}{$\uparrow$} \\
\colorbox{orange!20} {cutract (n=20)} & 72.23 & 70.12 & 71.15 & \textbf{72.50 \textcolor{ForestGreen}{$\uparrow$}} & 69.97 & 71.54 \textcolor{ForestGreen}{$\uparrow$} & 64.01 & 67.98 \textcolor{ForestGreen}{$\uparrow$} & 66.05 & 66.59 \textcolor{ForestGreen}{$\uparrow$} & 52.38 & 67.02 \textcolor{ForestGreen}{$\uparrow$} & 64.44 & 70.42 \textcolor{ForestGreen}{$\uparrow$} & 68.41 & 69.24 \textcolor{ForestGreen}{$\uparrow$} & 68.94 & 70.22 \textcolor{ForestGreen}{$\uparrow$} \\
\colorbox{orange!20} {maggic (n=20)} & 67.41 & 57.13 & 60.70 & \textbf{61.48 \textcolor{ForestGreen}{$\uparrow$}} & 57.54 & 58.69 \textcolor{ForestGreen}{$\uparrow$} & 52.75 & 54.51 \textcolor{ForestGreen}{$\uparrow$} & 54.59 & 55.39 \textcolor{ForestGreen}{$\uparrow$} & 50.29 & 55.64 \textcolor{ForestGreen}{$\uparrow$} & 54.72 & 57.38 \textcolor{ForestGreen}{$\uparrow$} & 55.84 & 56.15 \textcolor{ForestGreen}{$\uparrow$} & 54.08 & 56.19 \textcolor{ForestGreen}{$\uparrow$} \\
\colorbox{orange!20} {seer (n=20)} & 87.92 & 80.67 & 84.53 & \textbf{84.82 \textcolor{ForestGreen}{$\uparrow$}} & 83.34 & 83.71 \textcolor{ForestGreen}{$\uparrow$} & 74.34 & 78.73 \textcolor{ForestGreen}{$\uparrow$} & 80.59 & 80.60 \textcolor{ForestGreen}{$\uparrow$} & 47.57 & 74.43 \textcolor{ForestGreen}{$\uparrow$} & 76.06 & 79.98 \textcolor{ForestGreen}{$\uparrow$} & 79.23 & 80.02 \textcolor{ForestGreen}{$\uparrow$} & 74.53 & 78.73 \textcolor{ForestGreen}{$\uparrow$} \\
\colorbox{green!20} {compas (n=20)} & 67.51 & 63.11 & \textbf{68.01} & 67.91 & 62.07 & 64.43 \textcolor{ForestGreen}{$\uparrow$} & 55.67 & 62.56 \textcolor{ForestGreen}{$\uparrow$} & 57.67 & 60.87 \textcolor{ForestGreen}{$\uparrow$} & 53.33 & 63.59 \textcolor{ForestGreen}{$\uparrow$} & 59.49 & 64.62 \textcolor{ForestGreen}{$\uparrow$} & 61.06 & 61.59 \textcolor{ForestGreen}{$\uparrow$} & 58.30 & 62.58 \textcolor{ForestGreen}{$\uparrow$} \\
\colorbox{green!20} {adult (n=20)} & 84.17 & 77.45 & 50.39 & 71.48 \textcolor{ForestGreen}{$\uparrow$} & 49.23 & 72.37 \textcolor{ForestGreen}{$\uparrow$} & 72.23 & 76.86 \textcolor{ForestGreen}{$\uparrow$} & 74.35 & 75.04 \textcolor{ForestGreen}{$\uparrow$} & 67.00 & 77.25 \textcolor{ForestGreen}{$\uparrow$} & 67.46 & 76.48 \textcolor{ForestGreen}{$\uparrow$} & 73.75 & 73.67 & 73.20 & 76.90 \textcolor{ForestGreen}{$\uparrow$} \\
\colorbox{green!20} {drug (n=20)} & 77.81 & 70.84 & 75.08 & \textbf{75.29 \textcolor{ForestGreen}{$\uparrow$}} & 71.68 & 72.14 \textcolor{ForestGreen}{$\uparrow$} & 68.31 & 72.65 \textcolor{ForestGreen}{$\uparrow$} & 68.12 & 69.68 \textcolor{ForestGreen}{$\uparrow$} & 58.78 & 68.89 \textcolor{ForestGreen}{$\uparrow$} & 62.13 & 67.75 \textcolor{ForestGreen}{$\uparrow$} & 70.16 & 70.16 & 66.60 & 69.18 \textcolor{ForestGreen}{$\uparrow$} \\ \hline
\colorbox{orange!20} {covid (n=40)} & 74.41 & 70.77 & 73.40 & \textbf{73.95 \textcolor{ForestGreen}{$\uparrow$}} & 70.42 & 71.93 \textcolor{ForestGreen}{$\uparrow$} & 63.63 & 68.46 \textcolor{ForestGreen}{$\uparrow$} & 70.50 & 70.44 & 56.50 & 68.68 \textcolor{ForestGreen}{$\uparrow$} & 66.41 & 70.48 \textcolor{ForestGreen}{$\uparrow$} & 68.66 & 68.44 & 61.03 & 67.35 \textcolor{ForestGreen}{$\uparrow$} \\
\colorbox{orange!20} {cutract (n=40)} & 72.23 & 69.18 & 69.87 & \textbf{71.72 \textcolor{ForestGreen}{$\uparrow$}} & 68.47 & 69.56 \textcolor{ForestGreen}{$\uparrow$} & 63.01 & 67.87 \textcolor{ForestGreen}{$\uparrow$} & 65.63 & 67.27 \textcolor{ForestGreen}{$\uparrow$} & 54.39 & 68.44 \textcolor{ForestGreen}{$\uparrow$} & 61.40 & 67.98 \textcolor{ForestGreen}{$\uparrow$} & 67.86 & 67.95 \textcolor{ForestGreen}{$\uparrow$} & 59.79 & 66.62 \textcolor{ForestGreen}{$\uparrow$} \\
\colorbox{orange!20} {maggic (n=40)} & 67.41 & 58.26 & 59.29 & \textbf{60.77 \textcolor{ForestGreen}{$\uparrow$}} & 57.50 & 59.15 \textcolor{ForestGreen}{$\uparrow$} & 55.00 & 56.78 \textcolor{ForestGreen}{$\uparrow$} & 55.24 & 56.94 \textcolor{ForestGreen}{$\uparrow$} & 48.81 & 56.64 \textcolor{ForestGreen}{$\uparrow$} & 54.68 & 58.58 \textcolor{ForestGreen}{$\uparrow$} & 57.40 & 57.44 \textcolor{ForestGreen}{$\uparrow$} & 55.04 & 57.33 \textcolor{ForestGreen}{$\uparrow$} \\
\colorbox{orange!20} {seer (n=40)} & 87.92 & 82.93 & 84.29 & \textbf{84.93 \textcolor{ForestGreen}{$\uparrow$}} & 83.46 & 84.44 \textcolor{ForestGreen}{$\uparrow$} & 80.05 & 83.67 \textcolor{ForestGreen}{$\uparrow$} & 82.59 & 81.37 & 54.93 & 81.11 \textcolor{ForestGreen}{$\uparrow$} & 79.88 & 84.36 \textcolor{ForestGreen}{$\uparrow$} & 80.79 & 82.21 \textcolor{ForestGreen}{$\uparrow$} & 78.69 & 83.62 \textcolor{ForestGreen}{$\uparrow$} \\
\colorbox{green!20} {compas (n=40)} & 67.51 & 62.34 & 67.57 & \textbf{67.85 \textcolor{ForestGreen}{$\uparrow$}} & 61.34 & 62.84 \textcolor{ForestGreen}{$\uparrow$} & 56.29 & 61.02 \textcolor{ForestGreen}{$\uparrow$} & 58.85 & 60.11 \textcolor{ForestGreen}{$\uparrow$} & 58.88 & 64.37 \textcolor{ForestGreen}{$\uparrow$} & 58.61 & 63.54 \textcolor{ForestGreen}{$\uparrow$} & 60.83 & 60.95 \textcolor{ForestGreen}{$\uparrow$} & 55.94 & 61.04 \textcolor{ForestGreen}{$\uparrow$} \\
\colorbox{green!20} {adult (n=40)} & 84.17 & 79.44 & 48.31 & 73.82 \textcolor{ForestGreen}{$\uparrow$} & 49.21 & 74.27 \textcolor{ForestGreen}{$\uparrow$} & 71.82 & 79.11 \textcolor{ForestGreen}{$\uparrow$} & 71.51 & 77.99 \textcolor{ForestGreen}{$\uparrow$} & 66.77 & 78.81 \textcolor{ForestGreen}{$\uparrow$} & 71.13 & 79.71 \textcolor{ForestGreen}{$\uparrow$} & 77.90 & 78.84 \textcolor{ForestGreen}{$\uparrow$} & 72.58 & \textbf{80.02 \textcolor{ForestGreen}{$\uparrow$}} \\
\colorbox{green!20} {drug (n=40)} & 77.81 & 71.86 & 74.30 & \textbf{75.79 \textcolor{ForestGreen}{$\uparrow$}} & 71.33 & 72.76 \textcolor{ForestGreen}{$\uparrow$} & 69.46 & 72.74 \textcolor{ForestGreen}{$\uparrow$} & 71.08 & 73.07 \textcolor{ForestGreen}{$\uparrow$} & 64.89 & 73.64 \textcolor{ForestGreen}{$\uparrow$} & 62.51 & 70.97 \textcolor{ForestGreen}{$\uparrow$} & 69.23 & 69.78 \textcolor{ForestGreen}{$\uparrow$} & 65.22 & 70.30 \textcolor{ForestGreen}{$\uparrow$} \\ \hline
\colorbox{orange!20} {covid (n=100)} & 74.41 & 71.57 & 73.77 & \textbf{74.71 \textcolor{ForestGreen}{$\uparrow$}} & 70.71 & 72.76 \textcolor{ForestGreen}{$\uparrow$} & 69.05 & 72.13 \textcolor{ForestGreen}{$\uparrow$} & 71.60 & 73.22 \textcolor{ForestGreen}{$\uparrow$} & 63.52 & 72.04 \textcolor{ForestGreen}{$\uparrow$} & 64.25 & 72.64 \textcolor{ForestGreen}{$\uparrow$} & 70.08 & 70.78 \textcolor{ForestGreen}{$\uparrow$} & 69.05 & 71.96 \textcolor{ForestGreen}{$\uparrow$} \\
\colorbox{orange!20} {cutract (n=100)} & 72.23 & 70.96 & 70.20 & \textbf{72.51 \textcolor{ForestGreen}{$\uparrow$}} & 69.97 & 71.94 \textcolor{ForestGreen}{$\uparrow$} & 67.94 & 72.42 \textcolor{ForestGreen}{$\uparrow$} & 70.53 & 71.98 \textcolor{ForestGreen}{$\uparrow$} & 55.72 & 69.14 \textcolor{ForestGreen}{$\uparrow$} & 67.59 & 72.42 \textcolor{ForestGreen}{$\uparrow$} & 68.79 & 69.68 \textcolor{ForestGreen}{$\uparrow$} & 66.89 & 71.52 \textcolor{ForestGreen}{$\uparrow$} \\
\colorbox{orange!20} {maggic (n=100)} & 67.41 & 59.65 & 58.98 & \textbf{61.32 \textcolor{ForestGreen}{$\uparrow$}} & 55.71 & 58.90 \textcolor{ForestGreen}{$\uparrow$} & 57.20 & 59.34 \textcolor{ForestGreen}{$\uparrow$} & 57.26 & 58.28 \textcolor{ForestGreen}{$\uparrow$} & 49.54 & 57.91 \textcolor{ForestGreen}{$\uparrow$} & 56.36 & 60.11 \textcolor{ForestGreen}{$\uparrow$} & 58.89 & 58.99 \textcolor{ForestGreen}{$\uparrow$} & 56.17 & 58.86 \textcolor{ForestGreen}{$\uparrow$} \\
\colorbox{orange!20} {seer (n=100)} & 87.92 & 83.95 & 84.45 & \textbf{85.37 \textcolor{ForestGreen}{$\uparrow$}} & 83.92 & 85.08 \textcolor{ForestGreen}{$\uparrow$} & 81.60 & 85.14 \textcolor{ForestGreen}{$\uparrow$} & 83.04 & 84.83 \textcolor{ForestGreen}{$\uparrow$} & 70.32 & 83.83 \textcolor{ForestGreen}{$\uparrow$} & 81.16 & 85.03 \textcolor{ForestGreen}{$\uparrow$} & 81.82 & 82.49 \textcolor{ForestGreen}{$\uparrow$} & 78.88 & 84.50 \textcolor{ForestGreen}{$\uparrow$} \\
\colorbox{green!20} {compas (n=100)} & 67.51 & 62.56 & 68.02 & \textbf{68.19 \textcolor{ForestGreen}{$\uparrow$}} & 60.10 & 62.47 \textcolor{ForestGreen}{$\uparrow$} & 60.01 & 63.73 \textcolor{ForestGreen}{$\uparrow$} & 58.32 & 61.34 \textcolor{ForestGreen}{$\uparrow$} & 59.97 & 64.19 \textcolor{ForestGreen}{$\uparrow$} & 60.02 & 64.04 \textcolor{ForestGreen}{$\uparrow$} & 61.44 & 61.73 \textcolor{ForestGreen}{$\uparrow$} & 59.97 & 62.82 \textcolor{ForestGreen}{$\uparrow$} \\
\colorbox{green!20} {adult (n=100)} & 84.17 & 81.24 & 46.09 & 74.57 \textcolor{ForestGreen}{$\uparrow$} & 47.56 & 73.97 \textcolor{ForestGreen}{$\uparrow$} & 74.29 & 80.45 \textcolor{ForestGreen}{$\uparrow$} & 75.93 & 78.22 \textcolor{ForestGreen}{$\uparrow$} & 77.09 & \textbf{81.66 \textcolor{ForestGreen}{$\uparrow$}} & 70.70 & 81.04 \textcolor{ForestGreen}{$\uparrow$} & 80.56 & 81.10 \textcolor{ForestGreen}{$\uparrow$} & 74.04 & 80.23 \textcolor{ForestGreen}{$\uparrow$} \\
\colorbox{green!20} {drug (n=100)} & 77.81 & 73.58 & 76.24 & \textbf{76.74 \textcolor{ForestGreen}{$\uparrow$}} & 69.46 & 71.05 \textcolor{ForestGreen}{$\uparrow$} & 68.19 & 73.28 \textcolor{ForestGreen}{$\uparrow$} & 72.43 & 73.79 \textcolor{ForestGreen}{$\uparrow$} & 67.26 & 75.28 \textcolor{ForestGreen}{$\uparrow$} & 62.67 & 73.12 \textcolor{ForestGreen}{$\uparrow$} & 70.90 & 71.53 \textcolor{ForestGreen}{$\uparrow$} & 68.22 & 73.59 \textcolor{ForestGreen}{$\uparrow$} \\ \hline
\colorbox{orange!20} {covid (n=200)} & 74.41 & 72.33 & 73.40 & \textbf{74.62 \textcolor{ForestGreen}{$\uparrow$}} & 70.70 & 73.12 \textcolor{ForestGreen}{$\uparrow$} & 71.07 & 73.89 \textcolor{ForestGreen}{$\uparrow$} & 72.47 & 74.44 \textcolor{ForestGreen}{$\uparrow$} & 65.55 & 73.07 \textcolor{ForestGreen}{$\uparrow$} & 65.04 & 72.90 \textcolor{ForestGreen}{$\uparrow$} & 71.68 & 71.87 \textcolor{ForestGreen}{$\uparrow$} & 67.89 & 72.38 \textcolor{ForestGreen}{$\uparrow$} \\
\colorbox{orange!20} {cutract (n=200)} & 72.23 & 71.75 & 71.39 & 73.01 \textcolor{ForestGreen}{$\uparrow$} & 70.28 & 72.39 \textcolor{ForestGreen}{$\uparrow$} & 69.28 & 72.41 \textcolor{ForestGreen}{$\uparrow$} & 71.83 & \textbf{74.03 \textcolor{ForestGreen}{$\uparrow$}} & 66.66 & 72.49 \textcolor{ForestGreen}{$\uparrow$} & 68.77 & 73.16 \textcolor{ForestGreen}{$\uparrow$} & 70.23 & 70.80 \textcolor{ForestGreen}{$\uparrow$} & 66.61 & 71.87 \textcolor{ForestGreen}{$\uparrow$} \\
\colorbox{orange!20} {maggic (n=200)} & 67.41 & 61.39 & 58.92 & \textbf{61.41 \textcolor{ForestGreen}{$\uparrow$}} & 57.33 & 60.16 \textcolor{ForestGreen}{$\uparrow$} & 58.48 & 61.33 \textcolor{ForestGreen}{$\uparrow$} & 56.26 & 57.20 \textcolor{ForestGreen}{$\uparrow$} & 50.74 & 59.60 \textcolor{ForestGreen}{$\uparrow$} & 55.95 & 60.75 \textcolor{ForestGreen}{$\uparrow$} & 60.73 & 60.78 \textcolor{ForestGreen}{$\uparrow$} & 57.18 & 60.23 \textcolor{ForestGreen}{$\uparrow$} \\
\colorbox{orange!20} {seer (n=200)} & 87.92 & 84.63 & 84.39 & 85.56 \textcolor{ForestGreen}{$\uparrow$} & 83.48 & 84.80 \textcolor{ForestGreen}{$\uparrow$} & 82.04 & 85.34 \textcolor{ForestGreen}{$\uparrow$} & 84.39 & \textbf{86.57 \textcolor{ForestGreen}{$\uparrow$}} & 82.15 & 86.03 \textcolor{ForestGreen}{$\uparrow$} & 77.73 & 85.19 \textcolor{ForestGreen}{$\uparrow$} & 83.38 & 84.15 \textcolor{ForestGreen}{$\uparrow$} & 79.71 & 85.26 \textcolor{ForestGreen}{$\uparrow$} \\
\colorbox{green!20} {compas (n=200)} & 67.51 & 63.27 & 67.02 & \textbf{68.15 \textcolor{ForestGreen}{$\uparrow$}} & 60.48 & 63.39 \textcolor{ForestGreen}{$\uparrow$} & 60.58 & 64.32 \textcolor{ForestGreen}{$\uparrow$} & 60.60 & 63.52 \textcolor{ForestGreen}{$\uparrow$} & 61.11 & 65.08 \textcolor{ForestGreen}{$\uparrow$} & 56.58 & 63.60 \textcolor{ForestGreen}{$\uparrow$} & 61.99 & 62.80 \textcolor{ForestGreen}{$\uparrow$} & 60.15 & 63.99 \textcolor{ForestGreen}{$\uparrow$} \\
\colorbox{green!20} {adult (n=200)} & 84.17 & 82.12 & 40.96 & 75.84 \textcolor{ForestGreen}{$\uparrow$} & 49.89 & 76.11 \textcolor{ForestGreen}{$\uparrow$} & 78.18 & 82.32 \textcolor{ForestGreen}{$\uparrow$} & 81.66 & 83.17 \textcolor{ForestGreen}{$\uparrow$} & 80.06 & \textbf{83.32 \textcolor{ForestGreen}{$\uparrow$}} & 74.31 & 82.64 \textcolor{ForestGreen}{$\uparrow$} & 82.26 & 82.39 \textcolor{ForestGreen}{$\uparrow$} & 75.21 & 82.02 \textcolor{ForestGreen}{$\uparrow$} \\
\colorbox{green!20} {drug (n=200)} & 77.81 & 76.10 & 75.58 & 76.06 \textcolor{ForestGreen}{$\uparrow$} & 70.66 & 72.81 \textcolor{ForestGreen}{$\uparrow$} & 71.31 & 75.98 \textcolor{ForestGreen}{$\uparrow$} & 69.61 & 71.79 \textcolor{ForestGreen}{$\uparrow$} & 72.35 & \textbf{77.41 \textcolor{ForestGreen}{$\uparrow$}} & 65.25 & 75.26 \textcolor{ForestGreen}{$\uparrow$} & 74.38 & 74.78 \textcolor{ForestGreen}{$\uparrow$} & 68.39 & 74.33 \textcolor{ForestGreen}{$\uparrow$} \\ \hline
\bottomrule
\end{tabular}}
\label{table:results_comparison}
\end{table*}
\endgroup

\textbf{Experimental setup.}
We compare \name~ (with GPT-4 \citep{openai2023gpt4} and GPT-3.5 \citep{brown2020language}) against a variety of baselines for tabular data generation and augmentation: CTGAN \citep{xu2019modeling}, TVAE \citep{xu2019modeling},  Normalizing Flows \citep{papamakarios2021normalizing}, TabDDPM \citep{kotelnikov2022tabddpm}, SMOTE \citep{chawla2002smote} and GReaT \citep{borisov2023language}, which fine-tunes an LLM.
We evaluate performance on 7 real-world datasets with different feature counts and representative of the diverse domains where \name~ can have impact. For each dataset, we vary the number of samples available in $\Dtrain$, repeating each experiment for $10$ seeds. 
 
 While we do not know the exact makeup of the pretraining data for LLMs like GPT-4, there is the possibility that open-source data might be included. This poses the risk of memorization as the primary source of performance gain. To disentangle the role of memorization, we select 4 real-world medical datasets (Maggic \citep{pocock2013predicting}, Covid \citep{covid}, SEER \citep{seer}, CUTRACT \citep{cutract}) that require an authorization process to access, hence are unlikely to form part of the LLMs training corpus. We use common open-source datasets (Adult and Drug from the UCI repository \citep{asuncion2007uci} and Compas \citep{compas}) that are highly reflective of data scarce domains. Further experimental details can be found in Appendix \ref{appx:B}.

\subsection{Overall performance: downstream utility} \label{performance}
We assess overall performance based on \emph{Utility} of the augmented data, which we evaluate in terms of AUC on the real $\Dtest$, \textcolor{TealBlue}{when using four different types of downstream models (see Appendix \ref{appx:B})}. This setup mirrors the widely adopted Train-on-synthetic-Test-on-real (TSTR) \citep{esteban2017real}. Additionally, we compare the performance to training on the small $\Dtrain$, as well as training on the large held-out $\Doracle$, the latter serving as an upper bound.
\textbf{GPT-4 + Curation has best overall performance.} Table \ref{table:results_comparison} shows the performance of the proposed \name~ (GPT-~4 and GPT-3.5) vs baselines --- both \emph{with} and \emph{without} our curation mechanism.  We find that the GPT-4 + Curation variant of \name~ outperforms baselines in almost all settings ($20/28$). Interestingly, its performance is close to or even exceeds the performance of $\Doracle$. Table \ref{table:rank} further shows that GPT-4 + Curation ranks first on average vs all generative methods. 
\newpage
\textbf{Sample size sensitivity.}
We now investigate the performance gains of \name~ as we vary the number of samples $n$ in $\Dtrain$, in Table \ref{table:results_comparison} and Table \ref{table:rank}.
Performance improvements and high ranking across datasets for \name~(GPT-4+Curation) are especially noticeable in the low-data regime (i.e. $n < 100$). In this regime,
the limited size of $\Dtrain$ severely constrains the other baseline methods. In contrast, as illustrated in Sec. \ref{sec:generate}, \name~ can leverage GPT-4's prior knowledge to extrapolate beyond the small $\Dtrain$, thereby improving downstream performance. As expected, the performance gap between \name~ and other methods decreases as the size of $\Dtrain$ grows (e.g. $n = 200$), where sufficient training data helps other generators achieve good performance. We further decouple the prior knowledge of the LLM and the number of in-context samples in Appendix \ref{app:decouple}, showing the importance of the in-context samples to guide the LLM's generation.

% \begin{wraptable}{r}{0.68\textwidth}
\begin{table}[!h]
    \centering
    \caption{\footnotesize Average rank of approaches across the different datasets and seeds. \name~ w/ GPT-4 ranks first across all $n$ and curation improves all the generative models. }
    \scalebox{0.75}{
\begin{tabular}{c|cccc}
\hline
Method & n=20 & n=40 & n=100 & n=200 \\
\hline
\name~ w/ GPT-4 & \bf 2.71 $\pm$ 1.44   & \bf 2.14 $\pm$ 1.06   & \bf 2.29 $\pm$ 1.19   & \bf 3.29 $\pm$ 1.38   \\
\hline
GPT-4 & 3.86 $\pm$ 1.73   & 4.29 $\pm$ 1.83   & 6.00 $\pm$ 1.77   & 7.57 $\pm$ 1.65   \\
\hline
\name~ w/ GPT-3.5 & 4.14 $\pm$ 0.94   & 4.14 $\pm$ 0.71   & 6.86 $\pm$ 1.24   & 7.57 $\pm$ 0.70   \\
\hline
NFLOW (curated) & 6.00 $\pm$ 1.21   & 4.71 $\pm$ 0.80   & 4.00 $\pm$ 0.57   & 4.71 $\pm$ 0.63   \\
\hline
GPT-3.5 & 6.71 $\pm$ 1.52   & 7.29 $\pm$ 1.26   & 11.57 $\pm$ 0.94   & 12.57 $\pm$ 0.57   \\
\hline
TVAE (curated) & 7.14 $\pm$ 1.17   & 7.86 $\pm$ 1.30   & 6.43 $\pm$ 0.40   & 6.71 $\pm$ 0.52   \\
\hline
SMOTE (curated) & 7.71 $\pm$ 0.33   & 8.14 $\pm$ 0.91   & 7.71 $\pm$ 1.19   & 7.43 $\pm$ 1.07   \\
\hline
SMOTE & 7.86 $\pm$ 0.55   & 9.57 $\pm$ 0.80   & 9.57 $\pm$ 1.09   & 9.00 $\pm$ 1.03   \\
\hline
TabDDPM (curated) & 8.29 $\pm$ 0.98   & 8.00 $\pm$ 0.93   & 6.00 $\pm$ 0.95   & 5.14 $\pm$ 1.68   \\
\hline
CTGAN (curated) & 8.29 $\pm$ 1.42   & 7.14 $\pm$ 0.91   & 4.14 $\pm$ 0.62   & 3.71 $\pm$ 0.39   \\
\hline
GReaT (curated) & 8.57 $\pm$ 1.50   & 6.57 $\pm$ 1.21   & 6.29 $\pm$ 1.38   & 3.57 $\pm$ 0.92   \\
\hline
TabDDPM & 10.14 $\pm$ 1.19   & 9.86 $\pm$ 1.15   & 10.00 $\pm$ 1.03   & 10.29 $\pm$ 1.02   \\
\hline
TVAE & 12.14 $\pm$ 0.89   & 14.00 $\pm$ 0.70   & 13.71 $\pm$ 0.39   & 14.43 $\pm$ 0.40   \\
\hline
NFLOW & 12.86 $\pm$ 0.47   & 14.14 $\pm$ 0.37   & 14.00 $\pm$ 0.45   & 15.29 $\pm$ 0.33   \\
\hline
CTGAN & 13.86 $\pm$ 0.68   & 13.14 $\pm$ 0.47   & 12.86 $\pm$ 0.37   & 12.00 $\pm$ 0.53   \\
\hline
GReaT & 15.71 $\pm$ 0.26   & 15.00 $\pm$ 0.53   & 14.57 $\pm$ 1.03   & 12.71 $\pm$ 0.96   \\
\hline
\end{tabular}
}
    \label{table:rank}
\end{table}

\textbf{Curation generally helps all generative models.} Our curation mechanism consistently benefits all generative models for the different $n$. It ensures that only high-quality samples are retained, which is crucial for good data augmentation and downstream performance and has been overlooked in previous works. This explains why the combination of the best generative model and curation (\name) gives the best results and highest rankings in the low-data regime (e.g. $n=20$). 
In addition to GPT-4 and GPT-3.5, we show the versatility of our proposed curation mechanism to provide benefit with other open-source LLM backbones, including Mistral-7b \cite{jiang2023mistral}, LLAMA-2 \cite{touvron2023llama} and Mixtral \cite{jiang2024mixtral} (cf. Appendix \ref{app:curate-open-source}).

\textbf{Performance benefits maintained for private and public datasets.} One may hypothesize that the strong LLM (e.g. GPT-4) performance is explained by datasets being part of the LLMs' training corpus, hence possibly being memorized. 
We show in Table \ref{table:results_comparison} that it is unlikely, as we retain strong performance for both \colorbox{green!20} {open-source datasets}, as well as \colorbox{orange!20}{private medical datasets} which require authorization processes for access and are unlikely to be part of the LLM pretraining dataset.

\textit{Remark on ICL versus fine-tuning.} Our results in Table \ref{table:results_comparison} and Table \ref{table:rank} indicate that ICL is better than fine-tuning (GReaT baseline) in the low-data regime. This highlights the difficulty of fine-tuning in this regime, where it is easy to overfit to $\Dtrain$.  As we increase the number of samples, this baseline, coupled with curation, improves to the level of \name ~ (GPT-4).

\subsection{Hardness: a proxy signal to flag poor quality synthetic datasets}\label{reliability}

 Having a systematic way to assess datasets generated by LLMs like GPT-4 is important because their black-box nature provides little control on their generation quality. This contrasts conventional generators for which training loss is an exploitable signal. Hence, we ask: could we have a signal to identify a potential problematic dataset generated by an LLM without an exhaustive manual review? For example, GPT-4 produced low-quality synthetic data for the Adult dataset (across the different sample sizes) resulting in poor downstream performance. While curation improves it, downstream performance is still suboptimal. 

 \begin{figure}[!h]

\centering
\includegraphics[width = \linewidth]{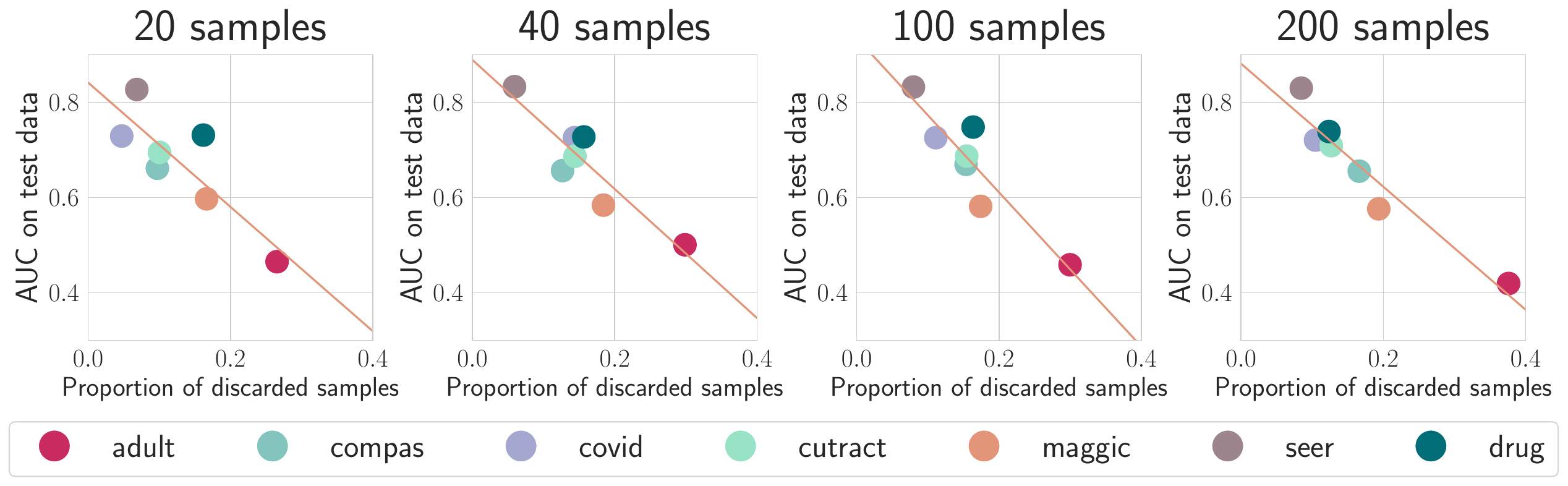}
\caption{The proportion of discarded samples $\Dsyn$ is a proxy for test performance. This negative linear relationship where each point is a synthetic dataset generated by GPT-4 (e.g. Adult, Covid, Compas) allows us to flag datasets that will lead to unreliable downstream performance.}
\label{reliability-diagram}
     \rule{\linewidth}{.5pt}
\end{figure}
 
 Addressing this question is important, since datasets are rarely created by the ML model builder in real-world ML workflows, but rather by specialist data teams or data owners \citep{gebru2021datasheets, sambasivan2021everyone,goncalves2020generation}.
Thus, having a signal to preemptively flag a potentially suboptimal generated dataset spares investment in both storing the subpar data and/or training a model likely to underperform on real data.

To address this, we posit that $\Dsyn$ should intuitively be considered imperfect if curation discards many of its samples, since the number of discarded samples measures the quality of samples with respect to the small but gold-standard $\Dtrain$. 

Hence, we investigate the relationship between test performance (AUC) and the proportion of samples discarded by curation. Fig. \ref{reliability-diagram}, where each point is a synthetic dataset generated by GPT-4 (e.g. Adult, Compas), shows a strong negative linear relationship between these two quantities. This holds across the different $n$ with slopes fairly stable around $-1.4$. 
This relationship corroborates the poor quality of the dataset generated by GPT-4 on the Adult dataset, providing a useful proxy that $\Dsyn$ is unlikely to lead to good downstream performance.

\section{Discussion} 

\vspace{2mm}
 
We introduce \name, an approach for data augmentation in the low-data setting. \name~ exploits the prior knowledge of LLMs along with our curation mechanism for improved downstream performance. 

\vspace{2mm}

As empirically shown, \name~ outperforms traditional generative models---most noticeably on underrepresented subgroups, for which data augmentation is of utmost importance. \name~ is grounded in the ICL capability of LLMs. Further improvements may be achieved through different tuning and prompting of the LLM, as shown in different domains \citep{meng2023tuning, liu2023pre}. Improving LLM tuning and prompting is beyond the scope of our work, but we regard this as a promising avenue for future work. 
\vspace{2mm}

While the key contribution of this work is the curation of LLM outputs, overall downstream performance gains are still fundamentally tied to the LLM backbone. Specifically, a practical consideration is that using less parameterized LLMs leads to poorer uncurated data. That said, our curation mechanism naturally addresses this aspect and improves downstream performance (see Appendix \ref{app:curate-open-source}).

Finally, while \name~ addresses the data logjam issue, increasing access to ML across regions, domains and societies is also about more than just technology. We believe broader engagement and discussion with various stakeholders is crucial to responsibly expand ML access, thereby realizing the benefits of ML in an equitable way.

\vspace{2mm}

\section*{Impact statement}

Data scarcity and computational limitations are deterrents for developing ML. These challenges should inspire cutting-edge ML research \citep{de2018machine}. We believe \name~ takes a step in this direction toward improving the use of ML in low-data settings, across \textit{society} (e.g. underrepresented subgroups \citep{suresh2021framework}), \textit{domains} (e.g. healthcare \citep{alami2020artificial,owoyemi2020artificial}) and \textit{regions} (e.g. LMICs). 

 However, with that in mind, LLMs may make errors and may reflect or exacerbate societal biases that are present in their data \citep{li2023ethics}. Though the curation in {\name} improves synthetic data quality, it does not directly aim to remove biases. The quality and fairness of generated data should always be evaluated. We believe broader engagement and discussion with various stakeholders is required before methods like {\name} should be applied to real-world sensitive settings like healthcare and finance, as well as more research into LLM bias and potential mitigation strategies. We provide a more detailed discussion in \cref{subsec:bias}.

Additionally, in this work, we evaluate \name~
using multiple real-world datasets. The private datasets are \emph{de-identified} and used in accordance with the guidance of
the respective data providers. We follow recommendations to use the Azure OpenAI service when using GPT-4 and GPT-3.5 models, where via the agreement we ensure the medical data is not sent for human review or stored, hence respecting the guidelines given by the dataset providers.

\section*{Acknowledgments} The authors are grateful to Fergus Imrie, Andrew Rashbass and the anonymous
ICML reviewers for their useful comments and feedback. Nabeel Seedat is supported by the Cystic Fibrosis Trust, Nicolas Huynh by Illumina, and Boris van Breugel by the Office of Naval
Research UK. This work was supported by Microsoft's Accelerate Foundation Models Academic Research initiative.

\bibliographystyle{icml2024}
\bibliography{references} 

\clearpage
\onecolumn
\appendix

\addcontentsline{toc}{section}{Appendix}
\part{Appendix: Curated LLM: Synergy of LLMs and Data Curation for tabular augmentation in low-data regimes}

\mtcsetdepth{parttoc}{3} 
\parttoc

\newpage

\section{Extended Related Work}\label{appendix:extended}
This paper primarily engages with the work on data augmentation when we have limited data, where our primary goal is synthetic data generation to augment the dataset. Generating synthetic datasets not only helps improve downstream performance, but it is also a flexible solution as it doesn't tie the data consumer to any particular downstream model. That said, beyond the major difference of synthetic data generation, for completeness we contrast our setting of learning from limited data with other seemingly similar settings and highlight their differences.

\textbf{Contrasting learning w/ limited data vs other settings.} The challenge of learning from limited data, while seemingly related to several other learning paradigms, presents distinct differences and unique intricacies that warrant dedicated study.

\textit{Transfer learning} \citep{pan2009survey}, \textit{domain adaptation} \citep{farahani2021brief}, and \textit{few-shot learning} \citep{wang2020generalizing} employ additional data resources or rely on specific task-related assumptions to improve learning performance. These methods exploit large labeled data from a source domain, unlabeled data in a target domain, or leverage knowledge from related tasks respectively.  \textcolor{TealBlue}{For example, \cite{levin2022transfer} and \cite{jin2023benchmarking} use models trained on labeled data from a source domain, while \cite{ruiz2023enabling} and \cite{margeloiu2022graph} leverage knowledge-graphs.}
This is in contrast to our setting, considered of learning with limited data, which must function with whatever scarce labeled data it has, without making any assumptions about the availability of additional data or tasks.

\textbf{Detailed contrast between \name~ and transfer learning / meta-learning / few-shot learning.}

In addition to the above, we emphasize below three dimensions along which \name~ differs from the transfer learning and meta-learning literature (which permit to do few-shot learning). Specifically, we highlight three specific dimensions which explain why transfer learning and meta-learning generally cannot apply to the setting considered in \name~. We provide empirical evidence on why the following dimensions are important (notably the point on the choice of downstream backbone). Specifically, we compare \name~ with TabPFN \cite{hollmann2022tabpfn}, a few-shot learning method designed for small tabular problems (see \cref{appendix:tabpfn}).

\begin{enumerate}
    \item \textit{Access to external datasets}: our problem setting in \name~ assumes access to a single small training set $D_{\mathrm{train}}$, without access to any external/additional datasets. This mirrors the unique characteristics of low-to-middle income countries (LMICs), where data scarcity may be pervasive, hence making it unrealistic to assume that external datasets are available to the practitioner. In contrast, transfer learning and meta-learning make more stringent assumptions on the data requirements. (i) \textit{Transfer learning} (Definition 3 in \cite{zhuang2020comprehensive}): one typically assumes access to at least one additional source dataset $D_{\mathrm{source}}$, usually bigger than $D_{\mathrm{train}}$. This external $D_{\mathrm{source}}$ can then be used to pretrain a model, which is adapted using $D_{\mathrm{train}}$. (ii) \textit{Meta-learning} \cite{hospedales2021meta}: one assumes access to a set of $m$ tasks, which define a set of source datasets 
 $\{D_{\mathrm{source}}^{(i)} \mid i \in [m]\}$. It is also worth noting that a common assumption in meta-learning is that the source datasets and $D_{\mathrm{train}}$ share the same feature space, restricting its applicability.
    To summarize, both these learning paradigms often rely on external data, an assumption not required in \name~. We acknowledge that \name~ can be seen as a form of transfer, in its broad definition, since it uses prior knowledge, with the LLM. However, a key point is that it does not require external/additional datasets.

\item \textit{Flexibility of the backbone model}: since \name~ is a data augmentation method for tabular data, it enables the practitioner to use any downstream model backbone, such as neural networks or tree-based methods (XGBoost, Random forest, CatBoost). This flexibility is important, as it allows the practitioner to choose the best-suited model for the task at hand. For example, one can use any tree-based method with \name~, which is appealing given that tree-based methods are often preferred over neural networks for tabular data \cite{grinsztajn2022tree}, \cite{shwartz2022tabular}. In contrast, most approaches to transfer learning and meta-learning are not flexible as they traditionally require neural networks as backbone models. This stems from their methodology, where fine-tuning a pretrained model on $D_{\mathrm{train}}$ is a prevailing approach \cite{finn2017model}.
\item   \textit{Ease of use}: \name~ is distinct in that it is easy to use, without the need for costly or complex operations, such as fine-tuning large pretrained models or using them at inference time, which may be impossible in LMICs, due to the cost associated to these operations.
\end{enumerate}

\textit{Active learning} \citep{settles2009active} and \textit{semi-supervised learning} \citep{vanEngelen2019ASO,chapelle} also operate under the premise of having access to plentiful unlabeled data and the capacity to interactively query labels. However, in our setting, considered learning with limited data does not inherently assume such capabilities, focusing instead on limited labeled data only. Furthermore, active learning primarily focuses on the iterative process of selecting data samples that, when labeled, are expected to most significantly improve the model's performance. This selection is typically based on criteria such as uncertainty sampling which focuses on \textbf{epistemic uncertainty} \citep{mussmann2018relationship, houlsby2011bayesian, kirsch2019batchbald, nguyen2022measure}. The primary objective is to minimize labeling effort while maximizing the model's learning efficiency. Additionally, active learning would aims to label instances based on epistemic uncertainty where the model struggles to make accurate predictions, yet the samples themselves are correct.
In contrast, {\name} leverage training dynamics based on \textbf{aleatoric uncertainty} and confidence and is designed to discard samples that might jeopardize the downstream accuracy. These samples can be considered to have inherent issues or are erroneous, such as being "mislabeled".
To summarize, in active learning, epistemic uncertainty is used to identify data points that, if labeled, would yield the most significant insights for model training. In our approach, they serve to identify and exclude/filter data points that could potentially deteriorate the model's performance.

\textit{Self-supervised learning} \citep{liu2021self} leverages large amounts of unlabeled data to learn useful representations for downstream tasks. However, in our setting, considered learning with limited data does not inherently assume such access to vast amounts of unlabeled data.

\textbf{Data-centric AI.} Ensuring high data quality is a critical but often overlooked problem in ML, where the focus is
optimizing models \citep{sambasivan2021everyone}. Even when it is considered, the process of assessing
datasets is adhoc or artisanal \citep{seedat2022dc}. However, the recent push of data-centric AI \citep{liang2022advances, polyzotis2021can,zha2023data, seedat2023dissecting} aims to develop systematic tools to curate existing datasets.  
Our work contributes to this nascent body of work \citep{seedat2023triage} – presenting \name, which, to the best of our knowledge, is the first systematic data-centric framework looking at how we can tailor synthetic datasets  (rather than real datasets) to downstream task use with data curation.

\textbf{Why Data Augmentation?}
Data augmentation is a flexible approach to address the low-data regime. 
An alternative might be to resort to a pretrained black-box model for classification, which could be for example via in-context learning for classification \citep{dong2022survey}. However, such a solution is inadequate for several reasons, many of which would prevent real-world utility (e.g. in LMICs): 

$\blacktriangleright$ \emph{Not economical over the long term:} While using an LLM like GPT for classification may seem attractive due to its few-shot capabilities, it is likely not economically viable in real-world settings, especially in LMICs. The reason is classifying each sample will incur a cost to call the LLM, hence scales linearly with the number of test samples. Over time, the cumulative cost of these calls will surpass the once-off fixed cost associated with generating data. With data augmentation, once the dataset is augmented, there are no additional deployment time costs associated with the LLM. Indeed, the downstream models e.g. a random forest or XGBoost have negligible inference costs.
 
$\blacktriangleright$ \emph{Control, interpretability and auditability:} Relying on a large, pre-trained LLM as a black-box classifier raises several concerns. (1) we have no control over our downstream classifier and its architecture, (2) lack of interpretability and auditability of the LLM when issuing predictions. In contrast, training a downstream model on augmented data maintains the ability to understand and explain  how the model is making decisions (e.g. feature importance). This is especially crucial in contexts where accountability, transparency, and validation of machine learning processes are paramount.

$\blacktriangleright$ \emph{Independence and self-sufficiency:} Relying on third-party services for continuous classification means being dependent on their availability, pricing models, and potential changes in the LLM version. By augmenting data and training a downstream classifier on the augmented dataset, we ensure that there is no external dependencies such as increasing costs or reduced performance with LLM version updates. 

$\blacktriangleright$  \emph{Hardware and financial constraints:}  Even if we opt for an open-source LLM (e.g. Falcon \citep{penedo2023refinedweb} or LLaMA-2 \citep{touvron2023llama}), deploying and running it locally demands significant computational resources. Typically, these models require GPUs with high amounts of VRAM for optimal performance (e.g. needing around 40 GB hencing requiring an A100 GPU for Falcon-40b and LLaMA-2 65B). Such high-end GPUs are expensive, and are likely to be inaccessible in a LMIC setting. Furthermore, renting hardware by the hour can quickly become prohibitively expensive. Data augmentation, on the other hand, can often be performed on modest hardware, and once the augmented dataset is created, many classifiers can be trained without the need for high-end GPUs, making the entire process more financially accessible.

In conclusion, while large language models offer vast knowledge, for low-data settings in low-income countries, data augmentation provides a more cost-effective, controllable, and interpretable solution for building robust classifiers.

\clearpage

\newpage

\section{Experimental Details} \label{appx:B}

We provide details on our datasets used, as well as, other experimental specifics including: generation, curation, downstream model, prompt template.

\subsection{Datasets}
We summarize the different datasets we use in this paper in Table \ref{tab:dataset}. The datasets vary in number of samples, number of features and domain. 

\begin{table}[h]
\centering
\caption{Summary of the datasets used. * Denotes private/proprietary datasets.}
\scalebox{1}{
\begin{tabular}{l|lll}
\toprule

Name &  $n$ samples & $n$ features & Domain \\ 
\midrule
Adult Income \citep{asuncion2007uci} & 30k & 12 & Finance \\ 
Compas \citep{compas}  & 5k & 13 & Criminal justice \\ 
*Covid-19 \citep{covid} & 7k & 29 & Healthcare/Medicine \\
*CUTRACT Prostate \citep{cutract} & 2k & 12 & Healthcare/Medicine \\ 
Drug \citep{fehrman2017five}  & 2k & 27 & Healthcare/Medicine \\  
*MAGGIC \citep{pocock2013predicting}  & 41k & 29 & Healthcare/Medicine \\ 
*SEER Prostate \citep{seer} & 20k & 12 & Healthcare/Medicine \\ 
\bottomrule
\end{tabular}}
\vspace{.5cm}
\label{tab:dataset}
\end{table}

The private datasets are de-identified
and used in accordance with the guidance of the respective data providers. We follow recommendations to use the
Azure OpenAI service when using GPT-4 and GPT-3.5 models, where via the agreement we ensure the medical data is
not sent for human review or stored, hence respecting the
guidelines given by the dataset providers.

We detail the dataset splits used in Sec. \ref{performance}. For each dataset and number of samples $n \in \{ 20, 40, 100, 200\}$, we sample a training set $\Dtrain$ such that $\vert \Dtrain \vert = n$, and each target class has the same number of samples.  We then split the remaining samples into two non-overlapping datasets, $\Doracle$ and $\Dtest$, which have the same cardinality. This procedure is repeated $n_{\mathrm{seed}} = 10$ times, thus leading to different training and test sets. Note that the different generative models use the same $\Dtrain$ and $\Dtest$ for a given seed.

\textcolor{TealBlue}{
\textbf{Motivation for the choice of datasets.}}
\begin{enumerate}
    \item \textcolor{TealBlue}{\textbf{Open-source:} Adult, Drug and Compas are widely used open-source datasets used in the tabular data literature. Adult and Drug are both UCI datasets that have been used in many papers, while Compas is part of OpenML \cite{OpenML2013}. Our reason for selecting them is that, despite them being open-source, they are highly reflective of domains in which we might be unable to collect many samples --- hence in reality would often be in a low-data regime.}
    \item \textcolor{TealBlue}{\textbf{Private datasets:} We wanted to disentangle the possible role of memorization in the strong performance of the LLM. To ensure the datasets are not in the LLMs training corpus, we selected 4 private medical datasets that need an authorization process to access. Hence, these datasets would not be part of the LLMs training corpus given their proprietary nature and hence would be unseen to the LLM. While the private and unseen aspect was the main motivation, we also wish to highlight that these are real-world medical datasets. Consequently, this allows us to test a highly realistic problem setting.}
\end{enumerate}

\subsection{Data generation.}

\textbf{GPT-4 and GPT-3.5} We access GPT-4 \citep{openai2023gpt4} and GPT-3.5-Turbo \citep{brown2020language} through the API. We use a temperature of $0.9$.

\textbf{GReaT.} GReaT \cite{borisov2023language} is a generative model which fine-tunes an LLM based on a training set.
We use the implementation provided by authors.

\textbf{Generative model based approaches.}
For the other baselines used in \ref{performance}, we use the library SynthCity \citep{https://doi.org/10.48550/arxiv.2301.07573}, using the defaults. We detail each next.

\begin{itemize}
    \item TVAE: this is a conditional Variational Auto Encoder (VAE) for tabular data and is based on \citet{xu2019modeling}
    \item CTGAN: A conditional generative adversarial network which can handle tabular data and is based on \citet{xu2019modeling}
    \item NFLOW: Normalizing Flows are generative models which produce tractable distributions where both sampling and density evaluation can be efficient and exact.
    \item TabDDPM:  A diffusion model that can be universally applied to any tabular dataset, handles any type of feature and is based on \citet{kotelnikov2022tabddpm}
\end{itemize}

\textbf{Traditional Data Augmentation.} We use
SMOTE \citep{chawla2002smote} which augments data by considering nearest neighbors and performing linear interpolations. We use the implementation provided by \cite{JMLR:v18:16-365}, and set the number of neighbors $k$ to $5$.

\subsection{Data curation}

\paragraph{Learning dynamics computation}
We train an XGBoost with $100$ estimators on $\Dtrain$. We then compute predictive confidence and aleatoric uncertainty for the samples in $\Dsyn$.
\textcolor{TealBlue}{The motivation for the choice of an XGBoost backbone is that we cannot expect good performance by choosing “any” curation model, but rather we require a curation model with enough capacity and generalization properties — where boosting methods like XGBoost used in our work have shown to achieve best performance on tabular data. This leads to our guideline for the curation step: the model used for curation should be \textbf{at least as flexible} as the model that the practitioner intends to use for the downstream task.}

\paragraph{Learning dynamics thresholds}
 Recall that \name~ has two thresholds $\tau_{\mathrm{conf}}$ and $\tau_{\mathrm{al}}$ on the predictive confidence and aleatoric uncertainty respectively, as defined in \ref{sec:curation}. We set $\tau_{\mathrm{conf}} = 0.2$, in order to select high confidence samples. We adopt an adaptive threshold for $\tau_{\mathrm{al}}$  based on the dataset, such that $\tau_{\mathrm{al}} = 0.75 \cdot (\max(v_{al}(\Dsyn)) - \min(v_{al}(\Dsyn)))$. Note that by definition $v_{al}(\Dsyn)$ is bounded between $0$ and $0.25$.

\paragraph{Example of learning dynamics}
We include examples of learning dynamics computed for $20$ samples in Fig. \ref{fig:LD}.

\begin{figure}[!h]
    \centering
    \includegraphics[scale=0.4]{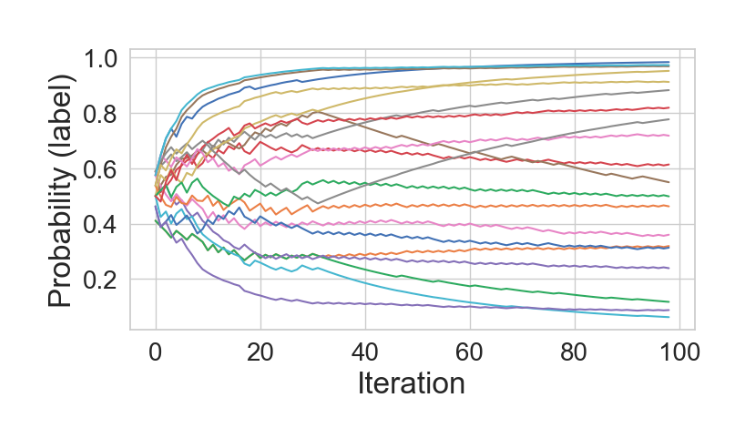}
    \caption{Learning dynamics computed for $20$ samples}
    \label{fig:LD}
\end{figure}

\subsection{Downstream task}
We compute downstream performance in Sec. \ref{performance} using four different downstream models: XGBoost, Random Forest, Decision tree, and Logistic Regression.

\newpage
\subsection{Prompt example} \label{app:prompt}
We include the template of the prompts used throughout the paper. We show how we include (1) in-context examples (demonstrations), (2) contextual information including dataset background and feature information and (3) the instruction.

\begin{lstlisting}[language=Python, caption=Template of the prompt]
    System role: 'You are a tabular synthetic data generation model.'

    You are a synthetic data generator. 
    Your goal is to produce data which mirrors \
    the given examples in causal structure and feature and label distributions \
    but also produce as diverse samples as possible. 

    I will give you real examples first.

    Context: Leverage your medical knowledge about covid and Brazil to generate 1000 realistic but diverse samples. 

    example data: {data}

    The output should be a markdown code snippet formatted in the following schema:

    "Sex_male": string  // feature column
    "Age": string  // feature column
    "Age_40": string  // feature column
    "Age_40_50": string  // feature column
    "Age_50_60": string  // feature column
    "Age_60_70": string  // feature column
    "Age_70": string  // feature column
    "Fever": string  // feature column
    "Cough": string  // feature column
    "Sore_throat": string  // feature column
    "Shortness_of_breath": string  // feature column
    "Respiratory_discomfort": string  // feature column
    "SPO2": string  // feature column
    "Dihareea": string  // feature column
    "Vomitting": string  // feature column
    "Cardiovascular": string  // feature column
    "Asthma": string  // feature column
    "Diabetis": string  // feature column
    "Pulmonary": string  // feature column
    "Immunosuppresion": string  // feature column
    "Obesity": string  // feature column
    "Liver": string  // feature column
    "Neurologic": string  // feature column
    "Renal": string  // feature column
    "Branca": string  // feature column
    "Preta": string  // feature column
    "Amarela": string  // feature column
    "Parda": string  // feature column
    "Indigena": string  // feature column
    "is_dead": string  // label if patient dead or not, is_dead

    DO NOT COPY THE EXAMPLES but generate realistic but new and diverse samples which have the correct label conditioned on the features.
        
\end{lstlisting}

\clearpage

\section{Additional Results}

\subsection{Decoupling prior knowledge and data model} \label{app:decouple}

Two components can be attributed to the good performances of {\name}: the background knowledge of the LLM, and its capacity to build a strong data model. 
In this subsection, we provide insights to understand the effect of the LLM's background knowledge (e.g. prior).
\begin{wraptable}{r}{0.5\textwidth}    \centering
    \caption{\textcolor{TealBlue}{Downstream accuracy when varying the number of in-context samples in the prompt to generate the augmented datasets.}}
    \begin{tabular}{c|c}
  \hline
  In-context samples & Downstream accuracy \\
  \hline
  $n=1$ (Prior) & $70.20\pm1.60$ \\
  \hline
  $n=20$ & $73.87\pm 0.50$ \\
  \hline
  $n=40$ & $73.95\pm0.67$ \\
  \hline
  $n=100$ & $74.71\pm0.34$ \\
  \hline
  $D_{\mathrm{oracle}}$ & $74.6\pm 0.15$ \\
  \hline
\end{tabular}
    \label{tab:background_exp}
\end{wraptable}
We considered the Covid dataset (private medical dataset, to avoid memorization issues) and generated data with GPT-4 (same as Section \ref{sec:generate}). We ablate the prompt used in our work (detailed in Appendix \ref{app:prompt}), and solely provide one in-context example in the prompt, in order to give the LLM the minimal amount of information about the desired structure of the dataset. 
This forces the LLM to rely on its own prior (background knowledge), and removes the effect of in-context examples which could be used to build a data model. 
We report the results for the prior and {\name} in Table \ref{tab:background_exp}.

From these results, we conclude that the LLM prior permits to obtain good downstream performance, but is outperformed by $\mathcal{D_{\mathrm{oracle}}}$ by a margin of $4.4\%$. Hence, we cannot solely rely on the prior. Furthermore, downstream performance increases as the number of in-context samples increases. This shows it is important to include in-context samples if we wish to obtain downstream performance close to $\mathcal{D}_{\mathrm{oracle}}$, as the LLM can build a good data model.
This implies that while the LLM uses background knowledge of similar datasets, it still needs in-context samples to refine its prior by creating a good data model.

We then quantify and visualize the strength of the prior, by studying how much the LLMs output distribution adapts to the in-context samples provided. We evaluate data generated by the prior of the LLM ($n=1$), and for $n=20,40,100$ on the Covid dataset. \textcolor{TealBlue}{In particular, we observe in Figure \ref{fig:subgroups_tsne} that there is a region in the oracle data which is not captured by the LLM’s prior output (the left part of the leftmost blob, circled in blue in Figure \ref{fig:subgroups_tsne}). However, as the number of in-context real examples increases in the prompt of the LLM, we observe that this steers the LLM to generate data which covers this region. 
This region is associated to the subgroup of people older than $87$ years old, and having many severe comorbidities (e.g. Diabetes, Cardiovascular diseases) and many respiratory symptoms. This subgroup, in the Oracle dataset, represents less than $3.5\%$ of the data, and is completely ignored by the GPT-4 prior. In particular, the prior defaults to more typical patients in the range $70$-$80$ years old.
On the contrary,  as $n$ increases, the LLM is guided by the in-context samples and generates samples from this subgroup, which are "rarer" or different from the general population. }

This shows that the LLM captures the distinct features of this particular region, and is not overwhelmed by the prior. Instead, the data in the form of in-context samples adapts it, and aligns the augmented dataset with the ground-truth distribution.

\begin{figure}[!h]
    \centering
    \includegraphics[width=0.75\linewidth]{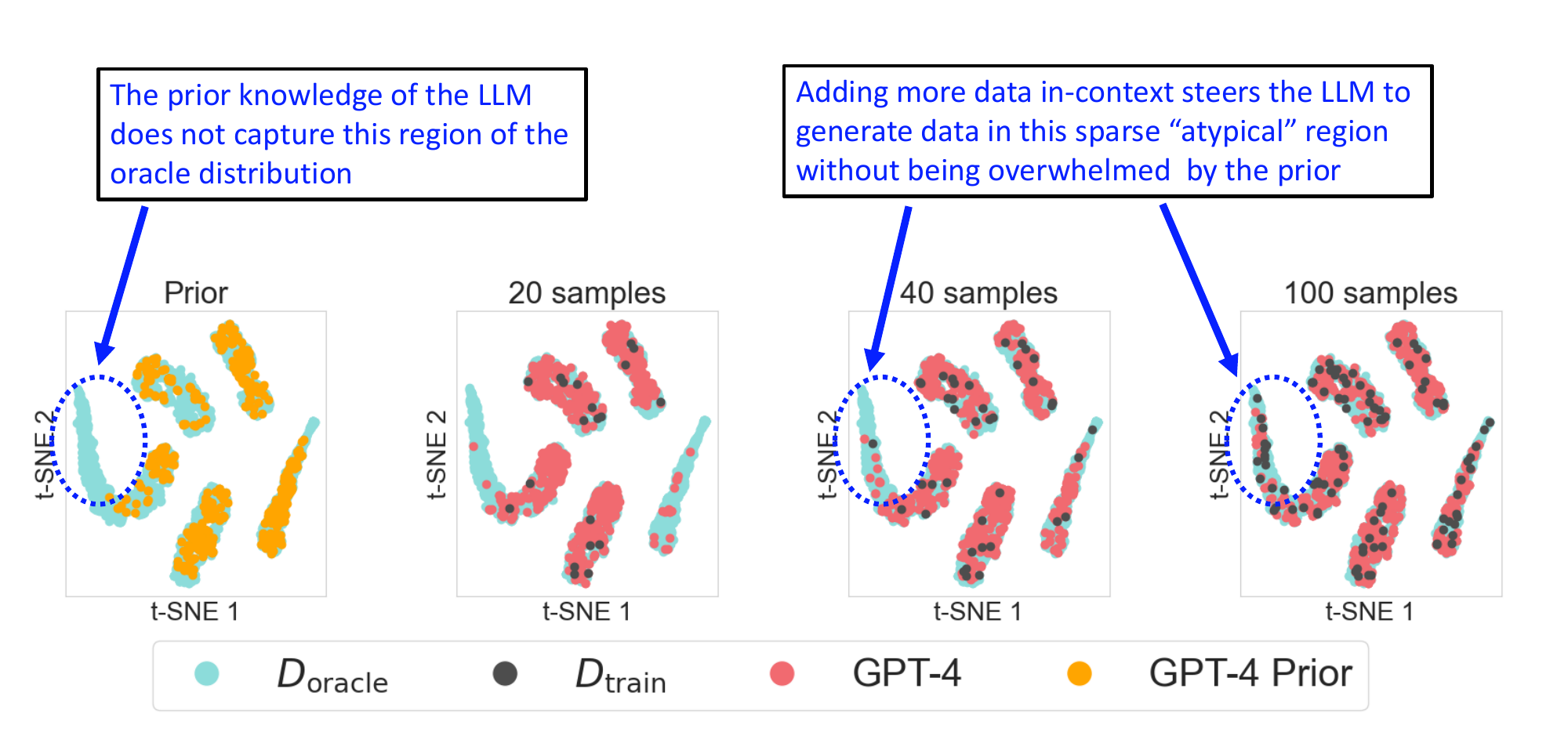}
    \caption{\textcolor{TealBlue}{The data generated by the LLM captures the distinct features in "atypical" regions of the Oracle manifold, as in-context samples are  added to the prompt. This shows that it is flexible enough to adapt its prior knowledge to the nuances of the data. The group encircled in blue represents patients who are $>88$ years old, representing around $3.5\%$ of the Oracle. This illustrates the added in-context samples can successfully guide the LLM to generate these rare samples. }}
    \label{fig:subgroups_tsne}
\end{figure}

\newpage
\subsection{Curation mechanism also improves open-source LLM generated data}\label{app:curate-open-source}
In this subsection, we demonstrate the wide applicability of the \name~ framework. Different factors may affect the choice of the LLM backbone, such as operational costs, and the desired parameterization of the LLM (for quality of generation). In addition to GPT-4 and GPT-3.5, we use the open source models Mistral-7b \cite{jiang2023mistral} and LLaMa-13b (4-bit quantized) \cite{touvron2023llama}, LLaMa-70b \cite{touvron2023llama} and Mixtral-8x7b \cite{jiang2024mixtral} to generate augmented datasets \footnote{For LLaMa-70b and Mixtral-8x7b, we only run open-source datasets due to data sharing restrictions with the endpoints for those two models.}. We then compute the downstream performance when training a model on both the uncurated and curated data. As can be seen in \cref{table:open-source-compare}, downstream performance with uncurated is lower for these open-source models compared to GPT-4 --- which is expected given their significantly smaller size (i.e. parameter count). 
However, the curation mechanism, which is the key contribution of \name, almost always improves downstream performance for all LLM backbones investigated.  Overall, this demonstrates the versatility and wide applicability of \name~ for tabular data augmentation in low-data regimes. 

\textbf{Practical tip:} The size of the LLM plays a role: the larger models (with more parameters) outperform those with fewer parameters (e.g. GPT-4 vs Mixtral-8x7b vs Mistral-7b). Hence, while curation helps improve all LLM-generated data, ideally the best LLM possible should be used.

% \begingroup
% \setlength{\tabcolsep}{1.5pt}
% \renewcommand{\arraystretch}{1}
\begin{table*}[!h]
\caption{AUC averaged over 4 downstream models on $\Dtest$ for GPT-4, GPT-3.5, Mistral-7b, LLAMA-13b, LLAMA-70b and Mixtral-8x7b (Curated and Uncurated)}
\vspace{2mm}
\centering
\scalebox{0.95}{
\begin{tabular}{l|cc|cc|cc|cc|cc|cc}
\toprule
& \multicolumn{4}{c}{Open AI} & \multicolumn{8}{c}{Open source}  \\
\cmidrule(lr){2-5}
\cmidrule(lr){6-13} 
Dataset & \multicolumn{2}{c}{GPT-4} & \multicolumn{2}{c}{GPT-3.5} & \multicolumn{2}{c}{Mistral-7b} & \multicolumn{2}{c}{LLAMA-13b} & \multicolumn{2}{c}{LLAMA-70b} & \multicolumn{2}{c}{Mixtral} \\
\cmidrule(lr){2-3}\cmidrule(lr){4-5}\cmidrule(lr){6-7}\cmidrule(lr){8-9}\cmidrule(lr){10-11}\cmidrule(lr){12-13}
& Uncur. & Cur. & Uncur. & Cur.  & Uncur. & Cur.  & Uncur. & Cur. & Uncur. & Cur. & Uncur. & Cur. \\
\hline
\colorbox{orange!20}{covid (n=20)} & 73.78 & \textbf{73.87 \textcolor{ForestGreen}{$\uparrow$}} & 69.85 & \textbf{71.41 \textcolor{ForestGreen}{$\uparrow$}} & 71.47 & \textbf{72.80 \textcolor{ForestGreen}{$\uparrow$}} & 63.25 & \textbf{67.24 \textcolor{ForestGreen}{$\uparrow$}} & N.A. & N.A. & N.A. & N.A. \\
\colorbox{orange!20}{cutract (n=20)} & 71.15 & \textbf{72.50 \textcolor{ForestGreen}{$\uparrow$}} & 69.97 & \textbf{71.54 \textcolor{ForestGreen}{$\uparrow$}} & 69.60 & \textbf{71.34 \textcolor{ForestGreen}{$\uparrow$}} & 67.84 & 66.71 & N.A. & N.A. & N.A. & N.A. \\
\colorbox{orange!20}{maggic (n=20)} & 60.70 & \textbf{61.48 \textcolor{ForestGreen}{$\uparrow$}} & 57.54 & \textbf{58.69 \textcolor{ForestGreen}{$\uparrow$}} & 53.64 & 52.06 & 53.30 & \textbf{53.96 \textcolor{ForestGreen}{$\uparrow$}} & N.A. & N.A. & N.A. & N.A. \\
\colorbox{orange!20}{seer (n=20)} & 84.53 & \textbf{84.82 \textcolor{ForestGreen}{$\uparrow$}} & 83.34 & \textbf{83.71 \textcolor{ForestGreen}{$\uparrow$}} & 83.60 & \textbf{85.18 \textcolor{ForestGreen}{$\uparrow$}} & 80.94 & \textbf{82.84 \textcolor{ForestGreen}{$\uparrow$}} & N.A. & N.A. & N.A. & N.A. \\
\colorbox{green!20}{compas (n=20)} & 68.01 & 67.91 & 62.07 & \textbf{64.43 \textcolor{ForestGreen}{$\uparrow$}} & 56.26 & \textbf{59.95 \textcolor{ForestGreen}{$\uparrow$}} & 60.08 & \textbf{60.34 \textcolor{ForestGreen}{$\uparrow$}} & 56.10	& \textbf{60.68 \textcolor{ForestGreen}{$\uparrow$}} & 54.34 &	\textbf{61.66 \textcolor{ForestGreen}{$\uparrow$}}	 \\
\colorbox{green!20}{adult (n=20)} & 50.39 & \textbf{71.48 \textcolor{ForestGreen}{$\uparrow$}} & 49.23 & \textbf{72.37 \textcolor{ForestGreen}{$\uparrow$}} & 47.68 & \textbf{65.84 \textcolor{ForestGreen}{$\uparrow$}} & 48.82 & \textbf{66.00 \textcolor{ForestGreen}{$\uparrow$}} & 51.96	& \textbf{68.40 \textcolor{ForestGreen}{$\uparrow$}} & 57.78 &	\textbf{76.84 \textcolor{ForestGreen}{$\uparrow$}}	 \\
\colorbox{green!20}{drug (n=20)} & 75.08 & \textbf{75.29 \textcolor{ForestGreen}{$\uparrow$}} & 71.68 & \textbf{72.14 \textcolor{ForestGreen}{$\uparrow$}} & 74.14 & \textbf{75.21 \textcolor{ForestGreen}{$\uparrow$}} & 67.96 & 67.87 & 66.77	& \textbf{69.12 \textcolor{ForestGreen}{$\uparrow$}} & 72.45	& 71.63	 \\ \hline\hline
\colorbox{orange!20}{covid (n=40)} & 73.40 & \textbf{73.95 \textcolor{ForestGreen}{$\uparrow$}} & 70.42 & \textbf{71.93 \textcolor{ForestGreen}{$\uparrow$}} & 69.61 & \textbf{71.47 \textcolor{ForestGreen}{$\uparrow$}} & 61.32 & \textbf{65.57 \textcolor{ForestGreen}{$\uparrow$}} & N.A. & N.A. & N.A. & N.A. \\
\colorbox{orange!20}{cutract (n=40)} & 69.87 & \textbf{71.72 \textcolor{ForestGreen}{$\uparrow$}} & 68.47 & \textbf{69.56 \textcolor{ForestGreen}{$\uparrow$}} & 68.74 & \textbf{72.36 \textcolor{ForestGreen}{$\uparrow$}} & 64.34 & \textbf{67.46 \textcolor{ForestGreen}{$\uparrow$}} & N.A. & N.A. & N.A. & N.A. \\
\colorbox{orange!20}{maggic (n=40)} & 59.29 & \textbf{60.77 \textcolor{ForestGreen}{$\uparrow$}} & 57.50 & \textbf{59.15 \textcolor{ForestGreen}{$\uparrow$}} & 52.43 & \textbf{53.79 \textcolor{ForestGreen}{$\uparrow$}} & 52.61 & \textbf{53.45 \textcolor{ForestGreen}{$\uparrow$}} & N.A. & N.A. & N.A. & N.A. \\
\colorbox{orange!20}{seer (n=40)} & 84.29 & \textbf{84.93 \textcolor{ForestGreen}{$\uparrow$}} & 83.46 & \textbf{84.44 \textcolor{ForestGreen}{$\uparrow$}} & 83.88 & \textbf{85.21 \textcolor{ForestGreen}{$\uparrow$}} & 78.82 & \textbf{82.65 \textcolor{ForestGreen}{$\uparrow$}} & N.A. & N.A. & N.A. & N.A. \\
\colorbox{green!20}{compas (n=40)} & 67.57 & \textbf{67.85 \textcolor{ForestGreen}{$\uparrow$}} & 61.34 & \textbf{62.84 \textcolor{ForestGreen}{$\uparrow$}} & 57.07 & \textbf{60.87 \textcolor{ForestGreen}{$\uparrow$}} & 59.52 & \textbf{61.46 \textcolor{ForestGreen}{$\uparrow$}} & 57.48	& \textbf{61.20\textcolor{ForestGreen}{$\uparrow$}} &  58.90	& \textbf{63.32\textcolor{ForestGreen}{$\uparrow$}}	 \\
\colorbox{green!20}{adult (n=40)} & 48.31 & \textbf{73.82 \textcolor{ForestGreen}{$\uparrow$}} & 49.21 & \textbf{74.27 \textcolor{ForestGreen}{$\uparrow$}} & 48.80 & \textbf{74.13 \textcolor{ForestGreen}{$\uparrow$}} & 54.34 & \textbf{69.31 \textcolor{ForestGreen}{$\uparrow$}} &	64.44	& \textbf{74.83\textcolor{ForestGreen}{$\uparrow$}} & 56.59 & 	\textbf{78.30 \textcolor{ForestGreen}{$\uparrow$}} \\
\colorbox{green!20}{drug (n=40)} & 74.30 & \textbf{75.79 \textcolor{ForestGreen}{$\uparrow$}} & 71.33 & \textbf{72.76 \textcolor{ForestGreen}{$\uparrow$}} & 73.12 & \textbf{74.12 \textcolor{ForestGreen}{$\uparrow$}} & 69.84 & \textbf{72.50 \textcolor{ForestGreen}{$\uparrow$}} & 64.14	& \textbf{68.14 \textcolor{ForestGreen}{$\uparrow$}} &  74.34	& \textbf{76.07 \textcolor{ForestGreen}{$\uparrow$}}	 \\ \hline\hline
\colorbox{orange!20}{covid (n=100)} & 73.77 & \textbf{74.71 \textcolor{ForestGreen}{$\uparrow$}} & 70.71 & \textbf{72.76 \textcolor{ForestGreen}{$\uparrow$}} & 71.02 & \textbf{73.37 \textcolor{ForestGreen}{$\uparrow$}} & 63.76 & \textbf{70.68 \textcolor{ForestGreen}{$\uparrow$}} & N.A. & N.A. & N.A. & N.A.  \\
\colorbox{orange!20}{cutract (n=100)} & 70.20 & \textbf{72.51 \textcolor{ForestGreen}{$\uparrow$}} & 69.97 & \textbf{71.94 \textcolor{ForestGreen}{$\uparrow$}} & 68.92 & \textbf{71.17 \textcolor{ForestGreen}{$\uparrow$}} & 64.81 & \textbf{69.85 \textcolor{ForestGreen}{$\uparrow$}}  & N.A. & N.A. & N.A. & N.A. \\
\colorbox{orange!20}{maggic (n=100)} & 58.98 & \textbf{61.32 \textcolor{ForestGreen}{$\uparrow$}} & 55.71 & \textbf{58.90 \textcolor{ForestGreen}{$\uparrow$}} & 52.53 & \textbf{53.36 \textcolor{ForestGreen}{$\uparrow$}} & 54.27 & 53.65 & N.A. & N.A. & N.A. & N.A.  \\
\colorbox{orange!20}{seer (n=100)} & 84.45 & \textbf{85.37 \textcolor{ForestGreen}{$\uparrow$}} & 83.92 & \textbf{85.08 \textcolor{ForestGreen}{$\uparrow$}} & 82.23 & \textbf{84.36 \textcolor{ForestGreen}{$\uparrow$}} & 81.00 & \textbf{81.99 \textcolor{ForestGreen}{$\uparrow$}}  & N.A. & N.A. & N.A. & N.A. \\
\colorbox{green!20}{compas (n=100)} & 68.02 & \textbf{68.19 \textcolor{ForestGreen}{$\uparrow$}} & 60.10 & \textbf{62.47 \textcolor{ForestGreen}{$\uparrow$}} & 53.74 & \textbf{61.28 \textcolor{ForestGreen}{$\uparrow$}} & 57.90 & \textbf{61.79 \textcolor{ForestGreen}{$\uparrow$}} & 60.96	& \textbf{63.36 \textcolor{ForestGreen}{$\uparrow$}} & 59.07	& \textbf{63.13 \textcolor{ForestGreen}{$\uparrow$}}	 \\
\colorbox{green!20}{adult (n=100)} & 46.09 & \textbf{74.57 \textcolor{ForestGreen}{$\uparrow$}} & 47.56 & \textbf{73.97 \textcolor{ForestGreen}{$\uparrow$}} & 40.51 & \textbf{71.08 \textcolor{ForestGreen}{$\uparrow$}} & 48.85 & \textbf{72.91 \textcolor{ForestGreen}{$\uparrow$}}  & 54.45	& \textbf{76.67 \textcolor{ForestGreen}{$\uparrow$}} & 54.27	& \textbf{77.61 \textcolor{ForestGreen}{$\uparrow$}}	 \\
\colorbox{green!20}{drug (n=100)} & 76.24 & \textbf{76.74 \textcolor{ForestGreen}{$\uparrow$}} & 69.46 & \textbf{71.05 \textcolor{ForestGreen}{$\uparrow$}} & 74.02 & \textbf{76.55 \textcolor{ForestGreen}{$\uparrow$}} & 66.86 & \textbf{75.31 \textcolor{ForestGreen}{$\uparrow$}} & 71.47	& \textbf{75.00 \textcolor{ForestGreen}{$\uparrow$}} & 73.22	& \textbf{75.83 \textcolor{ForestGreen}{$\uparrow$}}	 \\ \hline\hline
\colorbox{orange!20}{covid (n=200)} & 73.40 & \textbf{74.62 \textcolor{ForestGreen}{$\uparrow$}} & 70.70 & \textbf{73.12 \textcolor{ForestGreen}{$\uparrow$}} & 70.81 & \textbf{73.26 \textcolor{ForestGreen}{$\uparrow$}} & 62.53 & \textbf{70.67 \textcolor{ForestGreen}{$\uparrow$}}  & N.A. & N.A. & N.A. & N.A. \\
\colorbox{orange!20}{cutract (n=200)} & 71.39 & \textbf{73.01 \textcolor{ForestGreen}{$\uparrow$}} & 70.28 & \textbf{72.39 \textcolor{ForestGreen}{$\uparrow$}} & 67.69 & \textbf{70.29 \textcolor{ForestGreen}{$\uparrow$}} & 65.77 & \textbf{69.11 \textcolor{ForestGreen}{$\uparrow$}} & N.A. & N.A. & N.A. & N.A. \\
\colorbox{orange!20}{maggic (n=200)} & 58.92 & \textbf{61.41 \textcolor{ForestGreen}{$\uparrow$}} & 57.33 & \textbf{60.16 \textcolor{ForestGreen}{$\uparrow$}} & 52.78 & 52.40 & 52.56 & \textbf{52.90 \textcolor{ForestGreen}{$\uparrow$}} & N.A. & N.A. & N.A. & N.A.  \\
\colorbox{orange!20}{seer (n=200)} & 84.39 & \textbf{85.56 \textcolor{ForestGreen}{$\uparrow$}} & 83.48 & \textbf{84.80 \textcolor{ForestGreen}{$\uparrow$}} & 83.15 & \textbf{84.18 \textcolor{ForestGreen}{$\uparrow$}} & 80.88 & \textbf{82.74 \textcolor{ForestGreen}{$\uparrow$}} & N.A. & N.A. & N.A. & N.A. \\
\colorbox{green!20}{compas (n=200)} & 67.02 & \textbf{68.15 \textcolor{ForestGreen}{$\uparrow$}} & 60.48 & \textbf{63.39 \textcolor{ForestGreen}{$\uparrow$}} & 53.96 & \textbf{59.22 \textcolor{ForestGreen}{$\uparrow$}} & 57.56 & \textbf{59.97 \textcolor{ForestGreen}{$\uparrow$}} & 61.82 &	\textbf{63.53 \textcolor{ForestGreen}{$\uparrow$}} & 60.99	& \textbf{64.72 \textcolor{ForestGreen}{$\uparrow$}}	\\
\colorbox{green!20}{adult (n=200)} & 40.96 & \textbf{75.84 \textcolor{ForestGreen}{$\uparrow$}} & 49.89 & \textbf{76.11 \textcolor{ForestGreen}{$\uparrow$}} & 44.13 & \textbf{74.23 \textcolor{ForestGreen}{$\uparrow$}} & 48.25 & \textbf{78.35 \textcolor{ForestGreen}{$\uparrow$}} & 54.42 &	\textbf{76.23 \textcolor{ForestGreen}{$\uparrow$}} & 42.42 & 	\textbf{76.06 \textcolor{ForestGreen}{$\uparrow$}}	 \\
\colorbox{green!20}{drug (n=200)} & 75.58 & \textbf{76.06 \textcolor{ForestGreen}{$\uparrow$}} & 70.66 & \textbf{72.81 \textcolor{ForestGreen}{$\uparrow$}} & 71.89 & \textbf{76.54 \textcolor{ForestGreen}{$\uparrow$}} & 70.88 & \textbf{75.46 \textcolor{ForestGreen}{$\uparrow$}} & 69.12 & 	\textbf{74.22 \textcolor{ForestGreen}{$\uparrow$}} & 73.43	& \textbf{76.31 \textcolor{ForestGreen}{$\uparrow$}}	 \\
\hline
\hline
\bottomrule
\end{tabular}}
\label{table:open-source-compare}
\end{table*}

\clearpage

\subsection{Ablation for contextual information on Compas} \label{app:compas_context}
\textcolor{TealBlue}{We conduct a similar experiment as in Table \ref{tab:metrics_covid}, and use the dataset Compas. We report the results in Table \ref{tab:metrics-compas}.}

\begin{minipage}{1\textwidth}
    \centering
    \captionof{table}{\textcolor{TealBlue}{Including contextual information in the prompt improves  precision (P), recall (R), and utility (U) in low-sample settings (results shown for Compas)}.}
    \scalebox{0.8}{

\begin{tabular}{c|ccccccccc}
\hline
\makecell{$n_{\mathrm{samples}}$ \\ in $\Dtrain$} & \multicolumn3{c}{\makecell{GPT-4 \\  w/ context}} & \multicolumn3{c}{\makecell{GPT-4 \\  no context}} & \multicolumn3{c}{TVAE} \\
\cmidrule(lr){2-4}\cmidrule(lr){5-7}\cmidrule(lr){8-10}
 & P & R & U & P & R & U & P & R & U  \\
\hline
20 & $\bf 0.69_{\scriptsize(0.02)}$ & $\bf 0.88_{\scriptsize(0.02)}$ & $\bf 0.69_{\scriptsize(0.02)}$ & $0.27_{\scriptsize(0.03)}$ & $\bf 0.89_{\scriptsize(0.03)}$ & $0.60_{\scriptsize(0.03)}$ & $0.43_{\scriptsize(0.02)}$ & $0.43_{\scriptsize(0.05)}$ & $0.55_{\scriptsize(0.04)}$\\
40 & $\bf 0.70_{\scriptsize(0.0)}$ & $\bf 0.92_{\scriptsize(0.01)}$ & $\bf 0.65_{\scriptsize(0.03)}$ & $0.31_{\scriptsize(0.06)}$ & $0.84_{\scriptsize(0.03)}$ & $0.57_{\scriptsize(0.01)}$ & $0.54_{\scriptsize(0.02)}$ & $0.80_{\scriptsize(0.02)}$ & $0.50_{\scriptsize(0.04)}$\\
100 & $\bf 0.69_{\scriptsize(0.02)}$ & $\bf 0.89_{\scriptsize(0.02)}$ & $\bf 0.69_{\scriptsize(0.01)}$ & $0.34_{\scriptsize(0.1)}$ & $0.85_{\scriptsize(0.05)}$ & $0.62_{\scriptsize(0.01)}$ & $0.60_{\scriptsize(0.03)}$ & $0.86_{\scriptsize(0.02)}$ & $0.59_{\scriptsize(0.03)}$\\
200 & $\bf 0.70_{\scriptsize(0.01)}$ & $\bf 0.89_{\scriptsize(0.02)}$ & $\bf 0.69_{\scriptsize(0.01)}$ & $0.31_{\scriptsize(0.05)}$ & $0.87_{\scriptsize(0.03)}$ & $0.58_{\scriptsize(0.05)}$ & $0.65_{\scriptsize(0.02)}$ & $0.88_{\scriptsize(0.01)}$ & $0.63_{\scriptsize(0.01)}$\\
   \hline
\end{tabular}}
    \label{tab:metrics-compas}
\end{minipage}

These results highlight the importance of incorporating contextual information in the prompt, as it enables to exploit the prior knowledge of the LLM.

\subsection{Comparison to random noise baseline}

\textcolor{TealBlue}{We compare {\name} to a random noise baseline, where we augment the dataset with random additive Gaussian noise. In order to capture the correlations between the different features, we fit a Kernel Density Estimator with a Gaussian kernel and bandwidth given by Scott's rule. We then sample $1000$ points to create an augmented dataset $D_{\mathrm{syn}}$.}
\textcolor{TealBlue}{We report the performance gap between {\name} and this baseline (with and without curation) for the Covid and Compas datasets in Figure \ref{fig:random_baseline}.}
We observe that the random noise baseline does not match the performance of {\name} (i.e. has a performance gap), although the baseline naturally improves as the dataset $D_{\mathrm{train}}$ grows in size.

\begin{figure}[!h]
    \centering
    \includegraphics[scale=0.42]{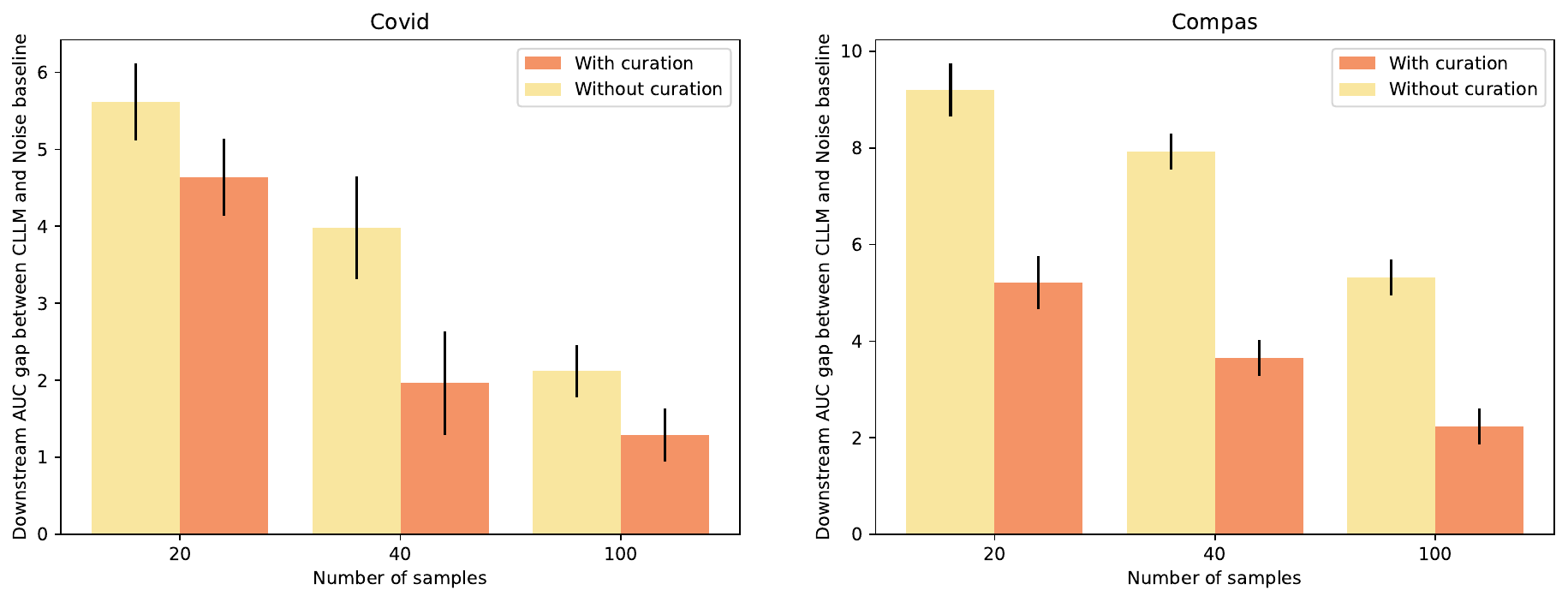}
    \caption{The random noise baseline does not match the performance of {\name}}
    \label{fig:random_baseline}
\end{figure}

\newpage

\subsection{Detailed results for Section \ref{performance}}
We report additional results for Sec. \ref{performance}, showing the AUC for each downstream model (XGBoost, Random forest, Logistic regression, Decision tree).
As we can see, the conclusion that curation helps improve downstream performance holds for each of these various downstream models, as is indicated by the green arrows in Tables \ref{table:results_rf}, \ref{table:results_xgb}, \ref{table:results_dt}, \ref{table:results_lr}.

\begingroup
\setlength{\tabcolsep}{1.5pt} %
\renewcommand{\arraystretch}{1} %
\begin{table}[!h]

\caption{AUC for the RF model on $\Dtest$ where curation improves performance for all methods across all sample sizes $n$, as indicated by \textcolor{ForestGreen}{$\uparrow$}.}
\centering
\scalebox{0.57}{
\begin{tabular}{l|cc|cc|cc|cc|cc|cc|cc|cc|cc}
\toprule
& \multicolumn{2}{c|}{Real data} & \multicolumn{4}{c|}{\name ~(OURS)} &  \multicolumn{12}{c}{Baselines} \\
\cmidrule(lr){2-3}\cmidrule(lr){4-7}\cmidrule(lr){8-19}
& \multicolumn{2}{c|}{} & \multicolumn{2}{c}{GPT-4} & \multicolumn{2}{c|}{GPT-3.5} & \multicolumn{2}{c}{CTGAN} & \multicolumn{2}{c}{TabDDPM} & \multicolumn{2}{c}{GReaT}   & \multicolumn{2}{c}{NFLOW} & \multicolumn{2}{c}{SMOTE} & \multicolumn{2}{c}{TVAE} \\
\cmidrule(lr){2-3}\cmidrule(lr){4-5}\cmidrule(lr){6-7}\cmidrule(lr){8-9}\cmidrule(lr){10-11}\cmidrule(lr){12-13}\cmidrule(lr){14-15}\cmidrule(lr){16-17}\cmidrule(lr){18-19}
Dataset & $\Doracle$ & $\Dtrain$ & Uncur. & Cur. & Uncur. & Cur. & Uncur. & Cur. & Uncur. & Cur. & Uncur. & Cur. & Uncur. & Cur. & Uncur. & Cur. & Uncur. & Cur. \\
\hline\hline
covid (n=20) & 76.11 & 72.52 & \textbf{75.67} & 75.59 & 72.37 & 73.32 \textcolor{ForestGreen}{$\uparrow$} & 61.66 & 65.24 \textcolor{ForestGreen}{$\uparrow$} & 70.22 & 70.13 & 57.78 & 68.24 \textcolor{ForestGreen}{$\uparrow$} & 65.54 & 70.72 \textcolor{ForestGreen}{$\uparrow$} & 71.40 & 71.42 \textcolor{ForestGreen}{$\uparrow$} & 63.90 & 67.69 \textcolor{ForestGreen}{$\uparrow$} \\
cutract (n=20) & 74.16 & 73.34 & 73.45 & \textbf{74.24 \textcolor{ForestGreen}{$\uparrow$}} & 72.22 & 73.51 \textcolor{ForestGreen}{$\uparrow$} & 66.24 & 70.45 \textcolor{ForestGreen}{$\uparrow$} & 69.73 & 70.49 \textcolor{ForestGreen}{$\uparrow$} & 52.61 & 68.11 \textcolor{ForestGreen}{$\uparrow$} & 66.63 & 72.39 \textcolor{ForestGreen}{$\uparrow$} & 71.42 & 72.34 \textcolor{ForestGreen}{$\uparrow$} & 71.18 & 71.89 \textcolor{ForestGreen}{$\uparrow$} \\
maggic (n=20) & 71.28 & 59.18 & 62.92 & \textbf{63.97 \textcolor{ForestGreen}{$\uparrow$}} & 60.10 & 61.25 \textcolor{ForestGreen}{$\uparrow$} & 53.21 & 55.14 \textcolor{ForestGreen}{$\uparrow$} & 55.61 & 56.50 \textcolor{ForestGreen}{$\uparrow$} & 50.92 & 57.23 \textcolor{ForestGreen}{$\uparrow$} & 55.91 & 59.07 \textcolor{ForestGreen}{$\uparrow$} & 58.20 & 58.55 \textcolor{ForestGreen}{$\uparrow$} & 55.43 & 57.71 \textcolor{ForestGreen}{$\uparrow$} \\
seer (n=20) & 90.09 & 85.22 & 86.30 & \textbf{86.82 \textcolor{ForestGreen}{$\uparrow$}} & 85.86 & 85.73 & 78.11 & 80.52 \textcolor{ForestGreen}{$\uparrow$} & 82.61 & 82.74 \textcolor{ForestGreen}{$\uparrow$} & 47.57 & 75.70 \textcolor{ForestGreen}{$\uparrow$} & 77.35 & 81.45 \textcolor{ForestGreen}{$\uparrow$} & 82.09 & 83.00 \textcolor{ForestGreen}{$\uparrow$} & 77.63 & 81.16 \textcolor{ForestGreen}{$\uparrow$} \\
compas (n=20) & 66.79 & 64.47 & \textbf{68.40} & 67.85 & 61.93 & 64.29 \textcolor{ForestGreen}{$\uparrow$} & 56.87 & 63.21 \textcolor{ForestGreen}{$\uparrow$} & 58.56 & 61.26 \textcolor{ForestGreen}{$\uparrow$} & 52.23 & 63.30 \textcolor{ForestGreen}{$\uparrow$} & 59.60 & 64.79 \textcolor{ForestGreen}{$\uparrow$} & 61.05 & 62.18 \textcolor{ForestGreen}{$\uparrow$} & 59.31 & 63.37 \textcolor{ForestGreen}{$\uparrow$} \\
adult (n=20) & 85.83 & 83.31 & 51.35 & 73.45 \textcolor{ForestGreen}{$\uparrow$} & 49.17 & 74.27 \textcolor{ForestGreen}{$\uparrow$} & 75.63 & 78.99 \textcolor{ForestGreen}{$\uparrow$} & 77.27 & 77.64 \textcolor{ForestGreen}{$\uparrow$} & 69.35 & 78.79 \textcolor{ForestGreen}{$\uparrow$} & 68.32 & 78.27 \textcolor{ForestGreen}{$\uparrow$} & 77.75 & 78.34 \textcolor{ForestGreen}{$\uparrow$} & 76.13 & 79.02 \textcolor{ForestGreen}{$\uparrow$} \\
drug (n=20) & 83.12 & 77.07 & \textbf{79.23} & 78.90 & 77.11 & 76.95 & 72.81 & 77.24 \textcolor{ForestGreen}{$\uparrow$} & 72.00 & 74.03 \textcolor{ForestGreen}{$\uparrow$} & 63.34 & 73.29 \textcolor{ForestGreen}{$\uparrow$} & 65.47 & 70.19 \textcolor{ForestGreen}{$\uparrow$} & 75.48 & 75.36 & 72.22 & 74.32 \textcolor{ForestGreen}{$\uparrow$} \\ \hline
covid (n=40) & 76.11 & 75.21 & 75.63 & \textbf{75.70 \textcolor{ForestGreen}{$\uparrow$}} & 72.27 & 73.59 \textcolor{ForestGreen}{$\uparrow$} & 66.64 & 70.75 \textcolor{ForestGreen}{$\uparrow$} & 75.56 & 75.39 & 56.91 & 70.88 \textcolor{ForestGreen}{$\uparrow$} & 70.47 & 73.15 \textcolor{ForestGreen}{$\uparrow$} & 73.59 & 73.63 \textcolor{ForestGreen}{$\uparrow$} & 64.27 & 69.71 \textcolor{ForestGreen}{$\uparrow$} \\
cutract (n=40) & 74.16 & 72.31 & 71.67 & \textbf{73.20 \textcolor{ForestGreen}{$\uparrow$}} & 69.92 & 71.01 \textcolor{ForestGreen}{$\uparrow$} & 64.67 & 69.21 \textcolor{ForestGreen}{$\uparrow$} & 69.16 & 69.67 \textcolor{ForestGreen}{$\uparrow$} & 53.41 & 68.93 \textcolor{ForestGreen}{$\uparrow$} & 61.23 & 69.21 \textcolor{ForestGreen}{$\uparrow$} & 70.82 & 71.21 \textcolor{ForestGreen}{$\uparrow$} & 59.99 & 67.56 \textcolor{ForestGreen}{$\uparrow$} \\
maggic (n=40) & 71.28 & 60.91 & 61.80 & \textbf{63.11 \textcolor{ForestGreen}{$\uparrow$}} & 59.38 & 61.50 \textcolor{ForestGreen}{$\uparrow$} & 56.49 & 58.17 \textcolor{ForestGreen}{$\uparrow$} & 56.50 & 58.21 \textcolor{ForestGreen}{$\uparrow$} & 48.43 & 57.84 \textcolor{ForestGreen}{$\uparrow$} & 55.47 & 59.94 \textcolor{ForestGreen}{$\uparrow$} & 59.82 & 60.11 \textcolor{ForestGreen}{$\uparrow$} & 56.76 & 58.92 \textcolor{ForestGreen}{$\uparrow$} \\
seer (n=40) & 90.09 & 86.96 & 86.01 & 86.29 \textcolor{ForestGreen}{$\uparrow$} & 86.73 & \textbf{87.09 \textcolor{ForestGreen}{$\uparrow$}} & 83.46 & 86.62 \textcolor{ForestGreen}{$\uparrow$} & 85.86 & 85.19 & 54.71 & 83.22 \textcolor{ForestGreen}{$\uparrow$} & 83.23 & 86.81 \textcolor{ForestGreen}{$\uparrow$} & 83.76 & 84.82 \textcolor{ForestGreen}{$\uparrow$} & 80.34 & 85.82 \textcolor{ForestGreen}{$\uparrow$} \\
compas (n=40) & 66.79 & 62.73 & 68.06 & \textbf{68.13 \textcolor{ForestGreen}{$\uparrow$}} & 61.17 & 62.27 \textcolor{ForestGreen}{$\uparrow$} & 56.05 & 60.92 \textcolor{ForestGreen}{$\uparrow$} & 59.76 & 61.03 \textcolor{ForestGreen}{$\uparrow$} & 57.82 & 64.25 \textcolor{ForestGreen}{$\uparrow$} & 58.89 & 63.79 \textcolor{ForestGreen}{$\uparrow$} & 61.20 & 61.25 \textcolor{ForestGreen}{$\uparrow$} & 55.89 & 60.54 \textcolor{ForestGreen}{$\uparrow$} \\
adult (n=40) & 85.83 & 83.61 & 50.26 & 75.55 \textcolor{ForestGreen}{$\uparrow$} & 46.68 & 76.63 \textcolor{ForestGreen}{$\uparrow$} & 75.13 & 81.56 \textcolor{ForestGreen}{$\uparrow$} & 76.10 & 80.40 \textcolor{ForestGreen}{$\uparrow$} & 68.04 & 81.08 \textcolor{ForestGreen}{$\uparrow$} & 73.29 & 81.46 \textcolor{ForestGreen}{$\uparrow$} & 80.35 & 81.10 \textcolor{ForestGreen}{$\uparrow$} & 75.81 & 82.11 \textcolor{ForestGreen}{$\uparrow$} \\
drug (n=40) & 83.12 & 78.81 & 78.35 & \textbf{79.54 \textcolor{ForestGreen}{$\uparrow$}} & 77.33 & 77.94 \textcolor{ForestGreen}{$\uparrow$} & 74.68 & 76.64 \textcolor{ForestGreen}{$\uparrow$} & 74.50 & 77.18 \textcolor{ForestGreen}{$\uparrow$} & 70.25 & 77.77 \textcolor{ForestGreen}{$\uparrow$} & 65.86 & 74.55 \textcolor{ForestGreen}{$\uparrow$} & 74.84 & 75.26 \textcolor{ForestGreen}{$\uparrow$} & 71.29 & 75.98 \textcolor{ForestGreen}{$\uparrow$} \\ \hline
covid (n=100) & 76.11 & 75.78 & 75.86 & \textbf{76.33 \textcolor{ForestGreen}{$\uparrow$}} & 73.02 & 74.40 \textcolor{ForestGreen}{$\uparrow$} & 72.00 & 74.74 \textcolor{ForestGreen}{$\uparrow$} & 74.64 & 75.74 \textcolor{ForestGreen}{$\uparrow$} & 65.76 & 74.69 \textcolor{ForestGreen}{$\uparrow$} & 67.05 & 75.47 \textcolor{ForestGreen}{$\uparrow$} & 74.00 & 74.09 \textcolor{ForestGreen}{$\uparrow$} & 72.11 & 74.44 \textcolor{ForestGreen}{$\uparrow$} \\
cutract (n=100) & 74.16 & 73.93 & 72.92 & 74.73 \textcolor{ForestGreen}{$\uparrow$} & 72.51 & 73.93 \textcolor{ForestGreen}{$\uparrow$} & 70.18 & 74.26 \textcolor{ForestGreen}{$\uparrow$} & 71.71 & 73.69 \textcolor{ForestGreen}{$\uparrow$} & 55.48 & 70.79 \textcolor{ForestGreen}{$\uparrow$} & 69.79 & \textbf{74.75 \textcolor{ForestGreen}{$\uparrow$}} & 71.18 & 72.47 \textcolor{ForestGreen}{$\uparrow$} & 68.47 & 73.78 \textcolor{ForestGreen}{$\uparrow$} \\
maggic (n=100) & 71.28 & 63.06 & 60.99 & \textbf{63.36 \textcolor{ForestGreen}{$\uparrow$}} & 57.97 & 60.86 \textcolor{ForestGreen}{$\uparrow$} & 58.98 & 60.76 \textcolor{ForestGreen}{$\uparrow$} & 58.88 & 60.13 \textcolor{ForestGreen}{$\uparrow$} & 49.66 & 59.57 \textcolor{ForestGreen}{$\uparrow$} & 57.82 & 62.24 \textcolor{ForestGreen}{$\uparrow$} & 61.55 & 61.90 \textcolor{ForestGreen}{$\uparrow$} & 57.46 & 60.55 \textcolor{ForestGreen}{$\uparrow$} \\
seer (n=100) & 90.00 & 87.53 & 86.33 & 87.31 \textcolor{ForestGreen}{$\uparrow$} & 86.40 & 87.06 \textcolor{ForestGreen}{$\uparrow$} & 84.59 & 87.27 \textcolor{ForestGreen}{$\uparrow$} & 85.89 & 87.00 \textcolor{ForestGreen}{$\uparrow$} & 70.32 & 85.97 \textcolor{ForestGreen}{$\uparrow$} & 84.30 & 87.52 \textcolor{ForestGreen}{$\uparrow$} & 85.12 & 85.82 \textcolor{ForestGreen}{$\uparrow$} & 81.92 & 86.41 \textcolor{ForestGreen}{$\uparrow$} \\
compas (n=100) & 66.79 & 63.18 & 68.44 & \textbf{68.67 \textcolor{ForestGreen}{$\uparrow$}} & 59.41 & 62.00 \textcolor{ForestGreen}{$\uparrow$} & 60.28 & 64.34 \textcolor{ForestGreen}{$\uparrow$} & 60.17 & 62.63 \textcolor{ForestGreen}{$\uparrow$} & 59.32 & 63.98 \textcolor{ForestGreen}{$\uparrow$} & 60.10 & 65.19 \textcolor{ForestGreen}{$\uparrow$} & 61.34 & 61.37 \textcolor{ForestGreen}{$\uparrow$} & 59.58 & 63.37 \textcolor{ForestGreen}{$\uparrow$} \\
adult (n=100) & 85.83 & 84.38 & 46.37 & 76.29 \textcolor{ForestGreen}{$\uparrow$} & 47.30 & 75.74 \textcolor{ForestGreen}{$\uparrow$} & 77.45 & 82.63 \textcolor{ForestGreen}{$\uparrow$} & 80.88 & 81.69 \textcolor{ForestGreen}{$\uparrow$} & 79.23 & 83.33 \textcolor{ForestGreen}{$\uparrow$} & 73.52 & 83.38 \textcolor{ForestGreen}{$\uparrow$} & 82.54 & 83.05 \textcolor{ForestGreen}{$\uparrow$} & 76.60 & 82.28 \textcolor{ForestGreen}{$\uparrow$} \\
drug (n=100) & 83.12 & 80.31 & 79.29 & 79.75 \textcolor{ForestGreen}{$\uparrow$} & 74.55 & 76.46 \textcolor{ForestGreen}{$\uparrow$} & 74.54 & 78.01 \textcolor{ForestGreen}{$\uparrow$} & 78.34 & 79.88 \textcolor{ForestGreen}{$\uparrow$} & 73.19 & 79.72 \textcolor{ForestGreen}{$\uparrow$} & 67.01 & 76.46 \textcolor{ForestGreen}{$\uparrow$} & 76.92 & 77.57 \textcolor{ForestGreen}{$\uparrow$} & 74.53 & 78.82 \textcolor{ForestGreen}{$\uparrow$} \\ \hline
covid (n=200) & 76.11 & 76.08 & 75.23 & 75.84 \textcolor{ForestGreen}{$\uparrow$} & 72.62 & 74.99 \textcolor{ForestGreen}{$\uparrow$} & 74.04 & 76.25 \textcolor{ForestGreen}{$\uparrow$} & 75.28 & \textbf{76.82 \textcolor{ForestGreen}{$\uparrow$}} & 67.08 & 75.62 \textcolor{ForestGreen}{$\uparrow$} & 67.73 & 75.67 \textcolor{ForestGreen}{$\uparrow$} & 74.82 & 75.27 \textcolor{ForestGreen}{$\uparrow$} & 70.64 & 75.03 \textcolor{ForestGreen}{$\uparrow$} \\
cutract (n=200) & 74.16 & 74.25 & 73.48 & 75.28 \textcolor{ForestGreen}{$\uparrow$} & 72.01 & 74.37 \textcolor{ForestGreen}{$\uparrow$} & 71.99 & 74.93 \textcolor{ForestGreen}{$\uparrow$} & 74.33 & \textbf{76.19 \textcolor{ForestGreen}{$\uparrow$}} & 68.05 & 74.68 \textcolor{ForestGreen}{$\uparrow$} & 71.47 & 75.67 \textcolor{ForestGreen}{$\uparrow$} & 72.44 & 72.85 \textcolor{ForestGreen}{$\uparrow$} & 68.55 & 74.43 \textcolor{ForestGreen}{$\uparrow$} \\
maggic (n=200) & 71.28 & 64.77 & 61.56 & 63.78 \textcolor{ForestGreen}{$\uparrow$} & 59.76 & 62.53 \textcolor{ForestGreen}{$\uparrow$} & 60.85 & 63.57 \textcolor{ForestGreen}{$\uparrow$} & 58.23 & 58.85 \textcolor{ForestGreen}{$\uparrow$} & 50.84 & 61.49 \textcolor{ForestGreen}{$\uparrow$} & 57.52 & 63.00 \textcolor{ForestGreen}{$\uparrow$} & 63.72 & 64.01 \textcolor{ForestGreen}{$\uparrow$} & 58.92 & 62.08 \textcolor{ForestGreen}{$\uparrow$} \\
seer (n=200) & 90.09 & 88.13 & 86.96 & 87.72 \textcolor{ForestGreen}{$\uparrow$} & 85.28 & 86.70 \textcolor{ForestGreen}{$\uparrow$} & 85.05 & 87.83 \textcolor{ForestGreen}{$\uparrow$} & 87.14 & \textbf{88.80 \textcolor{ForestGreen}{$\uparrow$}} & 84.86 & 88.14 \textcolor{ForestGreen}{$\uparrow$} & 80.80 & 87.46 \textcolor{ForestGreen}{$\uparrow$} & 86.57 & 87.11 \textcolor{ForestGreen}{$\uparrow$} & 82.30 & 87.59 \textcolor{ForestGreen}{$\uparrow$} \\
compas (n=200) & 66.79 & 63.61 & 67.68 & \textbf{68.80 \textcolor{ForestGreen}{$\uparrow$}} & 60.67 & 63.43 \textcolor{ForestGreen}{$\uparrow$} & 61.20 & 65.11 \textcolor{ForestGreen}{$\uparrow$} & 60.88 & 63.87 \textcolor{ForestGreen}{$\uparrow$} & 60.10 & 64.91 \textcolor{ForestGreen}{$\uparrow$} & 56.94 & 64.76 \textcolor{ForestGreen}{$\uparrow$} & 61.57 & 62.76 \textcolor{ForestGreen}{$\uparrow$} & 59.95 & 64.14 \textcolor{ForestGreen}{$\uparrow$} \\
adult (n=200) & 85.83 & 85.16 & 40.18 & 77.61 \textcolor{ForestGreen}{$\uparrow$} & 46.74 & 78.71 \textcolor{ForestGreen}{$\uparrow$} & 80.92 & 84.26 \textcolor{ForestGreen}{$\uparrow$} & 84.50 & \textbf{85.44 \textcolor{ForestGreen}{$\uparrow$}} & 82.73 & 85.05 \textcolor{ForestGreen}{$\uparrow$} & 77.48 & 84.72 \textcolor{ForestGreen}{$\uparrow$} & 84.26 & 84.33 \textcolor{ForestGreen}{$\uparrow$} & 78.12 & 83.94 \textcolor{ForestGreen}{$\uparrow$} \\
drug (n=200) & 83.12 & 81.47 & 78.70 & 78.98 \textcolor{ForestGreen}{$\uparrow$} & 75.87 & 77.59 \textcolor{ForestGreen}{$\uparrow$} & 77.19 & 80.03 \textcolor{ForestGreen}{$\uparrow$} & 73.25 & 75.75 \textcolor{ForestGreen}{$\uparrow$} & 78.50 & 81.13 \textcolor{ForestGreen}{$\uparrow$} & 67.83 & 78.81 \textcolor{ForestGreen}{$\uparrow$} & 79.50 & 80.16 \textcolor{ForestGreen}{$\uparrow$} & 74.81 & 79.45 \textcolor{ForestGreen}{$\uparrow$} \\ \hline
\bottomrule
\end{tabular}}
\label{table:results_rf}
\end{table}
\endgroup

\begingroup
\setlength{\tabcolsep}{1.5pt} %
\renewcommand{\arraystretch}{1} %
\begin{table}[!h]
\caption{AUC for the XGB model on $\Dtest$ where curation improves performance for all methods across all sample sizes $n$, as indicated by \textcolor{ForestGreen}{$\uparrow$}.}
\centering
\scalebox{0.575}{
\begin{tabular}{l|cc|cc|cc|cc|cc|cc|cc|cc|cc}
\toprule
& \multicolumn{2}{c|}{Real data} & \multicolumn{4}{c|}{\name ~(OURS)} &  \multicolumn{12}{c}{Baselines} \\
\cmidrule(lr){2-3}\cmidrule(lr){4-7}\cmidrule(lr){8-19}
& \multicolumn{2}{c|}{} & \multicolumn{2}{c}{GPT-4} & \multicolumn{2}{c|}{GPT-3.5} & \multicolumn{2}{c}{CTGAN} & \multicolumn{2}{c}{TabDDPM} & \multicolumn{2}{c}{GReaT}   & \multicolumn{2}{c}{NFLOW} & \multicolumn{2}{c}{SMOTE} & \multicolumn{2}{c}{TVAE} \\
\cmidrule(lr){2-3}\cmidrule(lr){4-5}\cmidrule(lr){6-7}\cmidrule(lr){8-9}\cmidrule(lr){10-11}\cmidrule(lr){12-13}\cmidrule(lr){14-15}\cmidrule(lr){16-17}\cmidrule(lr){18-19}
Dataset & $\Doracle$ & $\Dtrain$ & Uncur. & Cur. & Uncur. & Cur. & Uncur. & Cur. & Uncur. & Cur. & Uncur. & Cur. & Uncur. & Cur. & Uncur. & Cur. & Uncur. & Cur. \\
\hline\hline
covid (n=20) & 77.59 & 69.70 & \textbf{75.59} & 75.49 & 71.27 & 72.25 \textcolor{ForestGreen}{$\uparrow$} & 61.40 & 65.96 \textcolor{ForestGreen}{$\uparrow$} & 69.39 & 68.21 & 55.99 & 67.06 \textcolor{ForestGreen}{$\uparrow$} & 62.28 & 69.67 \textcolor{ForestGreen}{$\uparrow$} & 68.49 & 68.43 & 62.49 & 67.16 \textcolor{ForestGreen}{$\uparrow$} \\
cutract (n=20) & 73.14 & 71.04 & 73.26 & \textbf{74.30 \textcolor{ForestGreen}{$\uparrow$}} & 71.62 & 73.04 \textcolor{ForestGreen}{$\uparrow$} & 65.13 & 69.67 \textcolor{ForestGreen}{$\uparrow$} & 68.46 & 68.87 \textcolor{ForestGreen}{$\uparrow$} & 51.99 & 68.31 \textcolor{ForestGreen}{$\uparrow$} & 62.93 & 70.87 \textcolor{ForestGreen}{$\uparrow$} & 68.28 & 70.72 \textcolor{ForestGreen}{$\uparrow$} & 69.96 & 71.98 \textcolor{ForestGreen}{$\uparrow$} \\
maggic (n=20) & 71.38 & 57.98 & 61.54 & \textbf{62.81 \textcolor{ForestGreen}{$\uparrow$}} & 58.76 & 60.01 \textcolor{ForestGreen}{$\uparrow$} & 53.62 & 55.44 \textcolor{ForestGreen}{$\uparrow$} & 55.97 & 57.06 \textcolor{ForestGreen}{$\uparrow$} & 50.14 & 56.96 \textcolor{ForestGreen}{$\uparrow$} & 54.91 & 58.22 \textcolor{ForestGreen}{$\uparrow$} & 56.79 & 57.15 \textcolor{ForestGreen}{$\uparrow$} & 55.41 & 57.72 \textcolor{ForestGreen}{$\uparrow$} \\
seer (n=20) & 91.80 & 80.41 & \textbf{87.00} & 86.56 & 85.67 & 85.42 & 76.84 & 80.00 \textcolor{ForestGreen}{$\uparrow$} & 81.79 & 80.21 & 45.06 & 75.39 \textcolor{ForestGreen}{$\uparrow$} & 76.25 & 81.45 \textcolor{ForestGreen}{$\uparrow$} & 78.57 & 79.78 \textcolor{ForestGreen}{$\uparrow$} & 75.69 & 81.01 \textcolor{ForestGreen}{$\uparrow$} \\
compas (n=20) & 69.17 & 65.49 & \textbf{69.71} & 68.79 & 62.55 & 64.68 \textcolor{ForestGreen}{$\uparrow$} & 56.34 & 63.92 \textcolor{ForestGreen}{$\uparrow$} & 58.19 & 61.42 \textcolor{ForestGreen}{$\uparrow$} & 51.73 & 63.51 \textcolor{ForestGreen}{$\uparrow$} & 59.01 & 64.84 \textcolor{ForestGreen}{$\uparrow$} & 62.35 & 63.46 \textcolor{ForestGreen}{$\uparrow$} & 59.18 & 63.62 \textcolor{ForestGreen}{$\uparrow$} \\
adult (n=20) & 88.64 & 79.21 & 52.12 & 74.28 \textcolor{ForestGreen}{$\uparrow$} & 47.70 & 72.91 \textcolor{ForestGreen}{$\uparrow$} & 71.70 & 78.26 \textcolor{ForestGreen}{$\uparrow$} & 76.50 & 77.22 \textcolor{ForestGreen}{$\uparrow$} & 67.79 & 78.67 \textcolor{ForestGreen}{$\uparrow$} & 66.79 & 77.77 \textcolor{ForestGreen}{$\uparrow$} & 77.11 & 77.97 \textcolor{ForestGreen}{$\uparrow$} & 75.46 & \textbf{79.35 \textcolor{ForestGreen}{$\uparrow$}} \\
drug (n=20) & 81.06 & 69.26 & \textbf{78.17} & 77.28 & 75.23 & 75.71 \textcolor{ForestGreen}{$\uparrow$} & 70.60 & 74.31 \textcolor{ForestGreen}{$\uparrow$} & 70.29 & 71.25 \textcolor{ForestGreen}{$\uparrow$} & 57.18 & 69.67 \textcolor{ForestGreen}{$\uparrow$} & 62.96 & 69.12 \textcolor{ForestGreen}{$\uparrow$} & 70.84 & 72.30 \textcolor{ForestGreen}{$\uparrow$} & 69.78 & 72.45 \textcolor{ForestGreen}{$\uparrow$} \\ \hline
covid (n=40) & 77.59 & 71.57 & 75.23 & \textbf{75.36 \textcolor{ForestGreen}{$\uparrow$}} & 71.55 & 73.37 \textcolor{ForestGreen}{$\uparrow$} & 66.17 & 70.85 \textcolor{ForestGreen}{$\uparrow$} & 72.21 & 72.68 \textcolor{ForestGreen}{$\uparrow$} & 56.03 & 70.80 \textcolor{ForestGreen}{$\uparrow$} & 67.39 & 71.82 \textcolor{ForestGreen}{$\uparrow$} & 70.02 & 69.96 & 61.19 & 69.77 \textcolor{ForestGreen}{$\uparrow$} \\
cutract (n=40) & 73.14 & 69.50 & 71.26 & \textbf{73.57 \textcolor{ForestGreen}{$\uparrow$}} & 69.66 & 71.34 \textcolor{ForestGreen}{$\uparrow$} & 62.52 & 68.69 \textcolor{ForestGreen}{$\uparrow$} & 64.50 & 68.05 \textcolor{ForestGreen}{$\uparrow$} & 51.23 & 68.23 \textcolor{ForestGreen}{$\uparrow$} & 60.74 & 68.91 \textcolor{ForestGreen}{$\uparrow$} & 69.42 & 69.19 & 59.34 & 67.80 \textcolor{ForestGreen}{$\uparrow$} \\
maggic (n=40) & 71.38 & 58.76 & 60.27 & \textbf{61.88 \textcolor{ForestGreen}{$\uparrow$}} & 57.87 & 60.29 \textcolor{ForestGreen}{$\uparrow$} & 55.87 & 58.47 \textcolor{ForestGreen}{$\uparrow$} & 55.38 & 58.03 \textcolor{ForestGreen}{$\uparrow$} & 48.50 & 57.02 \textcolor{ForestGreen}{$\uparrow$} & 54.53 & 59.01 \textcolor{ForestGreen}{$\uparrow$} & 58.12 & 58.29 \textcolor{ForestGreen}{$\uparrow$} & 56.19 & 58.33 \textcolor{ForestGreen}{$\uparrow$} \\
seer (n=40) & 91.80 & 84.51 & 86.63 & \textbf{87.40 \textcolor{ForestGreen}{$\uparrow$}} & 86.27 & 86.80 \textcolor{ForestGreen}{$\uparrow$} & 82.48 & 86.55 \textcolor{ForestGreen}{$\uparrow$} & 81.62 & 81.58 & 52.80 & 83.03 \textcolor{ForestGreen}{$\uparrow$} & 80.04 & 85.84 \textcolor{ForestGreen}{$\uparrow$} & 82.53 & 83.62 \textcolor{ForestGreen}{$\uparrow$} & 79.35 & 85.81 \textcolor{ForestGreen}{$\uparrow$} \\
compas (n=40) & 69.17 & 62.90 & \textbf{69.24} & 68.90 & 61.74 & 63.21 \textcolor{ForestGreen}{$\uparrow$} & 56.63 & 62.55 \textcolor{ForestGreen}{$\uparrow$} & 59.07 & 59.38 \textcolor{ForestGreen}{$\uparrow$} & 58.01 & 64.95 \textcolor{ForestGreen}{$\uparrow$} & 58.44 & 64.44 \textcolor{ForestGreen}{$\uparrow$} & 61.81 & 61.89 \textcolor{ForestGreen}{$\uparrow$} & 56.60 & 62.26 \textcolor{ForestGreen}{$\uparrow$} \\
adult (n=40) & 88.64 & 82.16 & 48.51 & 76.59 \textcolor{ForestGreen}{$\uparrow$} & 45.92 & 76.92 \textcolor{ForestGreen}{$\uparrow$} & 73.38 & 82.53 \textcolor{ForestGreen}{$\uparrow$} & 74.03 & 79.84 \textcolor{ForestGreen}{$\uparrow$} & 66.60 & 80.98 \textcolor{ForestGreen}{$\uparrow$} & 69.47 & 81.56 \textcolor{ForestGreen}{$\uparrow$} & 80.01 & 80.51 \textcolor{ForestGreen}{$\uparrow$} & 74.20 & \textbf{82.53 \textcolor{ForestGreen}{$\uparrow$}} \\
drug (n=40) & 81.06 & 72.72 & 77.03 & \textbf{78.85 \textcolor{ForestGreen}{$\uparrow$}} & 76.37 & 76.48 \textcolor{ForestGreen}{$\uparrow$} & 70.99 & 74.48 \textcolor{ForestGreen}{$\uparrow$} & 74.65 & 76.53 \textcolor{ForestGreen}{$\uparrow$} & 64.52 & 75.09 \textcolor{ForestGreen}{$\uparrow$} & 62.58 & 73.14 \textcolor{ForestGreen}{$\uparrow$} & 71.80 & 72.15 \textcolor{ForestGreen}{$\uparrow$} & 67.61 & 74.04 \textcolor{ForestGreen}{$\uparrow$} \\ \hline
covid (n=100) & 77.59 & 73.16 & 75.55 & \textbf{76.56 \textcolor{ForestGreen}{$\uparrow$}} & 72.69 & 74.36 \textcolor{ForestGreen}{$\uparrow$} & 70.54 & 73.99 \textcolor{ForestGreen}{$\uparrow$} & 72.99 & 74.50 \textcolor{ForestGreen}{$\uparrow$} & 63.40 & 73.98 \textcolor{ForestGreen}{$\uparrow$} & 63.50 & 73.79 \textcolor{ForestGreen}{$\uparrow$} & 72.64 & 73.35 \textcolor{ForestGreen}{$\uparrow$} & 70.45 & 73.76 \textcolor{ForestGreen}{$\uparrow$} \\
cutract (n=100) & 73.14 & 70.49 & 72.29 & 74.01 \textcolor{ForestGreen}{$\uparrow$} & 71.51 & 73.48 \textcolor{ForestGreen}{$\uparrow$} & 68.64 & 73.63 \textcolor{ForestGreen}{$\uparrow$} & 71.32 & \textbf{74.02 \textcolor{ForestGreen}{$\uparrow$}} & 54.74 & 69.69 \textcolor{ForestGreen}{$\uparrow$} & 66.97 & 72.98 \textcolor{ForestGreen}{$\uparrow$} & 67.71 & 69.42 \textcolor{ForestGreen}{$\uparrow$} & 66.44 & 72.32 \textcolor{ForestGreen}{$\uparrow$} \\
maggic (n=100) & 71.38 & 60.55 & 60.04 & \textbf{62.57 \textcolor{ForestGreen}{$\uparrow$}} & 56.98 & 60.49 \textcolor{ForestGreen}{$\uparrow$} & 58.21 & 60.42 \textcolor{ForestGreen}{$\uparrow$} & 58.11 & 59.42 \textcolor{ForestGreen}{$\uparrow$} & 49.71 & 58.77 \textcolor{ForestGreen}{$\uparrow$} & 55.36 & 60.58 \textcolor{ForestGreen}{$\uparrow$} & 60.02 & 60.15 \textcolor{ForestGreen}{$\uparrow$} & 56.47 & 60.28 \textcolor{ForestGreen}{$\uparrow$} \\
seer (n=100) & 91.80 & 85.52 & 86.81 & \textbf{87.64 \textcolor{ForestGreen}{$\uparrow$}} & 86.39 & 87.21 \textcolor{ForestGreen}{$\uparrow$} & 83.74 & 87.29 \textcolor{ForestGreen}{$\uparrow$} & 84.66 & 86.85 \textcolor{ForestGreen}{$\uparrow$} & 68.20 & 85.49 \textcolor{ForestGreen}{$\uparrow$} & 82.37 & 86.80 \textcolor{ForestGreen}{$\uparrow$} & 83.18 & 83.87 \textcolor{ForestGreen}{$\uparrow$} & 80.45 & 86.49 \textcolor{ForestGreen}{$\uparrow$} \\
compas (n=100) & 69.17 & 62.71 & \textbf{69.37} & 69.37 & 59.83 & 62.80 \textcolor{ForestGreen}{$\uparrow$} & 60.56 & 64.90 \textcolor{ForestGreen}{$\uparrow$} & 58.20 & 62.00 \textcolor{ForestGreen}{$\uparrow$} & 58.51 & 64.60 \textcolor{ForestGreen}{$\uparrow$} & 59.76 & 64.68 \textcolor{ForestGreen}{$\uparrow$} & 61.33 & 62.03 \textcolor{ForestGreen}{$\uparrow$} & 60.37 & 63.42 \textcolor{ForestGreen}{$\uparrow$} \\
adult (n=100) & 88.64 & 82.78 & 44.60 & 76.75 \textcolor{ForestGreen}{$\uparrow$} & 46.34 & 76.83 \textcolor{ForestGreen}{$\uparrow$} & 74.97 & 82.89 \textcolor{ForestGreen}{$\uparrow$} & 77.38 & 81.42 \textcolor{ForestGreen}{$\uparrow$} & 78.84 & \textbf{83.96 \textcolor{ForestGreen}{$\uparrow$}} & 68.31 & 82.87 \textcolor{ForestGreen}{$\uparrow$} & 82.58 & 83.24 \textcolor{ForestGreen}{$\uparrow$} & 76.15 & 82.47 \textcolor{ForestGreen}{$\uparrow$} \\
drug (n=100) & 81.06 & 75.49 & 79.12 & \textbf{79.81 \textcolor{ForestGreen}{$\uparrow$}} & 72.52 & 74.78 \textcolor{ForestGreen}{$\uparrow$} & 68.89 & 75.91 \textcolor{ForestGreen}{$\uparrow$} & 76.88 & 78.33 \textcolor{ForestGreen}{$\uparrow$} & 66.38 & 77.02 \textcolor{ForestGreen}{$\uparrow$} & 62.27 & 75.13 \textcolor{ForestGreen}{$\uparrow$} & 74.40 & 74.49 \textcolor{ForestGreen}{$\uparrow$} & 71.02 & 76.85 \textcolor{ForestGreen}{$\uparrow$} \\ \hline
covid (n=200) & 77.59 & 73.06 & 75.12 & \textbf{76.64 \textcolor{ForestGreen}{$\uparrow$}} & 71.74 & 74.66 \textcolor{ForestGreen}{$\uparrow$} & 72.57 & 75.50 \textcolor{ForestGreen}{$\uparrow$} & 73.56 & 75.88 \textcolor{ForestGreen}{$\uparrow$} & 66.02 & 74.51 \textcolor{ForestGreen}{$\uparrow$} & 63.29 & 74.62 \textcolor{ForestGreen}{$\uparrow$} & 73.34 & 73.89 \textcolor{ForestGreen}{$\uparrow$} & 68.14 & 74.21 \textcolor{ForestGreen}{$\uparrow$} \\
cutract (n=200) & 73.14 & 71.68 & 73.43 & 75.14 \textcolor{ForestGreen}{$\uparrow$} & 71.76 & 74.31 \textcolor{ForestGreen}{$\uparrow$} & 69.35 & 73.80 \textcolor{ForestGreen}{$\uparrow$} & 72.76 & \textbf{75.33 \textcolor{ForestGreen}{$\uparrow$}} & 65.83 & 73.43 \textcolor{ForestGreen}{$\uparrow$} & 68.05 & 74.27 \textcolor{ForestGreen}{$\uparrow$} & 70.06 & 71.35 \textcolor{ForestGreen}{$\uparrow$} & 66.65 & 73.34 \textcolor{ForestGreen}{$\uparrow$} \\
maggic (n=200) & 71.38 & 62.40 & 59.46 & \textbf{63.15 \textcolor{ForestGreen}{$\uparrow$}} & 58.05 & 61.68 \textcolor{ForestGreen}{$\uparrow$} & 59.04 & 62.58 \textcolor{ForestGreen}{$\uparrow$} & 57.63 & 59.23 \textcolor{ForestGreen}{$\uparrow$} & 50.63 & 60.09 \textcolor{ForestGreen}{$\uparrow$} & 55.55 & 61.36 \textcolor{ForestGreen}{$\uparrow$} & 61.73 & 61.88 \textcolor{ForestGreen}{$\uparrow$} & 57.09 & 61.15 \textcolor{ForestGreen}{$\uparrow$} \\
seer (n=200) & 91.80 & 85.87 & 86.75 & 88.08 \textcolor{ForestGreen}{$\uparrow$} & 85.74 & 86.97 \textcolor{ForestGreen}{$\uparrow$} & 83.97 & 87.64 \textcolor{ForestGreen}{$\uparrow$} & 86.21 & \textbf{88.46 \textcolor{ForestGreen}{$\uparrow$}} & 83.21 & 87.65 \textcolor{ForestGreen}{$\uparrow$} & 76.91 & 87.16 \textcolor{ForestGreen}{$\uparrow$} & 84.70 & 85.59 \textcolor{ForestGreen}{$\uparrow$} & 81.45 & 87.46 \textcolor{ForestGreen}{$\uparrow$} \\
compas (n=200) & 69.17 & 63.14 & 68.80 & \textbf{69.54 \textcolor{ForestGreen}{$\uparrow$}} & 61.62 & 64.59 \textcolor{ForestGreen}{$\uparrow$} & 60.65 & 65.67 \textcolor{ForestGreen}{$\uparrow$} & 60.97 & 65.12 \textcolor{ForestGreen}{$\uparrow$} & 60.66 & 65.67 \textcolor{ForestGreen}{$\uparrow$} & 56.49 & 64.34 \textcolor{ForestGreen}{$\uparrow$} & 61.68 & 62.93 \textcolor{ForestGreen}{$\uparrow$} & 60.32 & 64.18 \textcolor{ForestGreen}{$\uparrow$} \\
adult (n=200) & 88.64 & 84.12 & 40.38 & 77.39 \textcolor{ForestGreen}{$\uparrow$} & 47.40 & 79.40 \textcolor{ForestGreen}{$\uparrow$} & 80.00 & 84.61 \textcolor{ForestGreen}{$\uparrow$} & 83.73 & 85.73 \textcolor{ForestGreen}{$\uparrow$} & 81.92 & \textbf{85.76 \textcolor{ForestGreen}{$\uparrow$}} & 74.90 & 85.13 \textcolor{ForestGreen}{$\uparrow$} & 84.78 & 84.45 & 75.54 & 84.39 \textcolor{ForestGreen}{$\uparrow$} \\
drug (n=200) & 81.06 & 78.41 & 78.25 & 79.09 \textcolor{ForestGreen}{$\uparrow$} & 73.31 & 76.76 \textcolor{ForestGreen}{$\uparrow$} & 72.82 & 79.00 \textcolor{ForestGreen}{$\uparrow$} & 72.62 & 74.89 \textcolor{ForestGreen}{$\uparrow$} & 74.30 & \textbf{79.75 \textcolor{ForestGreen}{$\uparrow$}} & 63.91 & 77.60 \textcolor{ForestGreen}{$\uparrow$} & 77.64 & 77.92 \textcolor{ForestGreen}{$\uparrow$} & 70.16 & 77.89 \textcolor{ForestGreen}{$\uparrow$} \\ \hline
\bottomrule
\end{tabular}}
\label{table:results_xgb}
\vspace{20mm}
\end{table}
\endgroup

\begingroup
\setlength{\tabcolsep}{1.5pt} %
\renewcommand{\arraystretch}{1} %
\begin{table}[!t]
\vspace{10mm}
\caption{AUC for the DT model on $\Dtest$ where curation improve performance for all methods across all sample sizes $n$, as indicated by \textcolor{ForestGreen}{$\uparrow$}.}
\vspace{2mm}
\centering
\scalebox{0.575}{
\begin{tabular}{l|cc|cc|cc|cc|cc|cc|cc|cc|cc}
\toprule
& \multicolumn{2}{c|}{Real data} & \multicolumn{4}{c|}{\name ~(OURS)} &  \multicolumn{12}{c}{Baselines} \\
\cmidrule(lr){2-3}\cmidrule(lr){4-7}\cmidrule(lr){8-19}
& \multicolumn{2}{c|}{} & \multicolumn{2}{c}{GPT-4} & \multicolumn{2}{c|}{GPT-3.5} & \multicolumn{2}{c}{CTGAN} & \multicolumn{2}{c}{TabDDPM} & \multicolumn{2}{c}{GReaT}   & \multicolumn{2}{c}{NFLOW} & \multicolumn{2}{c}{SMOTE} & \multicolumn{2}{c}{TVAE} \\
\cmidrule(lr){2-3}\cmidrule(lr){4-5}\cmidrule(lr){6-7}\cmidrule(lr){8-9}\cmidrule(lr){10-11}\cmidrule(lr){12-13}\cmidrule(lr){14-15}\cmidrule(lr){16-17}\cmidrule(lr){18-19}
Dataset & $\Doracle$ & $\Dtrain$ & Uncur. & Cur. & Uncur. & Cur. & Uncur. & Cur. & Uncur. & Cur. & Uncur. & Cur. & Uncur. & Cur. & Uncur. & Cur. & Uncur. & Cur. \\
\hline\hline
covid (n=20) & 63.46 & 61.92 & 67.52 & \textbf{67.67 \textcolor{ForestGreen}{$\uparrow$}} & 61.07 & 64.89 \textcolor{ForestGreen}{$\uparrow$} & 52.94 & 56.62 \textcolor{ForestGreen}{$\uparrow$} & 58.95 & 59.19 \textcolor{ForestGreen}{$\uparrow$} & 51.83 & 58.73 \textcolor{ForestGreen}{$\uparrow$} & 54.37 & 61.53 \textcolor{ForestGreen}{$\uparrow$} & 59.56 & 59.01 & 56.46 & 59.49 \textcolor{ForestGreen}{$\uparrow$} \\
cutract (n=20) & 62.49 & 64.62 & 62.40 & \textbf{66.03 \textcolor{ForestGreen}{$\uparrow$}} & 62.10 & 65.15 \textcolor{ForestGreen}{$\uparrow$} & 55.67 & 61.73 \textcolor{ForestGreen}{$\uparrow$} & 57.91 & 58.97 \textcolor{ForestGreen}{$\uparrow$} & 51.89 & 62.76 \textcolor{ForestGreen}{$\uparrow$} & 57.12 & 64.85 \textcolor{ForestGreen}{$\uparrow$} & 62.85 & 62.38 & 60.25 & 62.34 \textcolor{ForestGreen}{$\uparrow$} \\
maggic (n=20) & 56.53 & 54.80 & 54.67 & \textbf{55.03 \textcolor{ForestGreen}{$\uparrow$}} & 52.54 & 53.54 \textcolor{ForestGreen}{$\uparrow$} & 50.90 & 52.08 \textcolor{ForestGreen}{$\uparrow$} & 52.39 & 52.70 \textcolor{ForestGreen}{$\uparrow$} & 49.70 & 53.08 \textcolor{ForestGreen}{$\uparrow$} & 51.52 & 53.23 \textcolor{ForestGreen}{$\uparrow$} & 52.55 & 52.44 & 50.91 & 51.52 \textcolor{ForestGreen}{$\uparrow$} \\
seer (n=20) & 78.86 & 71.09 & 76.17 & \textbf{77.17 \textcolor{ForestGreen}{$\uparrow$}} & 73.92 & 75.58 \textcolor{ForestGreen}{$\uparrow$} & 64.18 & 70.69 \textcolor{ForestGreen}{$\uparrow$} & 74.78 & 74.72 & 46.60 & 68.79 \textcolor{ForestGreen}{$\uparrow$} & 65.77 & 72.25 \textcolor{ForestGreen}{$\uparrow$} & 70.54 & 71.19 \textcolor{ForestGreen}{$\uparrow$} & 64.47 & 70.11 \textcolor{ForestGreen}{$\uparrow$} \\
compas (n=20) & 61.05 & 58.73 & 62.07 & \textbf{63.91 \textcolor{ForestGreen}{$\uparrow$}} & 57.17 & 60.85 \textcolor{ForestGreen}{$\uparrow$} & 53.83 & 58.83 \textcolor{ForestGreen}{$\uparrow$} & 51.77 & 56.11 \textcolor{ForestGreen}{$\uparrow$} & 52.39 & 60.96 \textcolor{ForestGreen}{$\uparrow$} & 54.64 & 60.27 \textcolor{ForestGreen}{$\uparrow$} & 58.16 & 57.57 & 53.52 & 58.97 \textcolor{ForestGreen}{$\uparrow$} \\
adult (n=20) & 73.84 & 68.03 & 52.80 & 67.10 \textcolor{ForestGreen}{$\uparrow$} & 49.15 & 66.52 \textcolor{ForestGreen}{$\uparrow$} & 61.57 & 68.93 \textcolor{ForestGreen}{$\uparrow$} & 70.46 & 71.19 \textcolor{ForestGreen}{$\uparrow$} & 62.04 & \textbf{71.87 \textcolor{ForestGreen}{$\uparrow$}} & 59.07 & 69.16 \textcolor{ForestGreen}{$\uparrow$} & 65.52 & 63.33 & 64.40 & 70.41 \textcolor{ForestGreen}{$\uparrow$} \\
drug (n=20) & 65.66 & 63.49 & 64.40 & \textbf{66.87 \textcolor{ForestGreen}{$\uparrow$}} & 62.69 & 64.21 \textcolor{ForestGreen}{$\uparrow$} & 56.76 & 64.18 \textcolor{ForestGreen}{$\uparrow$} & 61.04 & 62.48 \textcolor{ForestGreen}{$\uparrow$} & 52.22 & 62.62 \textcolor{ForestGreen}{$\uparrow$} & 53.50 & 61.30 \textcolor{ForestGreen}{$\uparrow$} & 62.34 & 60.69 & 55.00 & 58.89 \textcolor{ForestGreen}{$\uparrow$} \\ \hline
covid (n=40) & 63.46 & 65.10 & 66.55 & \textbf{67.85 \textcolor{ForestGreen}{$\uparrow$}} & 62.54 & 64.64 \textcolor{ForestGreen}{$\uparrow$} & 57.16 & 63.41 \textcolor{ForestGreen}{$\uparrow$} & 63.15 & 60.93 & 51.66 & 61.81 \textcolor{ForestGreen}{$\uparrow$} & 57.04 & 63.13 \textcolor{ForestGreen}{$\uparrow$} & 62.19 & 60.49 & 54.51 & 60.32 \textcolor{ForestGreen}{$\uparrow$} \\
cutract (n=40) & 62.49 & 63.01 & 62.49 & \textbf{65.29 \textcolor{ForestGreen}{$\uparrow$}} & 61.61 & 63.01 \textcolor{ForestGreen}{$\uparrow$} & 56.40 & 62.63 \textcolor{ForestGreen}{$\uparrow$} & 58.18 & 58.97 \textcolor{ForestGreen}{$\uparrow$} & 51.41 & 62.50 \textcolor{ForestGreen}{$\uparrow$} & 55.64 & 62.24 \textcolor{ForestGreen}{$\uparrow$} & 60.93 & 61.29 \textcolor{ForestGreen}{$\uparrow$} & 51.94 & 60.24 \textcolor{ForestGreen}{$\uparrow$} \\
maggic (n=40) & 56.53 & 53.74 & 53.20 & \textbf{54.71 \textcolor{ForestGreen}{$\uparrow$}} & 52.33 & 52.97 \textcolor{ForestGreen}{$\uparrow$} & 51.85 & 53.85 \textcolor{ForestGreen}{$\uparrow$} & 52.48 & 53.45 \textcolor{ForestGreen}{$\uparrow$} & 49.61 & 52.87 \textcolor{ForestGreen}{$\uparrow$} & 51.24 & 54.45 \textcolor{ForestGreen}{$\uparrow$} & 53.69 & 53.37 & 50.60 & 52.66 \textcolor{ForestGreen}{$\uparrow$} \\
seer (n=40) & 78.86 & 72.40 & 76.02 & 77.13 \textcolor{ForestGreen}{$\uparrow$} & 75.18 & \textbf{78.26 \textcolor{ForestGreen}{$\uparrow$}} & 68.31 & 73.89 \textcolor{ForestGreen}{$\uparrow$} & 77.94 & 73.24 & 51.45 & 72.32 \textcolor{ForestGreen}{$\uparrow$} & 67.78 & 75.95 \textcolor{ForestGreen}{$\uparrow$} & 70.32 & 72.81 \textcolor{ForestGreen}{$\uparrow$} & 68.79 & 74.48 \textcolor{ForestGreen}{$\uparrow$} \\
compas (n=40) & 61.05 & 58.23 & 62.00 & \textbf{63.43 \textcolor{ForestGreen}{$\uparrow$}} & 56.73 & 59.19 \textcolor{ForestGreen}{$\uparrow$} & 52.23 & 56.84 \textcolor{ForestGreen}{$\uparrow$} & 56.67 & 57.55 \textcolor{ForestGreen}{$\uparrow$} & 54.52 & 60.09 \textcolor{ForestGreen}{$\uparrow$} & 54.47 & 58.92 \textcolor{ForestGreen}{$\uparrow$} & 56.59 & 56.47 & 52.62 & 57.20 \textcolor{ForestGreen}{$\uparrow$} \\
adult (n=40) & 74.84 & 69.75 & 49.47 & 68.90 \textcolor{ForestGreen}{$\uparrow$} & 47.85 & 66.79 \textcolor{ForestGreen}{$\uparrow$} & 60.55 & 70.14 \textcolor{ForestGreen}{$\uparrow$} & 64.06 & 70.56 \textcolor{ForestGreen}{$\uparrow$} & 58.77 & 71.63 \textcolor{ForestGreen}{$\uparrow$} & 60.26 & 71.50 \textcolor{ForestGreen}{$\uparrow$} & 69.43 & 70.58 \textcolor{ForestGreen}{$\uparrow$} & 59.76 & \textbf{72.41 \textcolor{ForestGreen}{$\uparrow$}} \\
drug (n=40) & 65.66 & 64.18 & 62.86 & \textbf{65.90 \textcolor{ForestGreen}{$\uparrow$}} & 60.09 & 63.43 \textcolor{ForestGreen}{$\uparrow$} & 60.84 & 65.40 \textcolor{ForestGreen}{$\uparrow$} & 63.05 & 64.90 \textcolor{ForestGreen}{$\uparrow$} & 54.80 & 65.25 \textcolor{ForestGreen}{$\uparrow$} & 54.68 & 62.40 \textcolor{ForestGreen}{$\uparrow$} & 61.21 & 61.98 \textcolor{ForestGreen}{$\uparrow$} & 53.77 & 60.23 \textcolor{ForestGreen}{$\uparrow$} \\ \hline
covid (n=100) & 63.46 & 63.17 & 67.14 & \textbf{68.60 \textcolor{ForestGreen}{$\uparrow$}} & 62.40 & 66.02 \textcolor{ForestGreen}{$\uparrow$} & 61.34 & 64.63 \textcolor{ForestGreen}{$\uparrow$} & 63.33 & 66.42 \textcolor{ForestGreen}{$\uparrow$} & 55.86 & 64.53 \textcolor{ForestGreen}{$\uparrow$} & 55.44 & 64.91 \textcolor{ForestGreen}{$\uparrow$} & 61.96 & 63.40 \textcolor{ForestGreen}{$\uparrow$} & 59.75 & 64.47 \textcolor{ForestGreen}{$\uparrow$} \\
cutract (n=100) & 62.49 & 62.69 & 60.96 & 65.60 \textcolor{ForestGreen}{$\uparrow$} & 62.12 & 65.95 \textcolor{ForestGreen}{$\uparrow$} & 59.06 & \textbf{66.20 \textcolor{ForestGreen}{$\uparrow$}} & 63.38 & 65.72 \textcolor{ForestGreen}{$\uparrow$} & 52.62 & 61.73 \textcolor{ForestGreen}{$\uparrow$} & 58.11 & 65.08 \textcolor{ForestGreen}{$\uparrow$} & 60.15 & 61.34 \textcolor{ForestGreen}{$\uparrow$} & 58.82 & 63.76 \textcolor{ForestGreen}{$\uparrow$} \\
maggic (n=100) & 56.53 & 54.83 & 53.58 & \textbf{55.70 \textcolor{ForestGreen}{$\uparrow$}} & 50.72 & 53.98 \textcolor{ForestGreen}{$\uparrow$} & 52.29 & 55.06 \textcolor{ForestGreen}{$\uparrow$} & 53.93 & 54.61 \textcolor{ForestGreen}{$\uparrow$} & 49.71 & 53.34 \textcolor{ForestGreen}{$\uparrow$} & 51.82 & 54.53 \textcolor{ForestGreen}{$\uparrow$} & 54.11 & 54.34 \textcolor{ForestGreen}{$\uparrow$} & 51.31 & 53.33 \textcolor{ForestGreen}{$\uparrow$} \\
seer (n=100) & 78.86 & 73.83 & 76.50 & \textbf{77.84 \textcolor{ForestGreen}{$\uparrow$}} & 74.81 & 77.80 \textcolor{ForestGreen}{$\uparrow$} & 69.61 & 76.91 \textcolor{ForestGreen}{$\uparrow$} & 74.07 & 77.21 \textcolor{ForestGreen}{$\uparrow$} & 59.62 & 75.40 \textcolor{ForestGreen}{$\uparrow$} & 69.51 & 76.54 \textcolor{ForestGreen}{$\uparrow$} & 71.04 & 72.63 \textcolor{ForestGreen}{$\uparrow$} & 66.35 & 76.44 \textcolor{ForestGreen}{$\uparrow$} \\
compas (n=100) & 61.05 & 56.61 & 63.04 & \textbf{63.39 \textcolor{ForestGreen}{$\uparrow$}} & 55.57 & 57.40 \textcolor{ForestGreen}{$\uparrow$} & 55.07 & 58.95 \textcolor{ForestGreen}{$\uparrow$} & 53.38 & 55.69 \textcolor{ForestGreen}{$\uparrow$} & 54.61 & 59.31 \textcolor{ForestGreen}{$\uparrow$} & 54.38 & 57.92 \textcolor{ForestGreen}{$\uparrow$} & 56.69 & 57.01 \textcolor{ForestGreen}{$\uparrow$} & 54.34 & 56.72 \textcolor{ForestGreen}{$\uparrow$} \\
adult (n=100) & 73.84 & 72.19 & 46.68 & 66.58 \textcolor{ForestGreen}{$\uparrow$} & 47.94 & 67.37 \textcolor{ForestGreen}{$\uparrow$} & 62.42 & 71.78 \textcolor{ForestGreen}{$\uparrow$} & 68.06 & 70.77 \textcolor{ForestGreen}{$\uparrow$} & 66.29 & \textbf{74.36 \textcolor{ForestGreen}{$\uparrow$}} & 58.28 & 72.14 \textcolor{ForestGreen}{$\uparrow$} & 71.51 & 72.68 \textcolor{ForestGreen}{$\uparrow$} & 62.11 & 72.18 \textcolor{ForestGreen}{$\uparrow$} \\
drug (n=100) & 65.66 & 64.20 & 66.18 & \textbf{67.41 \textcolor{ForestGreen}{$\uparrow$}} & 60.77 & 61.21 \textcolor{ForestGreen}{$\uparrow$} & 57.67 & 63.41 \textcolor{ForestGreen}{$\uparrow$} & 63.26 & 64.55 \textcolor{ForestGreen}{$\uparrow$} & 54.89 & 65.72 \textcolor{ForestGreen}{$\uparrow$} & 52.06 & 63.38 \textcolor{ForestGreen}{$\uparrow$} & 61.21 & 62.30 \textcolor{ForestGreen}{$\uparrow$} & 55.40 & 63.32 \textcolor{ForestGreen}{$\uparrow$} \\ \hline
covid (n=200) & 63.46 & 63.45 & 67.19 & \textbf{68.74 \textcolor{ForestGreen}{$\uparrow$}} & 63.25 & 65.73 \textcolor{ForestGreen}{$\uparrow$} & 61.58 & 66.27 \textcolor{ForestGreen}{$\uparrow$} & 63.55 & 66.98 \textcolor{ForestGreen}{$\uparrow$} & 57.20 & 65.54 \textcolor{ForestGreen}{$\uparrow$} & 55.08 & 63.77 \textcolor{ForestGreen}{$\uparrow$} & 63.60 & 63.40 & 60.01 & 64.27 \textcolor{ForestGreen}{$\uparrow$} \\
cutract (n=200) & 62.49 & 63.53 & 63.36 & 65.33 \textcolor{ForestGreen}{$\uparrow$} & 63.01 & 65.47 \textcolor{ForestGreen}{$\uparrow$} & 60.51 & 65.01 \textcolor{ForestGreen}{$\uparrow$} & 62.55 & \textbf{66.94 \textcolor{ForestGreen}{$\uparrow$}} & 57.40 & 65.14 \textcolor{ForestGreen}{$\uparrow$} & 59.13 & 65.20 \textcolor{ForestGreen}{$\uparrow$} & 61.66 & 62.40 \textcolor{ForestGreen}{$\uparrow$} & 58.29 & 64.85 \textcolor{ForestGreen}{$\uparrow$} \\
maggic (n=200) & 56.53 & 55.06 & 52.76 & 54.41 \textcolor{ForestGreen}{$\uparrow$} & 52.41 & 54.34 \textcolor{ForestGreen}{$\uparrow$} & 52.47 & \textbf{55.21 \textcolor{ForestGreen}{$\uparrow$}} & 52.40 & 53.39 \textcolor{ForestGreen}{$\uparrow$} & 50.43 & 53.66 \textcolor{ForestGreen}{$\uparrow$} & 51.58 & 54.82 \textcolor{ForestGreen}{$\uparrow$} & 54.64 & 54.36 & 51.78 & 54.22 \textcolor{ForestGreen}{$\uparrow$} \\
seer (n=200) & 78.86 & 74.51 & 75.20 & 77.26 \textcolor{ForestGreen}{$\uparrow$} & 75.33 & 77.41 \textcolor{ForestGreen}{$\uparrow$} & 70.96 & 77.08 \textcolor{ForestGreen}{$\uparrow$} & 74.51 & \textbf{79.20 \textcolor{ForestGreen}{$\uparrow$}} & 71.44 & 78.66 \textcolor{ForestGreen}{$\uparrow$} & 65.84 & 76.60 \textcolor{ForestGreen}{$\uparrow$} & 73.11 & 74.77 \textcolor{ForestGreen}{$\uparrow$} & 66.37 & 76.36 \textcolor{ForestGreen}{$\uparrow$} \\
compas (n=200) & 61.05 & 57.20 & 61.09 & \textbf{62.89 \textcolor{ForestGreen}{$\uparrow$}} & 56.15 & 59.02 \textcolor{ForestGreen}{$\uparrow$} & 54.20 & 57.78 \textcolor{ForestGreen}{$\uparrow$} & 55.72 & 58.68 \textcolor{ForestGreen}{$\uparrow$} & 55.83 & 59.81 \textcolor{ForestGreen}{$\uparrow$} & 52.13 & 57.25 \textcolor{ForestGreen}{$\uparrow$} & 56.55 & 57.34 \textcolor{ForestGreen}{$\uparrow$} & 54.07 & 58.66 \textcolor{ForestGreen}{$\uparrow$} \\
adult (n=200) & 73.84 & 72.26 & 44.41 & 67.99 \textcolor{ForestGreen}{$\uparrow$} & 48.46 & 67.24 \textcolor{ForestGreen}{$\uparrow$} & 66.56 & 74.12 \textcolor{ForestGreen}{$\uparrow$} & 71.61 & 74.89 \textcolor{ForestGreen}{$\uparrow$} & 69.91 & \textbf{76.52 \textcolor{ForestGreen}{$\uparrow$}} & 60.70 & 74.05 \textcolor{ForestGreen}{$\uparrow$} & 73.11 & 74.08 \textcolor{ForestGreen}{$\uparrow$} & 63.10 & 73.82 \textcolor{ForestGreen}{$\uparrow$} \\
drug (n=200) & 65.66 & 67.04 & 66.29 & 66.57 \textcolor{ForestGreen}{$\uparrow$} & 61.88 & 64.34 \textcolor{ForestGreen}{$\uparrow$} & 59.10 & 66.06 \textcolor{ForestGreen}{$\uparrow$} & 63.59 & 63.40 & 59.11 & \textbf{69.04 \textcolor{ForestGreen}{$\uparrow$}} & 53.78 & 64.55 \textcolor{ForestGreen}{$\uparrow$} & 64.56 & 65.35 \textcolor{ForestGreen}{$\uparrow$} & 56.82 & 63.02 \textcolor{ForestGreen}{$\uparrow$} \\ \hline
\bottomrule
\end{tabular}}
\label{table:results_dt}
\end{table}
\endgroup

\begingroup
\setlength{\tabcolsep}{1.5pt} %
\renewcommand{\arraystretch}{1} %
\begin{table}[!t]

\caption{AUC for the LR model on $\Dtest$ where curation improves performance for all methods across all sample sizes $n$, as indicated by \textcolor{ForestGreen}{$\uparrow$}.}
\vspace{2mm}
\centering
\scalebox{0.575}{
\begin{tabular}{l|cc|cc|cc|cc|cc|cc|cc|cc|cc}
\toprule
& \multicolumn{2}{c|}{Real data} & \multicolumn{4}{c|}{\name ~(OURS)} &  \multicolumn{12}{c}{Baselines} \\
\cmidrule(lr){2-3}\cmidrule(lr){4-7}\cmidrule(lr){8-19}
& \multicolumn{2}{c|}{} & \multicolumn{2}{c}{GPT-4} & \multicolumn{2}{c|}{GPT-3.5} & \multicolumn{2}{c}{CTGAN} & \multicolumn{2}{c}{TabDDPM} & \multicolumn{2}{c}{GReaT}   & \multicolumn{2}{c}{NFLOW} & \multicolumn{2}{c}{SMOTE} & \multicolumn{2}{c}{TVAE} \\
\cmidrule(lr){2-3}\cmidrule(lr){4-5}\cmidrule(lr){6-7}\cmidrule(lr){8-9}\cmidrule(lr){10-11}\cmidrule(lr){12-13}\cmidrule(lr){14-15}\cmidrule(lr){16-17}\cmidrule(lr){18-19}
Dataset & $\Doracle$ & $\Dtrain$ & Uncur. & Cur. & Uncur. & Cur. & Uncur. & Cur. & Uncur. & Cur. & Uncur. & Cur. & Uncur. & Cur. & Uncur. & Cur. & Uncur. & Cur. \\
\hline\hline
covid (n=20) & 80.47 & 69.85 & 76.35 & \textbf{76.73 \textcolor{ForestGreen}{$\uparrow$}} & 74.69 & 75.18 \textcolor{ForestGreen}{$\uparrow$} & 59.99 & 66.87 \textcolor{ForestGreen}{$\uparrow$} & 68.78 & 69.84 \textcolor{ForestGreen}{$\uparrow$} & 63.94 & 71.81 \textcolor{ForestGreen}{$\uparrow$} & 69.28 & 72.32 \textcolor{ForestGreen}{$\uparrow$} & 68.36 & 68.44 \textcolor{ForestGreen}{$\uparrow$} & 63.92 & 70.12 \textcolor{ForestGreen}{$\uparrow$} \\
cutract (n=20) & 79.12 & 71.47 & \textbf{75.50} & 75.41 & 73.93 & 74.47 \textcolor{ForestGreen}{$\uparrow$} & 69.01 & 70.06 \textcolor{ForestGreen}{$\uparrow$} & 68.11 & 68.02 & 53.04 & 68.90 \textcolor{ForestGreen}{$\uparrow$} & 71.09 & 73.55 \textcolor{ForestGreen}{$\uparrow$} & 71.09 & 71.53 \textcolor{ForestGreen}{$\uparrow$} & 74.37 & 74.69 \textcolor{ForestGreen}{$\uparrow$} \\
maggic (n=20) & 70.46 & 56.55 & 63.67 & \textbf{64.12 \textcolor{ForestGreen}{$\uparrow$}} & 58.74 & 59.98 \textcolor{ForestGreen}{$\uparrow$} & 53.27 & 55.37 \textcolor{ForestGreen}{$\uparrow$} & 54.40 & 55.28 \textcolor{ForestGreen}{$\uparrow$} & 50.39 & 55.27 \textcolor{ForestGreen}{$\uparrow$} & 56.52 & 59.00 \textcolor{ForestGreen}{$\uparrow$} & 55.83 & 56.45 \textcolor{ForestGreen}{$\uparrow$} & 54.59 & 57.82 \textcolor{ForestGreen}{$\uparrow$} \\
seer (n=20) & 90.91 & 85.96 & 88.63 & \textbf{88.74 \textcolor{ForestGreen}{$\uparrow$}} & 87.90 & 88.11 \textcolor{ForestGreen}{$\uparrow$} & 78.22 & 83.69 \textcolor{ForestGreen}{$\uparrow$} & 83.17 & 84.74 \textcolor{ForestGreen}{$\uparrow$} & 51.07 & 77.84 \textcolor{ForestGreen}{$\uparrow$} & 84.86 & 84.75 & 85.72 & 86.11 \textcolor{ForestGreen}{$\uparrow$} & 80.33 & 82.63 \textcolor{ForestGreen}{$\uparrow$} \\
compas (n=20) & 73.02 & 63.74 & \textbf{71.88} & 71.10 & 66.62 & 67.89 \textcolor{ForestGreen}{$\uparrow$} & 55.62 & 64.29 \textcolor{ForestGreen}{$\uparrow$} & 62.15 & 64.70 \textcolor{ForestGreen}{$\uparrow$} & 56.97 & 66.59 \textcolor{ForestGreen}{$\uparrow$} & 64.71 & 68.59 \textcolor{ForestGreen}{$\uparrow$} & 62.68 & 63.16 \textcolor{ForestGreen}{$\uparrow$} & 61.19 & 64.36 \textcolor{ForestGreen}{$\uparrow$} \\
adult (n=20) & 88.37 & 79.24 & 45.28 & 71.07 \textcolor{ForestGreen}{$\uparrow$} & 50.89 & 75.77 \textcolor{ForestGreen}{$\uparrow$} & 80.03 & \textbf{81.27 \textcolor{ForestGreen}{$\uparrow$}} & 73.19 & 74.10 \textcolor{ForestGreen}{$\uparrow$} & 68.84 & 79.66 \textcolor{ForestGreen}{$\uparrow$} & 75.65 & 80.74 \textcolor{ForestGreen}{$\uparrow$} & 74.61 & 75.04 \textcolor{ForestGreen}{$\uparrow$} & 76.82 & 78.84 \textcolor{ForestGreen}{$\uparrow$} \\
drug (n=20) & 81.41 & 73.56 & \textbf{78.54} & 78.10 & 71.69 & 71.68 & 73.07 & 74.88 \textcolor{ForestGreen}{$\uparrow$} & 69.14 & 70.97 \textcolor{ForestGreen}{$\uparrow$} & 62.38 & 69.99 \textcolor{ForestGreen}{$\uparrow$} & 66.57 & 70.37 \textcolor{ForestGreen}{$\uparrow$} & 71.99 & 72.28 \textcolor{ForestGreen}{$\uparrow$} & 69.39 & 71.08 \textcolor{ForestGreen}{$\uparrow$} \\ \hline
covid (n=40) & 80.47 & 71.20 & 76.17 & \textbf{76.91 \textcolor{ForestGreen}{$\uparrow$}} & 75.32 & 76.13 \textcolor{ForestGreen}{$\uparrow$} & 64.56 & 68.82 \textcolor{ForestGreen}{$\uparrow$} & 71.08 & 72.77 \textcolor{ForestGreen}{$\uparrow$} & 61.40 & 71.21 \textcolor{ForestGreen}{$\uparrow$} & 70.74 & 73.84 \textcolor{ForestGreen}{$\uparrow$} & 68.85 & 69.65 \textcolor{ForestGreen}{$\uparrow$} & 64.13 & 69.58 \textcolor{ForestGreen}{$\uparrow$} \\
cutract (n=40) & 79.12 & 71.90 & 74.05 & \textbf{74.82 \textcolor{ForestGreen}{$\uparrow$}} & 72.69 & 72.89 \textcolor{ForestGreen}{$\uparrow$} & 68.43 & 70.97 \textcolor{ForestGreen}{$\uparrow$} & 70.67 & 72.39 \textcolor{ForestGreen}{$\uparrow$} & 61.50 & 74.10 \textcolor{ForestGreen}{$\uparrow$} & 67.98 & 71.55 \textcolor{ForestGreen}{$\uparrow$} & 70.26 & 70.12 & 67.87 & 70.88 \textcolor{ForestGreen}{$\uparrow$} \\
maggic (n=40) & 70.46 & 59.65 & 61.89 & \textbf{63.39 \textcolor{ForestGreen}{$\uparrow$}} & 60.41 & 61.84 \textcolor{ForestGreen}{$\uparrow$} & 55.81 & 56.64 \textcolor{ForestGreen}{$\uparrow$} & 56.58 & 58.08 \textcolor{ForestGreen}{$\uparrow$} & 48.68 & 58.82 \textcolor{ForestGreen}{$\uparrow$} & 57.50 & 60.90 \textcolor{ForestGreen}{$\uparrow$} & 57.98 & 58.00 \textcolor{ForestGreen}{$\uparrow$} & 56.61 & 59.42 \textcolor{ForestGreen}{$\uparrow$} \\
seer (n=40) & 90.91 & 87.86 & 88.50 & \textbf{88.89 \textcolor{ForestGreen}{$\uparrow$}} & 85.66 & 85.62 & 85.95 & 87.62 \textcolor{ForestGreen}{$\uparrow$} & 84.92 & 85.45 \textcolor{ForestGreen}{$\uparrow$} & 60.75 & 85.84 \textcolor{ForestGreen}{$\uparrow$} & 88.46 & 88.85 \textcolor{ForestGreen}{$\uparrow$} & 86.56 & 87.58 \textcolor{ForestGreen}{$\uparrow$} & 86.30 & 88.36 \textcolor{ForestGreen}{$\uparrow$} \\
compas (n=40) & 73.02 & 65.50 & \textbf{70.96} & 70.93 & 65.70 & 66.72 \textcolor{ForestGreen}{$\uparrow$} & 60.24 & 63.78 \textcolor{ForestGreen}{$\uparrow$} & 59.91 & 62.49 \textcolor{ForestGreen}{$\uparrow$} & 65.17 & 68.19 \textcolor{ForestGreen}{$\uparrow$} & 62.63 & 67.00 \textcolor{ForestGreen}{$\uparrow$} & 63.70 & 64.18 \textcolor{ForestGreen}{$\uparrow$} & 58.63 & 64.17 \textcolor{ForestGreen}{$\uparrow$} \\
adult (n=40) & 88.37 & 82.23 & 44.99 & 74.25 \textcolor{ForestGreen}{$\uparrow$} & 56.38 & 76.76 \textcolor{ForestGreen}{$\uparrow$} & 78.24 & 82.22 \textcolor{ForestGreen}{$\uparrow$} & 71.83 & 81.15 \textcolor{ForestGreen}{$\uparrow$} & 73.65 & 81.54 \textcolor{ForestGreen}{$\uparrow$} & 81.49 & \textbf{84.34 \textcolor{ForestGreen}{$\uparrow$}} & 81.80 & 83.16 \textcolor{ForestGreen}{$\uparrow$} & 80.54 & 83.02 \textcolor{ForestGreen}{$\uparrow$} \\
drug (n=40) & 81.41 & 71.74 & \textbf{78.98} & 78.88 & 71.54 & 73.21 \textcolor{ForestGreen}{$\uparrow$} & 71.34 & 74.44 \textcolor{ForestGreen}{$\uparrow$} & 72.12 & 73.68 \textcolor{ForestGreen}{$\uparrow$} & 70.00 & 76.45 \textcolor{ForestGreen}{$\uparrow$} & 66.91 & 73.78 \textcolor{ForestGreen}{$\uparrow$} & 69.09 & 69.75 \textcolor{ForestGreen}{$\uparrow$} & 68.21 & 70.94 \textcolor{ForestGreen}{$\uparrow$} \\ \hline
covid (n=100) & 80.47 & 74.19 & 76.53 & \textbf{77.34 \textcolor{ForestGreen}{$\uparrow$}} & 74.74 & 76.27 \textcolor{ForestGreen}{$\uparrow$} & 72.30 & 75.15 \textcolor{ForestGreen}{$\uparrow$} & 75.45 & 76.20 \textcolor{ForestGreen}{$\uparrow$} & 69.05 & 74.94 \textcolor{ForestGreen}{$\uparrow$} & 71.01 & 76.38 \textcolor{ForestGreen}{$\uparrow$} & 71.71 & 72.28 \textcolor{ForestGreen}{$\uparrow$} & 73.90 & 75.17 \textcolor{ForestGreen}{$\uparrow$} \\
cutract (n=100) & 79.12 & 76.73 & 74.62 & 75.70 \textcolor{ForestGreen}{$\uparrow$} & 73.75 & 74.40 \textcolor{ForestGreen}{$\uparrow$} & 73.89 & 75.61 \textcolor{ForestGreen}{$\uparrow$} & 75.70 & 74.49 & 60.03 & 74.35 \textcolor{ForestGreen}{$\uparrow$} & 75.49 & \textbf{76.86 \textcolor{ForestGreen}{$\uparrow$}} & 76.13 & 75.47 & 73.85 & 76.21 \textcolor{ForestGreen}{$\uparrow$} \\
maggic (n=100) & 70.46 & 60.14 & 61.31 & \textbf{63.64 \textcolor{ForestGreen}{$\uparrow$}} & 57.17 & 60.28 \textcolor{ForestGreen}{$\uparrow$} & 59.32 & 61.13 \textcolor{ForestGreen}{$\uparrow$} & 58.12 & 58.97 \textcolor{ForestGreen}{$\uparrow$} & 49.10 & 59.95 \textcolor{ForestGreen}{$\uparrow$} & 60.43 & 63.06 \textcolor{ForestGreen}{$\uparrow$} & 59.87 & 59.59 & 59.43 & 61.28 \textcolor{ForestGreen}{$\uparrow$} \\
seer (n=100) & 90.91 & 88.92 & 88.16 & 88.69 \textcolor{ForestGreen}{$\uparrow$} & 88.09 & 88.25 \textcolor{ForestGreen}{$\uparrow$} & 88.45 & 89.11 \textcolor{ForestGreen}{$\uparrow$} & 87.53 & 88.25 \textcolor{ForestGreen}{$\uparrow$} & 83.13 & 88.44 \textcolor{ForestGreen}{$\uparrow$} & 88.46 & \textbf{89.27 \textcolor{ForestGreen}{$\uparrow$}} & 87.95 & 87.65 & 86.81 & 88.67 \textcolor{ForestGreen}{$\uparrow$} \\
compas (n=100) & 73.02 & 67.75 & 71.21 & \textbf{71.34 \textcolor{ForestGreen}{$\uparrow$}} & 65.61 & 67.67 \textcolor{ForestGreen}{$\uparrow$} & 64.14 & 66.72 \textcolor{ForestGreen}{$\uparrow$} & 61.55 & 65.03 \textcolor{ForestGreen}{$\uparrow$} & 67.43 & 68.88 \textcolor{ForestGreen}{$\uparrow$} & 65.84 & 68.39 \textcolor{ForestGreen}{$\uparrow$} & 66.39 & 66.53 \textcolor{ForestGreen}{$\uparrow$} & 65.61 & 67.77 \textcolor{ForestGreen}{$\uparrow$} \\
adult (n=100) & 88.37 & 85.59 & 46.71 & 78.68 \textcolor{ForestGreen}{$\uparrow$} & 48.64 & 75.94 \textcolor{ForestGreen}{$\uparrow$} & 82.30 & 84.48 \textcolor{ForestGreen}{$\uparrow$} & 77.41 & 78.98 \textcolor{ForestGreen}{$\uparrow$} & 83.99 & 84.99 \textcolor{ForestGreen}{$\uparrow$} & 82.67 & \textbf{85.77 \textcolor{ForestGreen}{$\uparrow$}} & 85.63 & 85.43 & 81.32 & 83.97 \textcolor{ForestGreen}{$\uparrow$} \\
drug (n=100) & 81.41 & 74.34 & \textbf{80.35} & 80.01 & 70.00 & 71.76 \textcolor{ForestGreen}{$\uparrow$} & 71.66 & 75.80 \textcolor{ForestGreen}{$\uparrow$} & 71.26 & 72.40 \textcolor{ForestGreen}{$\uparrow$} & 74.57 & 78.65 \textcolor{ForestGreen}{$\uparrow$} & 69.34 & 77.50 \textcolor{ForestGreen}{$\uparrow$} & 71.04 & 71.75 \textcolor{ForestGreen}{$\uparrow$} & 71.91 & 75.38 \textcolor{ForestGreen}{$\uparrow$} \\ \hline
covid (n=200) & 80.47 & 76.73 & 76.06 & 77.25 \textcolor{ForestGreen}{$\uparrow$} & 75.18 & 77.09 \textcolor{ForestGreen}{$\uparrow$} & 76.11 & 77.54 \textcolor{ForestGreen}{$\uparrow$} & 77.51 & \textbf{78.09 \textcolor{ForestGreen}{$\uparrow$}} & 71.92 & 76.62 \textcolor{ForestGreen}{$\uparrow$} & 74.06 & 77.53 \textcolor{ForestGreen}{$\uparrow$} & 74.94 & 74.94 & 72.75 & 76.00 \textcolor{ForestGreen}{$\uparrow$} \\
cutract (n=200) & 79.12 & 77.53 & 75.28 & 76.30 \textcolor{ForestGreen}{$\uparrow$} & 74.36 & 75.38 \textcolor{ForestGreen}{$\uparrow$} & 75.25 & 75.88 \textcolor{ForestGreen}{$\uparrow$} & \textbf{77.66} & 77.66 & 75.37 & 76.69 \textcolor{ForestGreen}{$\uparrow$} & 76.45 & 77.50 \textcolor{ForestGreen}{$\uparrow$} & 76.77 & 76.59 & 72.94 & 74.86 \textcolor{ForestGreen}{$\uparrow$} \\
maggic (n=200) & 70.46 & 63.32 & 61.89 & \textbf{64.30 \textcolor{ForestGreen}{$\uparrow$}} & 59.10 & 62.10 \textcolor{ForestGreen}{$\uparrow$} & 61.57 & 63.98 \textcolor{ForestGreen}{$\uparrow$} & 56.77 & 57.35 \textcolor{ForestGreen}{$\uparrow$} & 51.05 & 63.17 \textcolor{ForestGreen}{$\uparrow$} & 59.17 & 63.84 \textcolor{ForestGreen}{$\uparrow$} & 62.86 & 62.86 & 60.94 & 63.47 \textcolor{ForestGreen}{$\uparrow$} \\
seer (n=200) & 90.91 & 90.01 & 88.65 & 89.20 \textcolor{ForestGreen}{$\uparrow$} & 87.57 & 88.14 \textcolor{ForestGreen}{$\uparrow$} & 88.20 & 88.80 \textcolor{ForestGreen}{$\uparrow$} & 89.69 & 89.81 \textcolor{ForestGreen}{$\uparrow$} & 89.10 & 89.68 \textcolor{ForestGreen}{$\uparrow$} & 87.35 & 89.54 \textcolor{ForestGreen}{$\uparrow$} & 89.13 & 89.14 \textcolor{ForestGreen}{$\uparrow$} & 88.74 & 89.65 \textcolor{ForestGreen}{$\uparrow$} \\
compas (n=200) & 73.02 & 69.14 & 70.53 & \textbf{71.36 \textcolor{ForestGreen}{$\uparrow$}} & 63.49 & 66.50 \textcolor{ForestGreen}{$\uparrow$} & 66.26 & 68.73 \textcolor{ForestGreen}{$\uparrow$} & 64.83 & 66.40 \textcolor{ForestGreen}{$\uparrow$} & 67.85 & 69.93 \textcolor{ForestGreen}{$\uparrow$} & 60.78 & 68.07 \textcolor{ForestGreen}{$\uparrow$} & 68.18 & 68.18 & 66.26 & 68.99 \textcolor{ForestGreen}{$\uparrow$} \\
adult (n=200) & 88.37 & 86.94 & 38.89 & 80.35 \textcolor{ForestGreen}{$\uparrow$} & 56.97 & 79.07 \textcolor{ForestGreen}{$\uparrow$} & 85.25 & 86.29 \textcolor{ForestGreen}{$\uparrow$} & 86.81 & 86.63 & 85.70 & 85.94 \textcolor{ForestGreen}{$\uparrow$} & 84.18 & 86.65 \textcolor{ForestGreen}{$\uparrow$} & 86.88 & 86.70 & 84.09 & 85.91 \textcolor{ForestGreen}{$\uparrow$} \\
drug (n=200) & 81.41 & 77.47 & 79.06 & 79.61 \textcolor{ForestGreen}{$\uparrow$} & 71.56 & 72.54 \textcolor{ForestGreen}{$\uparrow$} & 76.15 & 78.84 \textcolor{ForestGreen}{$\uparrow$} & 69.00 & 73.14 \textcolor{ForestGreen}{$\uparrow$} & 77.51 & 79.72 \textcolor{ForestGreen}{$\uparrow$} & 75.47 & \textbf{80.07 \textcolor{ForestGreen}{$\uparrow$}} & 75.82 & 75.69 & 71.76 & 76.95 \textcolor{ForestGreen}{$\uparrow$} \\ \hline
\bottomrule
\end{tabular}}
\label{table:results_lr}
\end{table}
\endgroup

\newpage
\clearpage
\subsection{Full results for performance evaluation}

\textcolor{TealBlue}{We report full results with standard deviation for the results from the main paper. The performance is AUC averaged over XGBoost, Random forest, Logistic regression, Decision tree.}

\begingroup
\setlength{\tabcolsep}{1.5pt} %
\renewcommand{\arraystretch}{1} %
\begin{table}[!h]
\vspace{0mm}
\caption{\textcolor{TealBlue}{AUC averaged over 4 downstream models on $\Dtest$ where curation improves performance for all methods across all sample sizes $n$, as indicated by \textcolor{ForestGreen}{$\uparrow$}. \name~ w/ GPT-4 (Curated) dataset provides the strongest performance for both \colorbox{orange!20} {private/proprietary datasets} and \colorbox{green!20} {public datasets}}}
\vspace{2mm}
\centering
\scalebox{0.5}{
\begin{tabular}{l|cc|cc|cc|cc|cc|cc|cc|cc|cc}
\toprule
& \multicolumn{2}{c|}{Real data} & \multicolumn{4}{c|}{\name ~(OURS)} &  \multicolumn{12}{c}{Baselines} \\
\cmidrule(lr){2-3}\cmidrule(lr){4-7}\cmidrule(lr){8-19}
& \multicolumn{2}{c|}{} & \multicolumn{2}{c}{GPT-4} & \multicolumn{2}{c|}{GPT-3.5} & \multicolumn{2}{c}{CTGAN} & \multicolumn{2}{c}{TabDDPM} & \multicolumn{2}{c}{GReaT}   & \multicolumn{2}{c}{NFLOW} & \multicolumn{2}{c}{SMOTE} & \multicolumn{2}{c}{TVAE} \\
\cmidrule(lr){2-3}\cmidrule(lr){4-5}\cmidrule(lr){6-7}\cmidrule(lr){8-9}\cmidrule(lr){10-11}\cmidrule(lr){12-13}\cmidrule(lr){14-15}\cmidrule(lr){16-17}\cmidrule(lr){18-19}
Dataset & $\Doracle$ & $\Dtrain$ & Uncur. & Cur. & Uncur. & Cur. & Uncur. & Cur. & Uncur. & Cur. & Uncur. & Cur. & Uncur. & Cur. & Uncur. & Cur. & Uncur. & Cur. \\
\hline\hline
covid (n=20) & $74.41_{(0.11)}$ & $68.50_{(1.57)}$ & $73.78_{(0.31)}$ & $\textbf{73.87}_{(0.50)}$ & $69.85_{(0.75)}$ & $71.41_{(0.92)}$ & $59.00_{(2.25)}$ & $63.67_{(2.51)}$ & $66.84_{(1.66)}$ & $66.85_{(1.56)}$ & $57.38_{(1.47)}$ & $66.46_{(0.80)}$ & $62.87_{(0.98)}$ & $68.56_{(1.07)}$ & $66.95_{(1.66)}$ & $66.82_{(1.89)}$ & $61.69_{(2.72)}$ & $66.11_{(2.79)}$ \\
cutract (n=20) & $72.23_{(0.65)}$ & $70.12_{(1.16)}$ & $71.15_{(0.46)}$ & $\textbf{72.50}_{(0.76)}$ & $69.97_{(1.13)}$ & $71.54_{(1.38)}$ & $64.01_{(2.26)}$ & $67.98_{(1.63)}$ & $66.05_{(1.14)}$ & $66.59_{(1.49)}$ & $52.38_{(1.78)}$ & $67.02_{(0.97)}$ & $64.44_{(1.05)}$ & $70.42_{(1.25)}$ & $68.41_{(1.34)}$ & $69.24_{(1.25)}$ & $68.94_{(1.38)}$ & $70.22_{(1.12)}$ \\
maggic (n=20) & $67.41_{(0.06)}$ & $57.13_{(0.85)}$ & $60.70_{(0.38)}$ & $\textbf{61.48}_{(0.49)}$ & $57.54_{(0.83)}$ & $58.69_{(0.76)}$ & $52.75_{(1.34)}$ & $54.51_{(1.40)}$ & $54.59_{(1.02)}$ & $55.39_{(1.07)}$ & $50.29_{(0.52)}$ & $55.64_{(0.43)}$ & $54.72_{(1.38)}$ & $57.38_{(1.06)}$ & $55.84_{(1.21)}$ & $56.15_{(1.20)}$ & $54.08_{(1.05)}$ & $56.19_{(0.76)}$ \\
seer (n=20) & $87.92_{(0.08)}$ & $80.67_{(1.67)}$ & $84.53_{(0.33)}$ & $\textbf{84.82}_{(0.47)}$ & $83.34_{(0.95)}$ & $83.71_{(0.52)}$ & $74.34_{(4.11)}$ & $78.73_{(3.13)}$ & $80.59_{(1.32)}$ & $80.60_{(1.32)}$ & $47.57_{(2.51)}$ & $74.43_{(3.36)}$ & $76.06_{(1.68)}$ & $79.98_{(1.65)}$ & $79.23_{(2.17)}$ & $80.02_{(2.08)}$ & $74.53_{(1.37)}$ & $78.73_{(1.24)}$ \\
compas (n=20) & $67.51_{(0.03)}$ & $63.11_{(0.96)}$ & $\textbf{68.01}_{(0.44)}$ & $67.91_{(0.55)}$ & $62.07_{(1.51)}$ & $64.43_{(1.38)}$ & $55.67_{(1.60)}$ & $62.56_{(1.61)}$ & $57.67_{(1.90)}$ & $60.87_{(1.33)}$ & $53.33_{(1.85)}$ & $63.59_{(1.30)}$ & $59.49_{(1.72)}$ & $64.62_{(1.14)}$ & $61.06_{(1.92)}$ & $61.59_{(2.15)}$ & $58.30_{(1.77)}$ & $62.58_{(1.87)}$ \\
adult (n=20) & $84.17_{(0.10)}$ & $77.45_{(1.25)}$ & $50.39_{(3.99)}$ & $71.48_{(2.34)}$ & $49.23_{(2.57)}$ & $72.37_{(2.26)}$ & $72.23_{(1.26)}$ & $76.86_{(1.25)}$ & $74.35_{(1.48)}$ & $75.04_{(1.82)}$ & $67.00_{(4.72)}$ & $77.25_{(1.49)}$ & $67.46_{(3.76)}$ & $76.48_{(1.77)}$ & $73.75_{(1.64)}$ & $73.67_{(1.46)}$ & $73.20_{(1.51)}$ & $76.90_{(1.64)}$ \\
drug (n=20) & $77.81_{(0.55)}$ & $70.84_{(2.25)}$ & $75.08_{(1.17)}$ & $\textbf{75.29}_{(1.11)}$ & $71.68_{(2.25)}$ & $72.14_{(2.64)}$ & $68.31_{(2.81)}$ & $72.65_{(2.00)}$ & $68.12_{(2.38)}$ & $69.68_{(2.41)}$ & $58.78_{(4.26)}$ & $68.89_{(3.57)}$ & $62.13_{(4.94)}$ & $67.75_{(3.72)}$ & $70.16_{(1.87)}$ & $70.16_{(1.75)}$ & $66.60_{(2.95)}$ & $69.18_{(2.59)}$ \\ \hline
covid (n=40) & $74.41_{(0.11)}$ & $70.77_{(0.96)}$ & $73.40_{(0.61)}$ & $\textbf{73.95}_{(0.67)}$ & $70.42_{(0.92)}$ & $71.93_{(0.60)}$ & $63.63_{(1.15)}$ & $68.46_{(0.93)}$ & $70.50_{(1.49)}$ & $70.44_{(1.37)}$ & $56.50_{(0.83)}$ & $68.68_{(1.39)}$ & $66.41_{(2.51)}$ & $70.48_{(1.48)}$ & $68.66_{(1.48)}$ & $68.44_{(1.28)}$ & $61.03_{(0.96)}$ & $67.35_{(0.60)}$ \\ 
cutract (n=40) & $72.23_{(0.65)}$ & $69.18_{(0.65)}$ & $69.87_{(0.62)}$ & $\textbf{71.72}_{(0.46)}$ & $68.47_{(0.59)}$ & $69.56_{(0.50)}$ & $63.01_{(2.33)}$ & $67.87_{(1.36)}$ & $65.63_{(2.71)}$ & $67.27_{(2.38)}$ & $54.39_{(1.59)}$ & $68.44_{(0.41)}$ & $61.40_{(1.95)}$ & $67.98_{(1.06)}$ & $67.86_{(0.72)}$ & $67.95_{(0.59)}$ & $59.79_{(1.85)}$ & $66.62_{(1.06)}$ \\
maggic (n=40) & $67.50_{(0.04)}$ & $58.26_{(0.55)}$ & $59.29_{(0.50)}$ & $\textbf{60.77}_{(0.36)}$ & $57.50_{(0.69)}$ & $59.15_{(0.48)}$ & $55.00_{(1.27)}$ & $56.78_{(1.17)}$ & $55.24_{(0.63)}$ & $56.94_{(0.50)}$ & $48.81_{(0.73)}$ & $56.64_{(0.63)}$ & $54.68_{(0.53)}$ & $58.58_{(0.54)}$ & $57.40_{(0.57)}$ & $57.44_{(0.67)}$ & $55.04_{(1.14)}$ & $57.33_{(1.06)}$ \\
seer (n=40) & $87.92_{(0.08)}$ & $82.93_{(0.55)}$ & $84.29_{(0.39)}$ & $\textbf{84.93}_{(0.46)}$ & $83.46_{(1.20)}$ & $84.44_{(0.69)}$ & $80.05_{(0.93)}$ & $83.67_{(0.51)}$ & $82.59_{(1.48)}$ & $81.37_{(1.11)}$ & $54.93_{(2.06)}$ & $81.11_{(1.24)}$ & $79.88_{(0.53)}$ & $84.36_{(0.62)}$ & $80.79_{(0.67)}$ & $82.21_{(0.67)}$ & $78.69_{(2.32)}$ & $83.62_{(1.06)}$ \\
compas (n=40) & $67.51_{(0.03)}$ & $62.34_{(0.79)}$ & $67.57_{(0.40)}$ & $\textbf{67.85}_{(0.37)}$ & $61.34_{(1.67)}$ & $62.84_{(1.21)}$ & $56.29_{(1.96)}$ & $61.02_{(1.69)}$ & $58.85_{(1.28)}$ & $60.11_{(1.22)}$ & $58.88_{(1.04)}$ & $64.37_{(0.91)}$ & $58.61_{(1.35)}$ & $63.54_{(1.11)}$ & $60.83_{(1.19)}$ & $60.95_{(1.22)}$ & $55.94_{(1.50)}$ & $61.04_{(1.56)}$ \\
adult (n=40) & $84.17_{(0.10)}$ & $79.44_{(1.03)}$ & $48.31_{(3.58)}$ & $73.82_{(2.19)}$ & $49.21_{(1.78)}$ & $74.27_{(1.49)}$ & $71.82_{(1.22)}$ & $79.11_{(0.71)}$ & $71.51_{(1.85)}$ & $77.99_{(0.65)}$ & $66.77_{(3.32)}$ & $78.81_{(1.36)}$ & $71.13_{(2.27)}$ & $79.71_{(1.02)}$ & $77.90_{(1.07)}$ & $78.84_{(1.19)}$ & $72.58_{(1.34)}$ & $\textbf{80.02}_{(0.76)}$ \\
drug (n=40) & $77.81_{(0.55)}$ & $71.86_{(1.07)}$ & $74.30_{(0.59)}$ & $\textbf{75.79}_{(0.39)}$ & $71.33_{(0.88)}$ & $72.76_{(0.97)}$ & $69.46_{(2.23)}$ & $72.74_{(1.70)}$ & $71.08_{(1.72)}$ & $73.07_{(1.03)}$ & $64.89_{(1.39)}$ & $73.64_{(0.87)}$ & $62.51_{(3.03)}$ & $70.97_{(1.91)}$ & $69.23_{(1.68)}$ & $69.78_{(1.46)}$ & $65.22_{(1.37)}$ & $70.30_{(1.04)}$ \\ \hline
covid (n=100) & $74.41_{(0.11)}$ & $71.57_{(0.48)}$ & $73.77_{(0.27)}$ & $\textbf{74.71}_{(0.34)}$ & $70.71_{(0.46)}$ & $72.76_{(0.44)}$ & $69.05_{(0.96)}$ & $72.13_{(0.66)}$ & $71.60_{(0.59)}$ & $73.22_{(0.46)}$ & $63.52_{(1.29)}$ & $72.04_{(0.57)}$ & $64.25_{(1.51)}$ & $72.64_{(0.64)}$ & $70.08_{(0.67)}$ & $70.78_{(0.59)}$ & $69.05_{(0.48)}$ & $71.96_{(0.49)}$ \\
cutract (n=100) & $72.23_{(0.65)}$ & $70.96_{(0.68)}$ & $70.20_{(0.45)}$ & $\textbf{72.51}_{(0.55)}$ & $69.97_{(0.97)}$ & $71.94_{(0.89)}$ & $67.94_{(1.01)}$ & $72.42_{(0.66)}$ & $70.53_{(1.51)}$ & $71.98_{(1.39)}$ & $55.72_{(2.04)}$ & $69.14_{(0.93)}$ & $67.59_{(0.71)}$ & $72.42_{(0.54)}$ & $68.79_{(0.83)}$ & $69.68_{(0.76)}$ & $66.89_{(1.03)}$ & $71.52_{(0.70)}$ \\
maggic (n=100) & $67.50_{(0.04)}$ & $59.65_{(0.50)}$ & $58.98_{(0.23)}$ & $\textbf{61.32}_{(0.42)}$ & $55.71_{(0.83)}$ & $58.90_{(0.72)}$ & $57.20_{(0.91)}$ & $59.34_{(0.64)}$ & $57.26_{(0.50)}$ & $58.28_{(0.46)}$ & $49.54_{(0.71)}$ & $57.91_{(0.74)}$ & $56.36_{(0.54)}$ & $60.11_{(0.54)}$ & $58.89_{(0.51)}$ & $58.99_{(0.42)}$ & $56.17_{(0.68)}$ & $58.86_{(0.69)}$ \\
seer (n=100) & $87.92_{(0.08)}$ & $83.95_{(0.32)}$ & $84.45_{(0.38)}$ & $\textbf{85.37}_{(0.47)}$ & $83.92_{(0.81)}$ & $85.08_{(0.32)}$ & $81.60_{(0.73)}$ & $85.14_{(0.36)}$ & $83.04_{(0.78)}$ & $84.83_{(0.49)}$ & $70.32_{(2.52)}$ & $83.83_{(0.39)}$ & $81.16_{(1.03)}$ & $85.03_{(0.39)}$ & $81.82_{(0.45)}$ & $82.49_{(0.46)}$ & $78.88_{(0.71)}$ & $84.50_{(0.44)}$ \\
compas (n=100) & $67.51_{(0.03)}$ & $62.56_{(0.72)}$ & $68.02_{(0.29)}$ & $\textbf{68.19}_{(0.37)}$ & $60.10_{(1.60)}$ & $62.47_{(1.11)}$ & $60.01_{(1.16)}$ & $63.73_{(0.99)}$ & $58.32_{(0.91)}$ & $61.34_{(0.94)}$ & $59.97_{(0.82)}$ & $64.19_{(0.83)}$ & $60.02_{(0.63)}$ & $64.04_{(0.60)}$ & $61.44_{(0.84)}$ & $61.73_{(0.81)}$ & $59.97_{(1.00)}$ & $62.82_{(1.05)}$ \\
adult (n=100) & $84.17_{(0.10)}$ & $81.24_{(0.48)}$ & $46.09_{(1.86)}$ & $74.57_{(1.74)}$ & $47.56_{(3.43)}$ & $73.97_{(1.57)}$ & $74.29_{(1.23)}$ & $80.45_{(0.81)}$ & $75.93_{(1.45)}$ & $78.22_{(1.22)}$ & $77.09_{(0.74)}$ & $\textbf{81.66}_{(0.53)}$ & $70.70_({1.02)}$ & $81.04_{(0.37)}$ & $80.56_{(0.53)}$ & $81.10_{(0.44)}$ & $74.04_{(1.44)}$ & $80.23_{(0.98)}$ \\
drug (n=100) & $77.81_{(0.55)}$ & $73.58_{(0.72)}$ & $76.24_{(0.64)}$ & $\textbf{76.74}_{(0.51)}$ & $69.46_{(2.43)}$ & $71.05_{(2.30)}$ & $68.19_{(2.06)}$ & $73.28_{(1.41)}$ & $72.43_{(0.96)}$ & $73.79_{(0.65)}$ & $67.26_{(1.22)}$ & $75.28_{(0.46)}$ & $62.67_{(2.23)}$ & $73.12_{(0.97)}$ & $70.90_{(1.08)}$ & $71.53_{(1.13)}$ & $68.22_{(1.60)}$ & $73.59_{(0.97)}$ \\ \hline
covid (n=200) & $74.41_{(0.11)}$ & $72.33_{(0.52)}$ & $73.40_{(0.25)}$ & $\textbf{74.62}_{(0.13)}$ & $70.70_{(0.73)}$ & $73.12_{(0.48)}$ & $71.07_{(0.35)}$ & $73.89_{(0.40)}$ & $72.47_{(0.50)}$ & $74.44_{(0.44)}$ & $65.55_{(0.67)}$ & $73.07_{(0.38)}$ & $65.04_{(0.69)}$ & $72.90_{(0.37)}$ & $71.68_{(0.49)}$ & $71.87_{(0.49)}$ & $67.89_{(0.56)}$ & $72.38_{(0.42)}$ \\
cutract (n=200) & $72.23_{(0.65)}$ & $71.75_{(0.71)}$ & $71.39_{(0.76)}$ & $73.01_{(0.85)}$ & $70.28_{(0.67)}$ & $72.39_{(0.84)}$ & $69.28_{(0.55)}$ & $72.41_{(0.65)}$ & $71.83_{(0.65)}$ & $\textbf{74.03}_{(0.69)}$ & $66.66_{(0.94)}$ & $72.49_{(0.71)}$ & $68.77_{(0.71)}$ & $73.16_{(0.68)}$ & $70.23_{(0.83)}$ & $70.80_{(0.79)}$ & $66.61_{(0.76)}$ & $71.87_{(0.71)}$ \\
maggic (n=200) & $67.50_{(0.04)}$ & $61.39_{(0.44)}$ & $58.92_{(0.47)}$ & $\textbf{61.41}_{(0.39)}$ & $57.33_{(0.51)}$ & $60.16_{(0.41)}$ & $58.48_{(0.59)}$ & $61.33_{(0.57)}$ & $56.26_{(0.82)}$ & $57.20_{(0.91)}$ & $50.74_{(0.74)}$ & $59.60_{(0.55)}$ & $55.95_{(0.72)}$ & $60.75_{(0.39)}$ & $60.73_{(0.44)}$ & $60.78_{(0.49)}$ & $57.18_{(0.56)}$ & $60.23_{(0.44)}$ \\
seer (n=200) & $87.92_{(0.08)}$ & $84.63_{(0.35)}$ & $84.39_{(0.19)}$ & $85.56_{(0.30)}$ & $83.48_{(0.41)}$ & $84.80_{(0.34)}$ & $82.04_{(0.70)}$ & $85.34_{(0.36)}$ & $84.39_{(0.44)}$ & $\textbf{86.57}_{(0.27)}$ & $82.15_{(0.43)}$ & $86.03_{(0.23)}$ & $77.73_{(1.65)}$ & $85.19_{(0.22)}$ & $83.38_{(0.32)}$ & $84.15_{(0.42)}$ & $79.71_{(0.58)}$ & $85.26_{(0.25)}$ \\
compas (n=200) & $67.51_{(0.03)}$ & $63.27_{(0.60)}$ & $67.02_{(0.53)}$ & $\textbf{68.15}_{(0.40)}$ & $60.48_{(1.86)}$ & $63.39_{(1.28)}$ & $60.58_{(0.83)}$ & $64.32_{(0.78)}$ & $60.60_{(0.90)}$ & $63.52_{(1.04)}$ & $61.11_{(0.63)}$ & $65.08_{(0.53)}$ & $56.58_{(1.03)}$ & $63.60_{(0.83)}$ & $61.99_{(0.64)}$ & $62.80_{(0.61)}$ & $60.15_{(1.14)}$ & $63.99_{(0.71)}$ \\
adult (n=200) & $84.17_{(0.10)}$ & $82.12_{(0.41)}$ & $40.96_{(2.47)}$ & $75.84_{(1.43)}$ & $49.89_{(3.22)}$ & $76.11_{(1.43)}$ & $78.18_{(0.26)}$ & $82.32_{(0.29)}$ & $81.66_{(0.19)}$ & $83.17_{(0.20)}$ & $80.06_{(0.53)}$ & $\textbf{83.32}_{(0.35)}$ & $74.31_{(0.81)}$ & $82.64_{(0.33)}$ & $82.26_{(0.38)}$ & $82.39_{(0.33)}$ & $75.21_{(0.69)}$ & $82.02_{(0.25)}$ \\
drug (n=200) & $77.81_{(0.55)}$ & $76.10_{(0.45)}$ & $75.58_{(0.55)}$ & $76.06_{(0.42)}$ & $70.66_{(1.56)}$ & $72.81_{(1.18)}$ & $71.31_{(1.17)}$ & $75.98_{(0.76)}$ & $69.61_{(2.76)}$ & $71.79_{(1.99)}$ & $72.35_{(0.61)}$ & $\textbf{77.41}_{(0.56)}$ & $65.25_{(2.47)}$ & $75.26_{(0.77)}$ & $74.38_{(0.50)}$ & $74.78_{(0.54)}$ & $68.39_{(1.26)}$ & $74.33_{(0.61)}$ \\
\bottomrule
\end{tabular}}
\label{table:results_comparison2-full}
\vspace{-3mm}
\end{table}
\endgroup

\subsection{\name~ with context in low-resource languages}

We investigate how the language of tabular column names affect the performance of \name. To do so,  for a given dataset, we translate the feature names from English to 2 low-resource languages \cite{ranathunga2023neural}, i.e. to (i) \textit{Swahili} and (ii) \textit{Hausa}. Changing the language of the features alters the contextual information provided to the LLM to generate synthetic data.

We report the results in  \cref{tab:low-resource}. As expected, we observe for both Swahili and Hausa a performance drop compared to using English feature names, thereby mirroring the results in \cref{sec:generate} on the importance of contextual information in the prompt. However, we note that \name~ for both Swahili and Hausa remains highly competitive compared to other baselines, notably thanks to the curation mechanism.

\begin{table}[!h]
\centering
\caption{Test AUC in settings where the names of features have been translated to Swahili and Hausa, with and without curation}
\begin{tabular}{l|cc|cc|cc}
\hline
Dataset  & \makecell{Uncurated \\ \textit{English}} & \makecell{Curated \\ \textit{English}} & \makecell{Uncurated \\ \textit{Swahili}} & \makecell{Curated \\ \textit{Swahili}} & \makecell{Uncurated \\ \textit{Hausa}} & \makecell{Curated \\ \textit{Hausa}} \\ \hline
covid  ($n=20$) & 73.78 & 73.87 & 71.32 & 73.06 & 68.75 & 70.01 \\ 
seer  ($n=20$) & 84.53 & 84.82 & 83.92 & 85.47 & 81.57 & 82.73 \\ 
compas ($n=20$) & 68.01 & 67.91 & 64.42 & 61.61 & 62.85 & 63.06 \\ \hline
covid  $n=40$ & 73.40 & 73.95 & 71.53 & 73.20 & 70.15 & 72.13 \\ 
seer   $n=40$ & 84.29 & 84.93 & 79.71 & 84.25 & 81.47 & 84.66 \\ 
compas $n=40$ & 67.57 & 67.85 & 64.92 & 65.78 & 62.31 & 63.20 \\ \hline
\end{tabular}

\label{tab:low-resource}
\end{table}

\textbf{Takeaways:}

\begin{enumerate}
    \item \name~ can provide valuable augmentation and performance gains even with feature names in low-resource languages. 
    \item For optimal performance, we recommend using \name~ with English feature names or translating the feature names to English before applying \name~.
\end{enumerate}

\subsection{Comparison of curation mechanism vs Data-IQ}

We compare \name's curation mechanism with the approach taken by Data-IQ \cite{seedat2022data}.

They differ along four dimensions:

\begin{enumerate}
    \item  \textbf{Problem setting} (data augmentation vs data understanding): \name~ focuses on the problem of data augmentation in low-data regimes by curating synthetic samples. In contrast, Data-IQ aims to understand and characterize subgroups within a given real dataset.
\item \textbf{Conceptually}: \name~ curates a large synthetic dataset $D_{\mathrm{syn}}$ with respect to a very small gold standard real dataset $D_{\mathrm{train}}$, whereas Data-IQ aims to characterize subgroups of learnable samples in a single large real dataset $D$ (e.g. to find the hard samples in $D$).
\item \textbf{Technically}: \name~ trains the curation model only on the small but gold standard $D_{\mathrm{train}}$. The learning dynamics of the synthetic samples in $D_{\mathrm{syn}}$ are then assessed with respect to this curation model. In contrast, Data-IQ computes learning dynamics for real samples in $D$ using a curator model trained on the same $D$ it assesses.
\item \textbf{Empirical performance}: In \name~, the curation aims at discarding synthetic samples which contradict the learning signal obtained from the real data. If we were to perform the curation like in Data-IQ, this implies that we would instead have to merge $D_{\mathrm{train}}$ and $D_{\mathrm{syn}}$, and train a curator model on $D_{\mathrm{merged}} = D_{\mathrm{train}} \cup D_{\mathrm{syn}}$ to assess $D_{\mathrm{syn}}$. Intuitively, with such an approach, the signal from the small $D_{\mathrm{train}}$ would be overshadowed by $D_{\mathrm{syn}}$, as the latter is a lot bigger in size ($1000$ samples), thus making the curation irrelevant.
\end{enumerate}

We now show empirical evidence regarding the last point, by comparing \name~ with a baseline where we train the curator on $D_{\mathrm{merged}} = D_{\mathrm{train}} \cup D_{\mathrm{syn}}$.
The results are in \cref{table:compare-diq}. We observe that:

\begin{enumerate}
    \item \name's curation outperforms Data-IQ curation on downstream performance across all datasets and $n$.
    \item Data-IQ curation does not always improve upon the uncurated baseline, which aligns with our intuition that $D_{\mathrm{syn}}$ overshadows $D_{\mathrm{train}}$ because of its size.
\end{enumerate}

\begin{table}[!ht]
\centering
\caption{Comparing the \name~ curation mechanism to Data-IQ curation \cite{seedat2022data}. We report the AUC averaged over 4 downstream models on $\mathcal{D}_{\mathrm{test}}$.}
\begin{tabular}{l|c|c|c}
\hline
Dataset & Uncurated & Curated (\name~) - Ours & Curated (Data-IQ) \\ \hline
covid (n=20)   & 73.78 & 73.87 & 73.45 \\ 
cutract (n=20) & 71.15 & 72.50 & 70.84 \\ 
maggic (n=20)  & 60.70 & 61.48 & 60.68 \\ 
seer (n=20)    & 84.53 & 84.82 & 83.97 \\ 
compas (n=20)  & 68.01 & 67.91 & 67.45 \\ 
adult (n=20)   & 50.39 & 71.48 & 43.59 \\ 
drug (n=20)    & 75.08 & 75.29 & 74.23 \\ \hline
covid (n=40)   & 73.40 & 73.95 & 73.36 \\ 
cutract (n=40) & 69.87 & 71.72 & 69.57 \\ 
maggic (n=40)  & 59.29 & 60.77 & 59.65 \\ 
seer (n=40)    & 84.29 & 84.93 & 84.47 \\ 
compas (n=40)  & 67.57 & 67.85 & 67.07 \\ 
adult (n=40)   & 48.31 & 73.82 & 48.92 \\ 
drug (n=40)    & 74.30 & 75.79 & 74.45 \\ \hline
covid (n=100)  & 73.77 & 74.71 & 73.77 \\ 
cutract (n=100)& 70.20 & 72.51 & 70.29 \\ 
maggic (n=100) & 58.98 & 61.32 & 58.66 \\ 
seer (n=100)   & 84.45 & 85.37 & 84.33 \\ 
compas (n=100) & 68.02 & 68.19 & 67.82 \\ 
adult (n=100)  & 46.09 & 74.57 & 45.13 \\ 
drug (n=100)   & 76.24 & 76.74 & 76.11 \\ \hline
covid (n=200)  & 73.40 & 74.62 & 73.27 \\ 
cutract (n=200)& 71.39 & 73.01 & 71.43 \\ 
maggic (n=200) & 58.92 & 61.41 & 58.71 \\ 
seer (n=200)   & 84.39 & 85.56 & 84.49 \\ 
compas (n=200) & 67.02 & 68.15 & 67.26 \\ 
adult (n=200)  & 40.96 & 75.84 & 41.01 \\ 
drug (n=200)   & 75.58 & 76.06 & 75.39 \\ \hline
\end{tabular}
\label{table:compare-diq}
\end{table}

\newpage
\subsection{Comparing \name~ vs TabPFN}\label{appendix:tabpfn}

We outlined in \cref{appendix:extended} various dimensions on which \name~ and transfer learning / meta-learning / few-shot learning differ. To provide further empirical evidence on why the above dimensions are important (notably the point on the choice of downstream backbone), we compare \name~ with TabPFN \cite{hollmann2022tabpfn}, a few-shot learning method designed for small tabular problems. We chose TabPFN because it meets our criteria where it does not require access to external datasets, since the model is pretrained on an extensive set of synthetic tabular datasets and can perform few-shot learning with its transformer backbone.

We use the pretrained model released by the authors at \url{https://github.com/automl/TabPFN} and show the results in \cref{tab:tabpfn}.

\begin{table}[!h]
\centering
\caption{Comparison of \name~ vs TabPFN. We report the AUC averaged over 4 downstream models on $\mathcal{D}_{\mathrm{test}}$.}
\begin{tabular}{l|cc|c}
\hline
Dataset & Uncurated & \makecell{Curated \\ (\name~)} & TabPFN \\ \hline
covid (n=20)   & 73.78 & 73.87 & 66.31 \\ 
cutract (n=20) & 71.15 & 72.50 & 63.86 \\ 
maggic (n=20)  & 60.70 & 61.48 & 55.49 \\ 
seer (n=20)    & 84.53 & 84.82 & 75.30 \\ 
compas (n=20)  & 68.01 & 67.91 & 57.04 \\ 
adult (n=20)   & 50.39 & 71.48 & 71.70 \\ 
drug (n=20)    & 75.08 & 75.29 & 69.18 \\ \hline
covid (n=40)   & 73.40 & 73.95 & 67.26 \\ 
cutract (n=40) & 69.87 & 71.72 & 65.93 \\ 
maggic (n=40)  & 59.29 & 60.77 & 57.75 \\ 
seer (n=40)    & 84.29 & 84.93 & 79.48 \\ 
compas (n=40)  & 67.57 & 67.85 & 57.72 \\ 
adult (n=40)   & 48.31 & 73.82 & 75.36 \\ 
drug (n=40)    & 74.30 & 75.79 & 71.32 \\ \hline
covid (n=100)  & 73.77 & 74.71 & 69.50 \\ 
cutract (n=100)& 70.20 & 72.51 & 70.01 \\ 
maggic (n=100) & 58.98 & 61.32 & 59.07 \\ 
seer (n=100)   & 84.45 & 85.37 & 80.96 \\ 
compas (n=100) & 68.02 & 68.19 & 63.42 \\ 
adult (n=100)  & 46.09 & 74.57 & 77.36 \\ 
drug (n=100)   & 76.24 & 76.74 & 72.68 \\ \hline
covid (n=200)  & 73.40 & 74.62 & 70.82 \\ 
cutract (n=200)& 71.39 & 73.01 & 71.59 \\ 
maggic (n=200) & 58.92 & 61.41 & 60.67 \\ 
seer (n=200)   & 84.39 & 85.56 & 82.05 \\ 
compas (n=200) & 67.02 & 68.15 & 65.51 \\ 
adult (n=200)  & 40.96 & 75.84 & 79.18 \\ 
drug (n=200)   & 75.58 & 76.06 & 74.29 \\ \hline
\end{tabular}

\label{tab:tabpfn}
\end{table}

\textbf{Takeaways:}
We see \name~ outperforms TabPFN on $6/7$ datasets, for the different $n$. The performance gains are especially noticeable in the ultra low-sample regime ($n=20$).

Finally, we acknowledge that \name~ is not a one-size-fits-all approach. When the assumptions underpinning transfer learning or meta-learning hold (e.g. availability of external datasets), combining ideas from these learning paradigms with the augmentation methodology of \name~ could constitute an interesting research direction, but this falls beyond the scope of our current work.

\subsection{\name~ in specialized domains.}

We conduct an experiment with the additional dataset Higgs \cite{misc_higgs_280}. We chose this dataset because it comes from a specialized domain (physics), where the dataset consists in kinematic properties of particles measured by the particle detectors of an accelerator, which is likely under-represented in the LLMs training corpus.

We note that the contextual information for this dataset is quite specific, as can be seen from the names of the features, which include for example "m\_bb", "m\_jj", "m\_jjj". This particular contextual information, which is not as semantically meaningful as other datasets, makes it interesting to compare the performance of \name~ with the traditional baselines. We show the results in \cref{tab:higgs_performance}.
As we can see, the performance of \name~ is good for the smaller $n$. As $n$ increases, the downstream task benefits more from the use of the other baselines. We highlight that in this case, our curation mechanism still benefits both \name~ and the baselines.

\begin{table}[!h]
\centering
\caption{\name~ performance on the Higgs dataset}
\begin{tabular}{l|cc|cc|cc|ccc}

Dataset      & \multicolumn{2}{c|}{\name~} & \multicolumn{2}{c|}{CTGAN} & \multicolumn{2}{c|}{TVAE} & \multicolumn{2}{c}{NFLOW} \\ \hline
             & Uncur & Cur  & Uncur & Cur  & Uncur & Cur  & Uncur & Cur  \\ \hline
higgs (n=20) & 65.82 & 70.25 & 62.73 & 67.31 & 59.14 & 69.89 & 56.45 & 69.63 \\ 
higgs (n=40) & 65.38 & 71.01 & 63.07 & 70.41 & 59.29 & 67.61 & 60.27 & 71.54 \\ 
higgs (n=100)& 68.62 & 73.42 & 68.46 & 75.59 & 68.90 & 76.05 & 59.73 & 75.86 \\ 
higgs (n=200)& 66.39 & 75.18 & 74.27 & 79.75 & 73.65 & 79.32 & 60.99 & 79.42 \\ \hline
\end{tabular}

\label{tab:higgs_performance}
\end{table}

\subsection{Behavior beyond the low-data regime.}

We examine the behavior of baselines beyond $n=200$ and the low-data regime. We provide a plot in Figure \ref{fig:large_n} for three datasets examining the behavior of CTGAN, TVAE and NFLOW for increasing $n$. We fix the \name~ method at $n=200$ (as we do not see too much of an additional performance increase, due to the LLM context window which limits the number of in-context samples we can provide).

\begin{figure}[!h]
    \centering
    \includegraphics[width=0.75\linewidth]{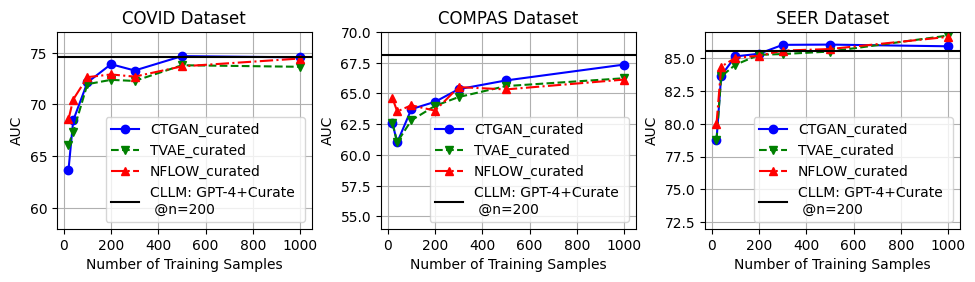}
    \caption{Behavior of baselines at high $n$, i.e. beyond the low data regime.}
    \label{fig:large_n}
\end{figure}

\textbf{Takeaways}:
\begin{enumerate}
    \item Other baselines initially improve with more real samples, however their performance gains tend to plateau out at around $n \in \{200, 400 \}$ samples. This suggests we only get minimal gains as increasing numbers of samples are added to the baselines.
\item \name~ either outperforms the baselines even for very high $n$ (Compas, Covid) or remains competitive (SEER), even when the number of samples used by CTGAN, TVAE and NFLOW is twice or thrice bigger.
\end{enumerate}

\subsection{Addressing potential biases from the LLM} \label{subsec:bias}

There are many challenges which arise from the use of LLMs. An example is potential biases from the LLM which might stem from their training data and might affect our synthetic data generated by the LLM.

We first describe some considerations which show how a practicioner using \name~ can address this issue.

\begin{enumerate}
    \item  \textbf{Choice of LLM:} Different LLMs may exhibit different levels of bias, due to their respective pretraining and alignment with human feedback. With \name~, the practitioner has a lot of flexibility in the choice of the LLM. This is an important aspect when it comes to bias. The practitioner may want to minimize the risk of having bias in the augmented data by selecting one particular LLM for their task based on prior knowledge about different LLMs and their respective potential bias \cite{gallegos2023bias}.

\item \textbf{Fairness analysis on the augmented data:} Practitioners can directly assess potential bias in the data generated by \name~: (i) by doing a standard fairness analysis and computing different fairness metrics \citep{dwork2012fairness, hardt2016equality, mehrabi2021survey} on the generated data, as can be done more generally when building a model with any given dataset. (ii) If some bias is detected, the practitioner can use any off-the-shelf debiasing method \citep{calmon2017optimized, feldman2015certifying}, or use a training objective to make the downstream model fair \citep{berk2017convex, zhang2018mitigating}. Note that this approach is possible because \name~ is a data augmentation method, which gives a lot of flexibility for the practitioner to adjust the data according to their needs.

\item \textbf{CLLM can reduce representation bias:} While these aforementioned considerations involve choices external to \name~, we also want to emphasize that our \name~ approach has the potential to already address bias indirectly. As shown in \cref{fig:nsweep}, underrepresented groups in the population benefit the most from \name~. This suggests that we could use \name~ to remove representation bias, by leveraging \name~ to augment $\mathcal{D}_{\mathrm{train}}$ with synthetic data from underrepresented groups. \cref{tab:subgroup} and \cref{tab:metrics_covid} show that data generated with \name~ better aligns with the ground-truth distribution, compared to using a widely used conventional generative approach (TVAE).

\item \textbf{Curation aligns the feature/label relationships:} The curation step of \name~ can already help indirectly address issues of bias in feature-label relationships if training data $\mathcal{D}_{\mathrm{train}}$ is unbiased, even if it is not its primary purpose. The curation mechanism discards samples in $\mathcal{D}_{\mathrm{syn}}$ which do not obey the same feature / label relationship as in $\mathcal{D}_{\mathrm{train}}$. The results shown in \cref{fig:bayes_classifier} demonstrate that the curation step aligns the feature/label relationship between the curated set and the ground-truth distribution. Thus, assuming that $\mathcal{D}_{\mathrm{train}}$ is not biased and has the correct $Y|X$, discriminatory correlations stemming from the LLM bias could be detected in the form of feature/label relationships which differ from those in the gold standard $\mathcal{D}_{\mathrm{train}}$.
\end{enumerate}

We now provide additional empirical evidence into how curation can help with bias, in a synthetic setup.

\textbf{Synthetic setup}: We consider a synthetic setup where $X$ and $Y$ are random variables such that $X \in \mathbb{R}^{2}$ denotes the features, and $Y \in \{0 , 1\}$ is the label. Furthermore, we let $X_{1}$ (the first coordinate of $X$) be a sensitive attribute. We then create a biased distribution such that a downstream predictor $\hat{Y}$ trained on this biased distribution will violate the fairness criterion of equalized odds \cite{hardt2016equality}.

Formally, we consider a mixture of Gaussians to define $X$, i.e. $X = ZA + (1-Z)B$ where $Z \sim Ber(1/2)$, $A \sim \mathcal{N}([-1.5,0], I_{2})$ and $B \sim \mathcal{N}([1.5,0], I_{2})$. Furthermore, we let $Y = Z$.

In order to introduce some bias in the data, we consider the variable $Y'$ such that $Y' = 1-Y$ if $|X_1| > 2.5$, else $Y' = Y$.

Let $\hat{Y}$ denote a downstream predictor, supposedly trained on samples drawn from the distribution of $(X,Y')$ (instead of the ground-truth distribution of $(X,Y)$). Intuitively, for $y \in \{0,1\}$, because of the definition of $Y'$, we can expect $P(\hat{Y} =y \mid~ \lvert X_1 \rvert >2.5, Y = y)$ to be much smaller than $P(\hat{Y} =y \mid \lvert X_1 \rvert \leq 2.5, Y = y)$, hence strongly violating equality of odds. To quantify that, we will compute the absolute equality of odds differences defined as 
$$\Delta_{Y=1}= \lvert P(\hat{Y} =~1 \mid~ \lvert~ X_1 \rvert \leq 2.5, Y = 1) - P(\hat{Y} =1 \mid~ \lvert X_1 \rvert > 2.5, Y = 1) \rvert$$ and $$\Delta_{Y=0}= \lvert P(\hat{Y} =0 \mid \lvert X_1 \rvert \leq 2.5, Y = 0) - P(\hat{Y} = 0 \mid \lvert X_1 \rvert > 2.5, Y = 0) \rvert$$

We investigate if curation can help address the bias. We generate a training dataset $\mathcal{D}_{\mathrm{train}}$ of size $n = 20$, by sampling independent samples from the distribution of $(X,Y)$. We generate an augmented dataset $\mathcal{D}_{\mathrm{syn}}$ of size $1000$, by sampling from the biased distribution of $(X,Y')$. We then curate $\mathcal{D}_{\mathrm{syn}}$ using the curation mechanism of \name~, and obtain $\mathcal{D}_{\mathrm{curated}} \subset \mathcal{D}_{\mathrm{syn}}$. Finally, we train three XGBoost models on each of the three datasets $\mathcal{D}_{\mathrm{train}}$, $\mathcal{D}_{\mathrm{syn}}$, and $\mathcal{D}_{\mathrm{curated}}$.

\textbf{Results.} We evaluate the performance of these downstream models on a held-out $\mathcal{D}_{\mathrm{test}}$. 

We then report the average test accuracy, along with $\Delta_{Y=0}$ and $\Delta_{Y=1}$ for 10 different seeds in Table \ref{table:bias}, which  demonstrate that the curation mechanism helps ensure that the bias present in the augmented dataset $\mathcal{D}_{\mathrm{syn}}$ does not propagate to the downstream model, as can be seen with the low values of $\Delta_{Y=0}$ and $\Delta_{Y=1}$ for the models trained on $\mathcal{D}_{\mathrm{curated}}$.

\begin{table}[h!]
    \centering
  \caption{Curation can help address bias.}
    \begin{tabular}{l|ccc}
        \toprule
        \textbf{Dataset used for downstream training} & \textbf{Test accuracy (\%)} ($\uparrow$) & $\Delta_{Y=1}$ ($\downarrow$) & $\Delta_{Y=0}$ ($\downarrow$) \\
        \midrule
        $\mathcal{D}_{\mathrm{syn}}$ (Biased) & 76.7 & 0.89 & 0.90 \\
        $\mathcal{D}_{\mathrm{train}}$ (Unbiased) & 90.0 & 0.17 & \textbf{0.07} \\
        $\mathcal{D}_{\mathrm{curated}}$ & \textbf{91.3} & \textbf{0.13} & 0.08 \\
        \bottomrule
    \end{tabular}

    \label{table:bias}
\end{table}

\subsection{Choice of thresholds}

In \name~ we have thresholds which are used as part of the curation mechanism. The intuition is that our choice of thresholds should discard hard samples, that is, samples for which the confidence is low while the aleatoric uncertainty is also low. Given this intuition, we set an adaptive threshold on the aleatoric uncertainty, with $\tau_{al} = 0.75 \cdot (\max(v_{al}(\mathcal{D}_{\mathrm{syn}}) - \min(v_{al}(\mathcal{D}_{\mathrm{syn}}) )$, where $v_{al}(\mathcal{D}_{\mathrm{syn}})$ denotes the set of aleatoric uncertainties for the samples in $\mathcal{D}_{\mathrm{syn}}$. Hence, this threshold is adaptive and depends on the dataset at hand, which reduces the number of degrees of freedom to 1, namely the choice of the confidence threshold. For the latter, we set $\tau_{conf} = 0.2$. While we do not claim that this value is optimal for all datasets, we find in practice that it is robust for datasets across different domains.

We explore other combinations of thresholds to confirm our intuition. In addition to the choice of thresholds used in our main experiments, we consider two alternatives:

\begin{itemize}
    \item Aggressive filtering: we set $\tau_{conf}= 0.95$ and $\tau_{al} = 0.2$.
    \item Permissive filtering: we set $\tau_{conf}= 0.5$ and $\tau_{al} = 0.05$
\end{itemize}

We then evaluate the test AUC for $n=20$, with an XGBoost as the downstream model, and GPT-3.5 as the LLM backbone. The results are shown in \cref{fig:thresholds}.

\begin{figure}[!h]
    \centering
    \includegraphics[width=0.5\linewidth]{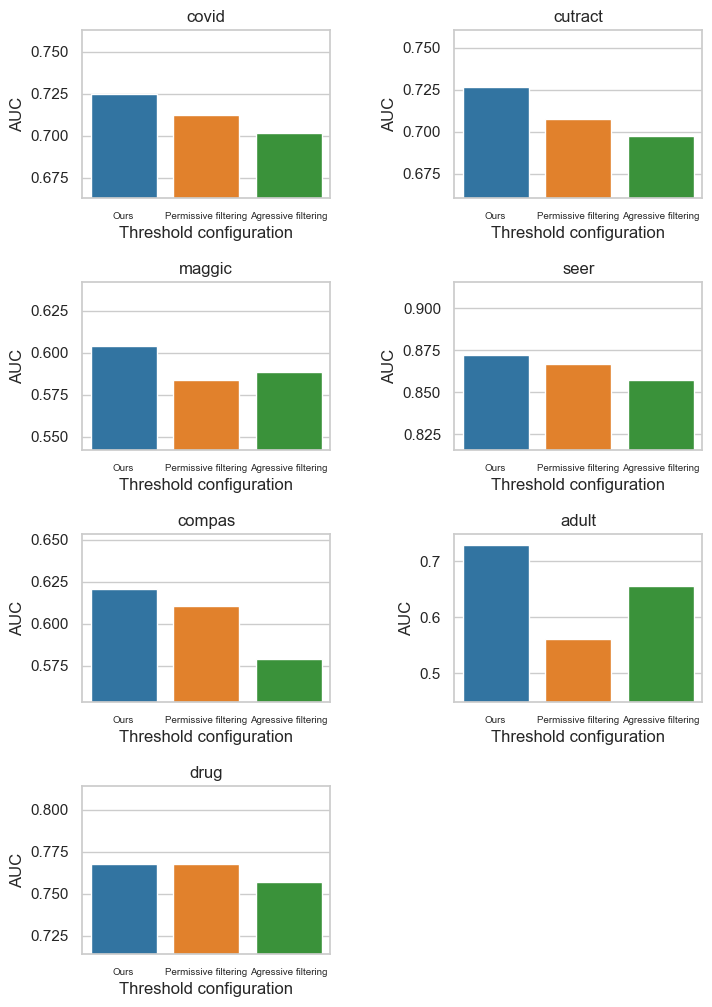}
    \caption{Assessment of the effects of different thresholds in \name~}
    \label{fig:thresholds}
\end{figure}

As we can see, our configuration strikes a good balance between the aggressive filtering and permissive filtering baselines, across the $7$ datasets.

In an ideal scenario, access to an external validation set could help determine optimal thresholds. However, given our focus on the low-sample regime ($n \leq 100$), we prioritized a versatile configuration that performs consistently across datasets and sample sizes.

\end{document}